%% file: main.tex
\documentclass[10pt,twocolumn,letterpaper]{article}

\usepackage{iccv}
\usepackage{times}
\usepackage{epsfig}
\usepackage{graphicx}
\usepackage{amsmath}
\usepackage{amssymb}

\usepackage{amsfonts} 
\usepackage{bbm}
\usepackage{booktabs}
\usepackage{xcolor}
\usepackage{makecell}
\usepackage{url}            
\usepackage{booktabs}       
\usepackage{amsfonts}       
\usepackage{nicefrac}       
\usepackage{microtype}      
\usepackage{graphicx}
\usepackage{graphics}
\usepackage{pifont}
\usepackage{color}
\usepackage{multirow}
\usepackage{adjustbox}
\usepackage{subcaption}
\usepackage{gensymb}
\usepackage{bm}
\usepackage{algorithm} 
\usepackage{algpseudocode}
\usepackage{wrapfig}
\usepackage{fancybox}
\usepackage{tikz}

\usepackage[accsupp]{axessibility}
\newcommand\method{{ClimateNeRF}}

\usepackage[pagebackref=true,breaklinks=true,letterpaper=true,colorlinks,bookmarks=false]{hyperref}
\usepackage[capitalize]{cleveref}
\crefname{section}{Sec.}{Secs.}
\Crefname{section}{Section}{Sections}
\Crefname{table}{Table}{Tables}
\crefname{table}{Tab.}{Tabs.}

\iccvfinalcopy %

\def\iccvPaperID{6482} %

\ificcvfinal\fi

\begin{document}
\include{macros}

\title{ClimateNeRF: Extreme Weather Synthesis in Neural Radiance Field}

\author{Yuan Li$^{1,2*}$ \quad Zhi-Hao Lin$^{1*}$ \quad  David Forsyth$^{1}$ \quad Jia-Bin Huang$^{3}$ \quad Shenlong Wang$^{1}$\\
$^{1}$University of Illinois at Urbana-Champaign \quad $^{2}$Zhejiang University \\ $^{3}$University of Maryland, College Park}

\twocolumn[{
\renewcommand\twocolumn[1][]{#1}
\maketitle
\input{figures/1_teaser-daf}

}]
\ificcvfinal\thispagestyle{empty}\fi

\input{0_abstraction.tex}
\input{1_introduction_v2}
\input{2_related_work.tex}

\input{3_method_v2.tex}

\input{4_experiment.tex}
\input{5_conclusion.tex}


\input{main.bbl}
\clearpage
\onecolumn
\appendix

\section*{\centering Supplementary Material}

\input{supp/abstract.tex}
\input{supp/implementation.tex}
\input{supp/ablation.tex}
\input{supp/control.tex}
\input{supp/qualitative.tex}
\input{supp/quantitative.tex}

\end{document}

%% file: macros.tex
\newcommand{\todocite}[1]{\textcolor{blue}{Citation needed []}}
\newcommand{\shenlongsay}[1]{\textcolor{blue}{[{\it Shenlong: #1}]}}

\newcommand{\todo}[1]{\textcolor{red}{[\textit{TODO: #1}]}}

\newcommand{\mfigure}[2]{\includegraphics[width=#1\linewidth]{#2}}
\newcommand{\mpage}[2]
{
\begin{minipage}{#1\linewidth}\centering
#2
\end{minipage}
}

\newcolumntype{L}[1]{>{\raggedright\let\newline\\\arraybackslash\hspace{0pt}}m{#1}}
\newcolumntype{C}[1]{>{\centering\let\newline\\\arraybackslash\hspace{0pt}}m{#1}}
\newcolumntype{R}[1]{>{\raggedleft\let\newline\\\arraybackslash\hspace{0pt}}m{#1}}

\newcommand{\xpar}[1]{\noindent\textbf{#1}\ \ }
\newcommand{\vpar}[1]{\vspace{3mm}\noindent\textbf{#1}\ \ }

\newcommand{\sect}[1]{Section~\ref{#1}}
\newcommand{\sects}[1]{Sections~\ref{#1}}
\newcommand{\eqn}[1]{Equation~\ref{#1}}
\newcommand{\eqns}[1]{Equations~\ref{#1}}
\newcommand{\fig}[1]{Figure~\ref{#1}}
\newcommand{\figs}[1]{Figures~\ref{#1}}
\newcommand{\tab}[1]{Table~\ref{#1}}
\newcommand{\tabs}[1]{Tables~\ref{#1}}

\newcommand{\ignorethis}[1]{}
\newcommand{\norm}[1]{\lVert#1\rVert}
\newcommand{\fcseven}{$\mbox{fc}_7$}

\renewcommand*{\thefootnote}{\fnsymbol{footnote}}

\def\naive{na\"{\i}ve\xspace}
\def\Naive{Na\"{\i}ve\xspace}

\makeatletter
\DeclareRobustCommand\onedot{\futurelet\@let@token\@onedot}
\def\@onedot{\ifx\@let@token.\else.\null\fi\xspace}

\def\iid{\emph{i.i.d}\onedot}
\def\eg{\emph{e.g}\onedot} \def\Eg{\emph{E.g}\onedot}
\def\ie{\emph{i.e}\onedot} \def\Ie{\emph{I.e}\onedot}
\def\cf{\emph{c.f}\onedot} \def\Cf{\emph{C.f}\onedot}
\def\etc{\emph{etc}\onedot} \def\vs{\emph{vs}\onedot}
\def\wrt{w.r.t\onedot} \def\dof{d.o.f\onedot}
\def\etal{\emph{et al}\onedot}
\makeatother

\definecolor{citecolor}{RGB}{34,139,34}
\definecolor{mydarkblue}{rgb}{0,0.08,1}
\definecolor{mydarkgreen}{rgb}{0.02,0.6,0.02}
\definecolor{mydarkred}{rgb}{0.8,0.02,0.02}
\definecolor{mydarkorange}{rgb}{0.40,0.2,0.02}
\definecolor{mypurple}{RGB}{111,0,255}
\definecolor{myred}{rgb}{1.0,0.0,0.0}
\definecolor{mygold}{rgb}{0.75,0.6,0.12}
\definecolor{myblue}{rgb}{0,0.2,0.8}
\definecolor{mydarkgray}{rgb}{0.66,0.66,0.66}

\newcommand{\myparagraph}[1]{\vspace{-6pt}\paragraph{#1}}

\newcommand{\bbR}{{\mathbb{R}}}
\newcommand{\bK}{\mathbf{K}}
\newcommand{\bX}{\mathbf{X}}
\newcommand{\bY}{\mathbf{Y}}
\newcommand{\bk}{\mathbf{k}}
\newcommand{\bx}{\mathbf{x}}
\newcommand{\by}{\mathbf{y}}
\newcommand{\bhy}{\hat{\mathbf{y}}}
\newcommand{\bty}{\tilde{\mathbf{y}}}
\newcommand{\bG}{\mathbf{G}}
\newcommand{\bI}{\mathbf{I}}
\newcommand{\bg}{\mathbf{g}}
\newcommand{\bS}{\mathbf{S}}
\newcommand{\bs}{\mathbf{s}}
\newcommand{\bM}{\mathbf{M}}
\newcommand{\bw}{\mathbf{w}}
\newcommand{\eye}{\mathbf{I}}
\newcommand{\bU}{\mathbf{U}}
\newcommand{\bV}{\mathbf{V}}
\newcommand{\bW}{\mathbf{W}}
\newcommand{\bn}{\mathbf{n}}
\newcommand{\bv}{\mathbf{v}}
\newcommand{\bq}{\mathbf{q}}
\newcommand{\bR}{\mathbf{R}}
\newcommand{\bi}{\mathbf{i}}
\newcommand{\bj}{\mathbf{j}}
\newcommand{\bp}{\mathbf{p}}
\newcommand{\bt}{\mathbf{t}}
\newcommand{\bJ}{\mathbf{J}}
\newcommand{\bu}{\mathbf{u}}
\newcommand{\bB}{\mathbf{B}}
\newcommand{\bD}{\mathbf{D}}
\newcommand{\bz}{\mathbf{z}}
\newcommand{\bP}{\mathbf{P}}
\newcommand{\bC}{\mathbf{C}}
\newcommand{\bA}{\mathbf{A}}
\newcommand{\bZ}{\mathbf{Z}}
\newcommand{\bff}{\mathbf{f}}
\newcommand{\bF}{\mathbf{F}}
\newcommand{\bo}{\mathbf{o}}
\newcommand{\bc}{\mathbf{c}}
\newcommand{\bT}{\mathbf{T}}
\newcommand{\bQ}{\mathbf{Q}}
\newcommand{\bL}{\mathbf{L}}
\newcommand{\bl}{\mathbf{l}}
\newcommand{\ba}{\mathbf{a}}
\newcommand{\bE}{\mathbf{E}}
\newcommand{\bH}{\mathbf{H}}
\newcommand{\bd}{\mathbf{d}}
\newcommand{\br}{\mathbf{r}}
\newcommand{\bb}{\mathbf{b}}
\newcommand{\bh}{\mathbf{h}}

\newcommand{\btheta}{\bm{\theta}}
\newcommand{\bhh}{\hat{\mathbf{h}}}
\newcommand{\ci}{{\cal I}}
\newcommand{\ct}{{\cal T}}
\newcommand{\co}{{\cal O}}
\newcommand{\ck}{{\cal K}}
\newcommand{\cu}{{\cal U}}
\newcommand{\cS}{{\cal S}}
\newcommand{\cQ}{{\cal Q}}
\newcommand{\cT}{{\cal S}}
\newcommand{\cC}{{\cal C}}
\newcommand{\cE}{{\cal E}}
\newcommand{\cF}{{\cal F}}
\newcommand{\cL}{{\cal L}}
\newcommand{\X}{{\cal{X}}}
\newcommand{\Y}{{\cal Y}}
\newcommand{\cH}{{\cal H}}
\newcommand{\cP}{{\cal P}}
\newcommand{\cN}{{\cal N}}
\newcommand{\cU}{{\cal U}}
\newcommand{\cV}{{\cal V}}
\newcommand{\cX}{{\cal X}}
\newcommand{\cY}{{\cal Y}}
\newcommand{\graph}{{\cal H}}
\newcommand{\bayes}{{\cal B}}
\newcommand{\cx}{{\cal X}}
\newcommand{\cg}{{\cal G}}
\newcommand{\cm}{{\cal M}}
\newcommand{\cM}{{\cal M}}
\newcommand{\cG}{{\cal G}}
\newcommand{\cR}{\cal{R}}
\newcommand{\R}{\cal{R}}
\newcommand{\eig}{\mathrm{eig}}

\newcommand{\bbS}{\mathbb{S}}

\newcommand{\D}{{\cal D}}
\newcommand{\bfp}{{\bf p}}
\newcommand{\bfd}{{\bf d}}

\newcommand{\cv}{{\cal V}}
\newcommand{\ce}{{\cal E}}
\newcommand{\cy}{{\cal Y}}
\newcommand{\cz}{{\cal Z}}
\newcommand{\cb}{{\cal B}}
\newcommand{\cq}{{\cal Q}}
\newcommand{\cd}{{\cal D}}
\newcommand{\bcf}{{\cal F}}
\newcommand{\cI}{\mathcal{I}}

\newcommand{\ut}{^{(t)}}
\newcommand{\up}{^{(t-1)}}

\newcommand{\bpi}{\boldsymbol{\pi}}
\newcommand{\bphi}{\boldsymbol{\phi}}
\newcommand{\bPhi}{\boldsymbol{\Phi}}
\newcommand{\bmu}{\boldsymbol{\mu}}
\newcommand{\bSigma}{\boldsymbol{\Sigma}}
\newcommand{\bGamma}{\boldsymbol{\Gamma}}
\newcommand{\bbeta}{\boldsymbol{\beta}}
\newcommand{\bomega}{\boldsymbol{\omega}}
\newcommand{\blambda}{\boldsymbol{\lambda}}
\newcommand{\bkappa}{\boldsymbol{\kappa}}
\newcommand{\btau}{\boldsymbol{\tau}}
\newcommand{\balpha}{\boldsymbol{\alpha}}
\def\bgamma{\boldsymbol\gamma}

\newcommand{\prox}{{\mathrm{prox}}}

\newcommand{\pardev}[2]{\frac{\partial #1}{\partial #2}}
\newcommand{\dev}[2]{\frac{d #1}{d #2}}
\newcommand{\dw}{\delta\bw}
\newcommand{\lab}{\mathcal{L}}
\newcommand{\unlab}{\mathcal{U}}
\newcommand{\ind}{1{\hskip -2.5 pt}\hbox{I}}
\newcommand{\ff}[2]{   \cf_{\prec (#1 \rightarrow #2)}}
\newcommand{\vv}[2]{   \cv_{\prec (#1 \rightarrow #2)}}
\newcommand{\dd}[2]{   \delta_{#1 \rightarrow #2}}
\newcommand{\ld}[2]{   \lambda_{#1 \rightarrow #2}}
\newcommand{\en}[2]{  \bD(#1|| #2)}
\newcommand{\ex}[3]{  \bE_{#1 \sim #2}\left[ #3\right]} 
\newcommand{\exd}[2]{  \bE_{#1 }\left[ #2\right]}

\newcommand{\se}[1]{\mathfrak{se}(#1)}
\newcommand{\SE}[1]{\mathbb{SE}(#1)}
\newcommand{\so}[1]{\mathfrak{so}(#1)}
\newcommand{\SO}[1]{\mathbb{SO}(#1)}

\newcommand{\poselow}{\xi}
\newcommand{\pose}{\bm{\poselow}}
\newcommand{\linpose}{\pose^\ell}
\newcommand{\cbpose}{\pose^c}
\newcommand{\rateparam}{v_i}
\newcommand{\bapose}{\bm{\poselow}_i}
\newcommand{\trackingpose}{\bm{\poselow}}
\newcommand{\rotlow}{\omega}
\newcommand{\rot}{\bm{\rotlow}}
\newcommand{\translow}{v}
\newcommand{\trans}{\bm{\translow}}
\newcommand{\hnorm}[1]{\left\lVert#1\right\rVert_{\gamma}}
\newcommand{\lnorm}[1]{\left\lVert#1\right\rVert}
\newcommand{\barate}{v_i}
\newcommand{\trackingrate}{v}
\newcommand{\imgpt}{\mathbf{u}_{i,k,j}}
\newcommand{\mappt}{\mathbf{X}_{j}}
\newcommand{\timet}[1]{\bar{t}_{#1}}
\newcommand{\mf}[1]{\text{MF}_{#1}}
\newcommand{\kmf}[1]{\text{KMF}_{#1}}
\newcommand{\Exp}{\text{Exp}}
\newcommand{\Log}{\text{Log}}

\newcommand{\shiftleft}[2]{\makebox[0pt][r]{\makebox[#1][l]{#2}}}
\newcommand{\shiftright}[2]{\makebox[#1][r]{\makebox[0pt][l]{#2}}}

%% file: figures/1_teaser-daf.tex
\begin{center}
\vspace{-9mm}
\setlength{\fboxsep}{0pt}
\raisebox{5\normalbaselineskip}[0pt][0pt]{\rotatebox[origin=c]{90}{Climate events}} \hfill
{\tikz{
\node[draw=black, line width=.4mm, inner sep=0pt] (flood) at (0,0)
    {\includegraphics[width=.46\textwidth]{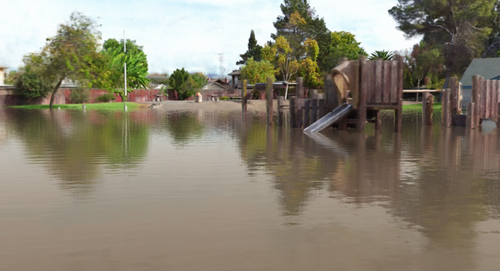}};
\node[draw=black, line width=.4mm, inner sep=0pt] (origin) at (-2.09,1.12)
    {\includegraphics[width=.2\textwidth]{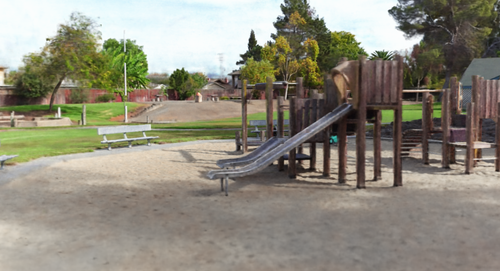}};
    \node[draw=black, draw opacity=1.0, line width=.3mm, fill opacity=0.7,fill=white, text opacity=1] at (-3.195, 0.45) {\small Original};
    \node[draw=black, draw opacity=1.0, line width=.3mm, fill opacity=0.7,fill=white, text opacity=1] at (-3.12, -1.81) {\large \ Flood \ };
}}
\hfill
{\tikz{
\node[draw=black, line width=.4mm, inner sep=0pt] (snow) at (0,0)
    {\includegraphics[width=.46\textwidth]{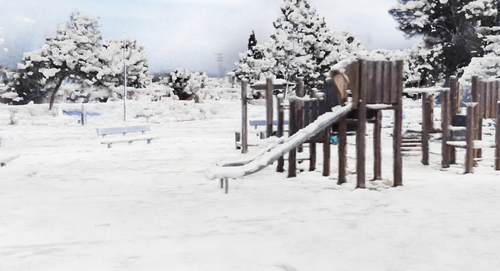}};
    \node[draw=black, draw opacity=1.0, line width=.3mm, fill opacity=0.7,fill=white, text opacity=1] at (-3.13, -1.81) {\large \ Snow \ };
}} \\

\raisebox{5\normalbaselineskip}[0pt][0pt]{\rotatebox[origin=c]{90}{Novel views}} \hfill
{\tikz{
\node[draw=black, line width=.4mm, inner sep=0pt] (view2) at (0,0)
    {\includegraphics[width=.46\textwidth]{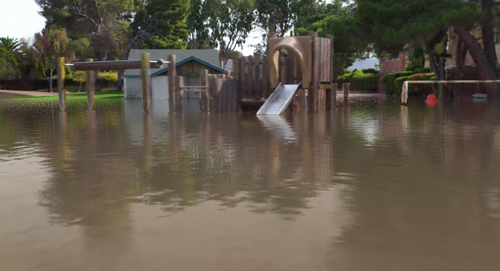}};
    \node[draw=black, draw opacity=1.0, line width=.3mm, fill opacity=0.7,fill=white, text opacity=1] at (-3.02, -1.82) {\large \ View 1 \ };
}}
\hfill
{\tikz{
\node[draw=black, line width=.4mm, inner sep=0pt]  (view3) at (0,0)
    {\includegraphics[width=.46\textwidth]{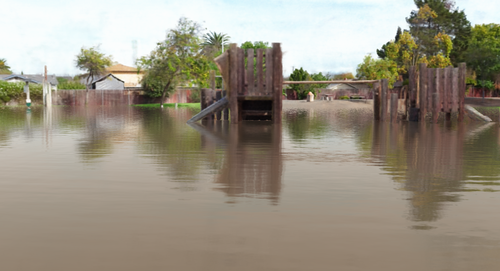}};
    \node[draw=black, draw opacity=1.0, line width=.3mm, fill opacity=0.7,fill=white, text opacity=1] at (-3.02, -1.82) {\large \ View 2 \ };
}}
\captionof{figure}{
\textbf{Extreme weather synthesis.}
We fuse NeRF modeling and physical simulation to produce 3D consistent renderings of scenes with simulated physical effects. We apply our method to climate effect simulation: what will it look like if the playground floods? or is covered in snow?  Note in particular reflections and ripple effects on water; accumulated snow on horizontal surfaces; trees darkened by wintertime; and consistency of geometry (but not ripples!). \href{https://climatenerf.github.io/}{Project page}}
\vspace{-3mm}
\label{fig:1_teaser}
\end{center}

%% file: 0_abstraction.tex
\footnotetext{
* Both authors contributed equally to this research. }

\begin{abstract}
  Physical simulations produce excellent predictions of weather effects.
  Neural radiance fields produce SOTA scene models. 
  We describe a novel NeRF-editing procedure that can fuse physical simulations with NeRF models of scenes,  producing realistic movies of physical phenomena in
those scenes.  
Our application -- Climate NeRF -- allows people to visualize what climate change outcomes will do {\em to them}.  

ClimateNeRF allows us to render realistic weather effects, including smog, snow, and flood.
Results can be controlled with physically meaningful variables like water level.
Qualitative and quantitative studies show that our
simulated results are significantly more realistic than those from 
SOTA 2D image editing and SOTA 3D NeRF stylization.
\end{abstract}

%% file: 1_introduction_v2.tex
\section{Introduction}

This paper describes a novel procedure that fuses graphic simulations
with NeRF models~\cite{mildenhall2020nerf, muller2022instant} of scenes to produce realistic movies of physical phenomena in those scenes.  
We apply our method to produce compelling simulations of possible extreme weather outcomes -- what would the playground look like after a minor flood?  a major flood?
a blizzard?

\input{figures/3_overview.tex}

Our application %
is aimed at an important problem.
Cumulative small changes are hard to reason about, and most people find
it difficult to visualize what climate change outcomes will do {\em to them} ~\cite{chapman2016climate,
  glaas2017visualization, herring2017communicating, leviston2014imagining}. 
Steps to slow $\mbox{CO}_2$ emissions (say, reducing fossil fuel use) or to moderate outcomes (say, building flood control
measures) come with immediate costs and distant benefits.  
It is hard to support such steps if one can't visualize their effects.

Traditional graphic simulations can produce realistic weather effects for 3D scenes in a conventional simulation pipeline~\cite{stomakhin2013material, feldman2002modeling, horvath2015empirical, zsolnai2022flow, blender,unrealengine, haas2014history}. 
But these methods operate on conventional polygon models.
Building polygon models that produce compelling renderings from a few images of a scene remains challenging. 
Neural radiance fields (NeRFs) produce photorealistic 3D scene models from few images~\cite{mildenhall2020nerf, barron2022mip, muller2022instant, chen2022tensorf, tancik2022block, wu2022diver, lin2022neurmips} but (as far as we know) have rarely been investigated together with graphics simulations. Our method draws from a large literature, reviewed below, that explores editing these models.

Although there are numerous variants of NeRF, most are primarily focused on the task of novel-view synthesis. 
Directly integrating existing NeRF models for graphics simulation yields unsatisfactory results (Fig.~\ref{fig:geometry_improvement}). 
Specifically, we found the geometry of current NeRF models is not sufficiently accurate to enable physics-inspired interactions, and they cannot provide a realistic and complete illumination field to relight the inserted physics entities, such as water and snow. 
To address this challenge, we have systematically investigated various techniques to improve NeRF models, focusing on answering the critical research question of {\it ``what constitutes a good simulatable NeRF representation?''}. 
Our resulting NeRF variant, ClimateNeRF, boasts high-quality geometry, a complete illumination field, and rich semantics, making it particularly well-suited for complex downstream editing and simulation tasks (Sec.~\ref{sec:modeling}).

ClimateNeRF allows us to render realistic weather effects, including smog, snow, and flood.  These effects are
consistent over frames so that compelling movies result.  
At a high level, we:  adjust scene images to reflect the global effects of physics; build a NeRF model of a
scene from those adjusted images; recover an approximate geometric representation; apply the physical simulation in that
geometry; then render using a novel ray tracer.  
Adjusting the images is essential.  
For example, trees tend to have less saturated images in winter.  
We use a novel style transfer approach in an NGP framework
to obtain these global effects without changing scene geometry (Sec.~\ref{sec:stylization}). 
Our ray tracer merges the physical and NeRF models by carefully accounting for ray effects during rendering (Sec.~\ref{sec:simulation}). An eye ray could, for example, first encounter a high NeRF density (and so return the usual result); or it
could strike an inserted water surface (and so be reflected to query the model again).

We demonstrate the effectiveness of ClimateNeRF in various 3D scenes from the Tanks and Temple, MipNeRF360, and
KITTI-360 datasets~\cite{barron2022mip, Knapitsch2017, liao2022kitti}.  We compared to the state-of-the-art 2D image editing
methods, such as stable diffusion inpainting~\cite{rombach2021highresolution}, ClimateGAN~\cite{schmidt2021climategan};
state-of-the-art 3D NeRF stylization~\cite{zhang2022arf}.  Both qualitative and quantitative studies show that our
simulated results are significantly more realistic than the other competing methods.  Furthermore, we also demonstrate
the controllability of our physically-inspired approaches, such as changing the water level, wind strength and
direction, and thickness of snow and smog.  

Our approach results in view consistency (so we can make movies, which is difficult to do with frame-by-frame synthesis); compelling photorealism (because the scene is a NeRF representation); and is controllable (because we can adjust physically meaningful parameters in the simulation).
As Fig.~\ref{fig:1_teaser} illustrates, results are photo-realistic, physically plausible, and temporally consistent.

Our contributions are three-fold:
\begin{enumerate}
    \item We investigate the challenging problem of repurposing NeRF for realistic simulation and develop a novel framework that can effectively reduce NeRF modeling errors and facilitate graphical simulation. 
    \item We propose ClimateNeRF, a novel solution that can render realistic weather effects, including smog, snow, and flood, over real-world images.
    \item We demonstrate our proposed approach can produce compelling free-viewpoint movies that are photorealistic, geometrically consistent, and controllable.
\end{enumerate}

%% file: figures/3_overview.tex
\begin{table*}
\centering
\footnotesize
\resizebox{\linewidth}{!}{
    \setlength{\tabcolsep}{0.05em}
    \begin{tabular}{cccc}
    \footnotesize
        \includegraphics[width=0.25\linewidth, trim={0 -0.9cm 0 0.9cm}]{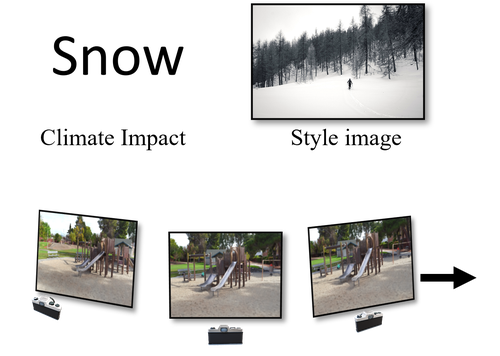} &
        \includegraphics[width=0.157\linewidth, trim={-0.1cm -0.1cm 0 0.05cm}]{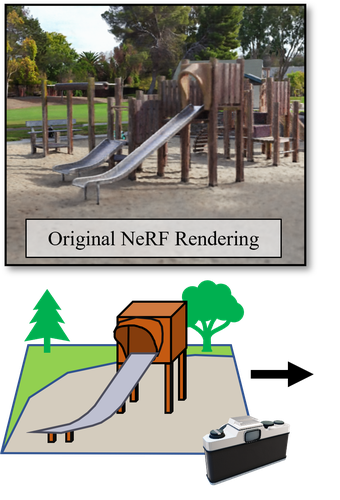} &
        \includegraphics[width=0.16\linewidth, trim={-0.3cm -0.15cm 0.15cm -0.2cm}]{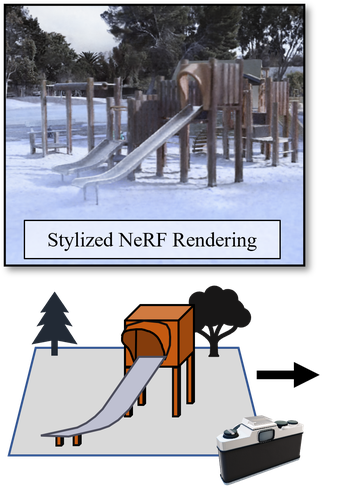} &
        \includegraphics[width=0.163\linewidth, trim={-0.05cm 0.02cm 0 0}]{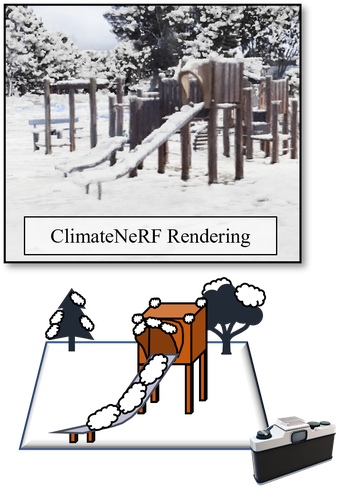} \\
        \shiftleft{40pt}{\scriptsize Multi-view Input Images} &
        \shiftleft{35pt}{\makecell[c]{\scriptsize 3D Scene Modeling\\\scriptsize Sect.~\ref{sec:modeling}}} &
        \shiftleft{40pt}{\makecell[c]{\scriptsize Neural Scene Stylization\\\scriptsize Sect.~\ref{sec:stylization}}}& 
        \shiftleft{45pt}{\makecell[c]{\scriptsize Physically-Inspired Simulation\\\scriptsize Sect.~\ref{sec:simulation}, ~\ref{sec:implementation}}} \\
    \end{tabular}
}
\vspace{-3mm}
\captionof{figure}{
\textbf{Method overview.} 
Our method takes multiple posed images, the targeted climate event simulation (e.g., snow),  and optionally a user-selected style image %
as inputs.
First, we reconstruct the 3D scene using instant NGP~\cite{muller2022instant} (a variant of NeRF) (Sect.~\ref{sec:modeling}). 
The reconstructed radiance fields allow us to synthesize high-quality novel view imagery of the scene efficiently. 
Second, we optionally finetune the learned instant-NGP model so that it captures the styles of the provided style image (Sect.~\ref{sec:stylization}).
Such 3D consistent stylization is particularly useful for modeling weather effects that are hard to capture via physical simulation.
Third, we simulate the climate events by integrating the relevant physical entities (snow, water, smog) to the scene and rendering physically plausible images.
}
\label{fig:3_overview}
\vspace{-4mm}
\end{table*}

%% file: 2_related_work.tex
\section{Related Work}

\noindent{\bf Climate simulation:}
The importance of making climate simulations accessible is well known. 
\cite{schmidt2019visualizing} collects climate
images dataset and performs image editing with CycleGAN~\cite{zhu2017unpaired}. 
~\cite{cosne2020using,schmidt2021climategan} leverage depth information to estimate water mask, and perform GAN-based image editing and inpainting. 
~\cite{hahner2019semantic} simulate fog and snow.
These methods offer realistic effects for a single image, but cannot
provide immersive, view-consistent climate simulation. 
In contrast, ClimateGAN allows the view to move without artifacts.

\noindent{\bf Novel view synthesis:}
Neural Radiance Field (NeRF)~\cite{mildenhall2020nerf, barron2021mip, barron2022mip, verbin2022refnerf} leverage differentiable rendering for scene reconstruction producing photorealistic novel view synthesis. 
Training and rendering can be accelerated using multiple smaller MLPs and customized CUDA kernel functions~\cite{reiser2021kilonerf, lin2022neurmips}.
Sample efficiency can be improved by modeling the light field~\cite{attal2022learning} or calculating ray intersections with explicit geometry, such as voxel grids~\cite{liu2020neural, wu2022diver, reiser2021kilonerf}, octrees~\cite{yu2021plenoctrees}, planes~\cite{Wizadwongsa2021NeX, lin2022neurmips}, and point cloud~\cite{xu2022point, zuo2022view}. 
Explicit geometric representations improve efficiencies~\cite{sun2022direct, yu2021plenoxels, martel2021acorn, xu2022point, zhang2022differentiable}.
We build on Instant-NGP~\cite{muller2022instant} as it offers fast learning and rendering, is more memory efficient than grid-like structures, and accelerates our simulation pipeline.  

\noindent{\bf Manipulating neural radiance fields:}
NeRF entangles lighting, geometry, and materials, making appearance editing hard. 
A line of
work~\cite{srinivasan2021nerv, rudnev2021neural, zhang2021nerfactor, zhang2021physg, li2022physically, boss2021nerd,
  boss2021neural} decomposes neural scene representations into familiar scene variables (like environment lighting,
surface normal, diffuse color, or BRDF). 
Each can then be edited with predictable results on rendering. 
Image segmentation and inpainting can be ``lifted'' to 3D, allowing object
removal~\cite{kobayashi2022distilledfeaturefields, benaim2022volumetricdisentanglement}.  
An alternative is to edit source images with text~\cite{wang2022clip}. 
In contrast, our method does not directly edit the NeRF while producing substantial changes to scene appearance.
Geometry editing by transforming a NeRF into a mesh and manipulating that (so editing shapes and compositing) has been
demonstrated~\cite{xu2022deforming,   Yuan22NeRFEditing, neumesh}.
In contrast, our method renders the original NeRF together with simulation results.

\input{figures/rendering}

\input{figures/geometry_improvement}

\noindent{\bf Style transfer:}
Data-driven 2D stylization~\cite{karras2019style, park2020swapping, jiang2020tsit} is a powerful tool to change image color and
texture, and can even change the content following masks and labels provided by users~\cite{liu2022asset, ling2021editgan, zhuang2021enjoy, abdal2020image2stylegan++}. 
Diffusion models~\cite{ruiz2022dreambooth, rombach2021highresolution, saharia2022photorealistic} can generate extraordinary image
quality based on text input. 
However, these 2D-based methods cannot currently generate 3D-consistent view synthesis; ours can. 
Another line of work~\cite{chiang2022stylizing, chen2022upst, zhang2022arf}  performs stylization on neural radiance fields. Given an style image, one changes the scene color into the same style.  
This line of work focuses on artistic effects; in contrast, our changes are driven by physical simulations.

\noindent{\bf Physically-based simulation of weather:}
Physical simulation of weather in computer graphics has too long a history to allow comprehensive review here.
Fournier and Reeves obtain excellent capillary wave simulations with simple Fourier transform reasoning~\cite{fournier1986simple} (and we
adopt their method), with modifications by~\cite{horvath2015empirical, tessendorf2001simulating};\cite{fedkiw2001visual}
simulates smoke; and~\cite{nishita1997modeling, feldman2002modeling, 
  stomakhin2013material} simulate snow in wind using metaballs and fluid dynamics.  
Our method demonstrates how to benefit from such simulations while retaining the excellent scene modelling properties of NeRF style models.

%% file: figures/rendering.tex
\begin{figure}[t]
  \centering
  \includegraphics[width=0.9\linewidth]{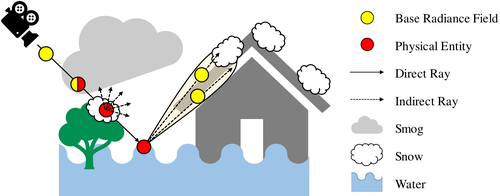}
   \caption{
\textbf{Rendering Procedure of ClimateNeRF.} 
We first determine the position of physical entities (smog particle, snow balls, water surface) with physical simulation. 
We can then render the scene with desired effects by modeling the light transport between the physical entities and the scene.
More specifically, we follow the volume rendering process and fuse the estimated color and density from 
1) the original radiance field (by querying the trained instant-NGP model) and 
2) the physical entities (by physically based rendering).
Our rendering procedure thus maintain the realism while achieving complex, yet physically plausible visual effects.
}
   \label{fig:3_rendering}
   \vspace{-5mm}
\end{figure}

%% file: figures/geometry_improvement.tex
\begin{table*}[t]
  \centering
  \footnotesize
  \setlength\tabcolsep{0.05em}
  \resizebox{\textwidth}{!}{
    \begin{tabular}{ccccc}

        \raisebox{5mm}[0pt][0pt]{\rotatebox[origin=c]{90}{\tiny{NGP~\cite{muller2022instant}}}} & 
        \tikz{
        \node[draw=black, line width=.3mm, inner sep=0pt] (ngp) at (0,0) {
        \includegraphics[width=0.118\textwidth]{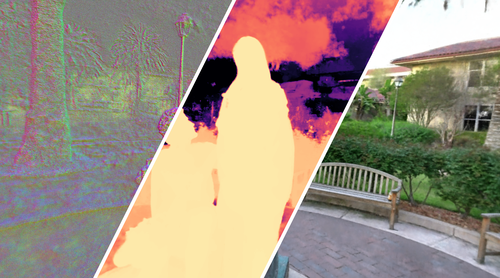}}} & 
        \tikz{
        \node[draw=black, line width=.3mm, inner sep=0pt] (ngp_smog) at (0,0) {
        \includegraphics[width=0.15\textwidth]{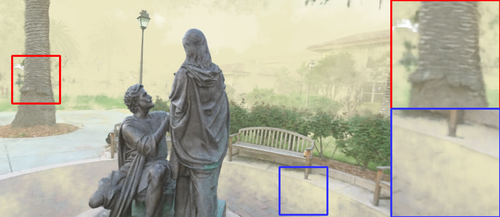}}} & 
        \tikz{
        \node[draw=black, line width=.3mm, inner sep=0pt] (ngp_flood) at (0,0) {
        \includegraphics[width=0.15\textwidth]{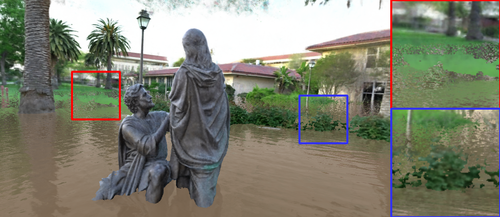}}} & 
        \tikz{
        \node[draw=black, line width=.3mm, inner sep=0pt] (ngp_snow) at (0,0) {
        \includegraphics[width=0.15\textwidth]{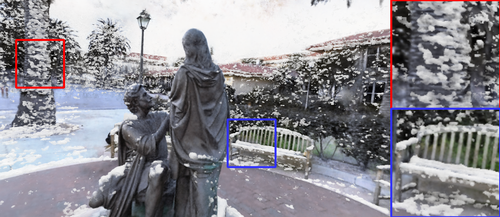}}} \\
        
        \raisebox{5mm}[0pt][0pt]{\rotatebox[origin=c]{90}{\tiny{Ours}}} & 
        \tikz{
        \node[draw=black, line width=.3mm, inner sep=0pt] (ours) at (0,0) {
        \includegraphics[width=0.118\textwidth]{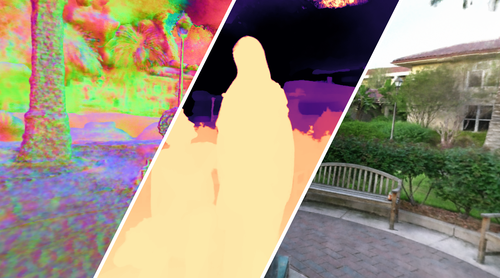}}} & 
        \tikz{
        \node[draw=black, line width=.3mm, inner sep=0pt] (ours_snow) at (0,0) {
        \includegraphics[width=0.15\textwidth]{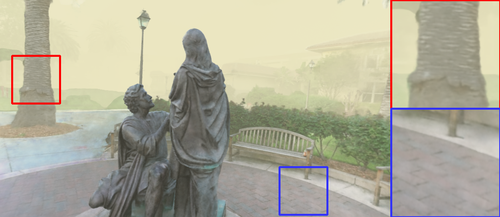}}} & 
        \tikz{
        \node[draw=black, line width=.3mm, inner sep=0pt] (ours_snow) at (0,0) {
        \includegraphics[width=0.15\textwidth]{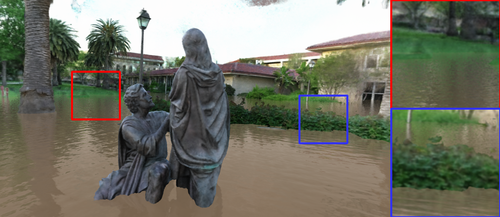}}} & 
        \tikz{
        \node[draw=black, line width=.3mm, inner sep=0pt] (ours_snow) at (0,0) {
        \includegraphics[width=0.15\textwidth]{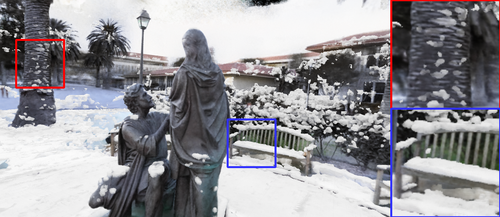}}} \\
       & \tiny{(a) Normal/Depth/RGB} & \tiny{(b) Smog} & \tiny{(c) Flood} & \tiny{(d) Snow} \\
    \end{tabular}
  }
  \vspace{-3mm}
   \captionof{figure}{\textbf{Climate simulations reveal NeRF modeling errors.} Top: NGP, bottom: our NeRF model. On the left, there are normal, depth map, and RGB images. Although the rendered RGB images of the two NeRF versions appear similar, NGP exhibits depth errors in the sky regions and more normal noises on surfaces. This results in geometric inaccuracies in fog and flood images, as well as unrealistic snow accumulation. } 
   \label{fig:geometry_improvement}
   \vspace{-3mm}
\end{table*}

%% file: 3_method_v2.tex
\section{Method}
\label{sec:method}

ClimateNeRF fuses physical simulations with NeRF models of scenes to produce realistic videos of climate change
effects. A simple example illustrates how components interact in our approach.  
Imagine we wish to build a model of a flooded scene in the Fall.  
We acquire images, apply a Fall style (Section~\ref{sec:stylization}), and build a NeRF from the
results (Section~\ref{sec:modeling}).  
We then use geometric information in that NeRF to compute a water surface.  
This is represented using a density field, a color field together with normal and BRDF representations
(Section~\ref{sec:simulation}). 
Finally, to render, we query the model with rays.  The details are elaborate
(Section~\ref{sec:implementation}), but the general idea is straightforward:  we edit the NeRF's density
and color functions to represent effects like flood; and we intercept rays to represent specular effects.
So if a ray encounters high density in the NeRF first, we use the NeRF integral for that ray;  but if the first collision is with the water surface, we reflect
that ray in the water surface, then query the NeRF with the reflected ray.
Fig.~\ref{fig:3_overview} provides an overview of our approach.   

\subsection{Repurposing NeRF for Simulation}
\label{sec:modeling}

NeRF builds a parametric scene representation that supports realistic rendering from
multiple images of a scene obtained at known poses. The scene is represented by a field
$(\sigma, \mathbf{c}) = F_{\theta}(\mathbf{x}, \mathbf{d})$,
which accepts position $\bx$ and direction $\bd$ and predicts density $\sigma \in \bbR$ and color $\mathbf{c} \in \bbR^3$. This function is encoded in a multi-layer perceptron (MLP) with learnable parameters
$\theta$.  
Rendering is done by querying radiance along appropriate choices of ray,
computed as a volume integral~\cite{kajiya1984ray}.  This integral is estimated
by drawing samples along the ray, evaluating the density and color of those samples,
then accumulating values.   The rendering process is differentiable so that NeRF can be trained by minimizing the image 
reconstruction loss over training views through gradient descent: $\min_\theta \sum_{\br}\|C(\mathbf{r}) -
C_{gt}(\mathbf{r})\|_2^2$.   

\input{figures/method/flood_render}

\paragraph{Repurposing NeRF for simulation:} 
Although there are numerous variants of NeRF, most are primarily focused on the task of novel-view synthesis~\cite{mildenhall2020nerf, barron2021mip, barron2022mip, muller2022instant, chen2022tensorf}. 
However, combining graphical simulation with these NeRF representations directly results in rendering artifacts, as shown in Fig.~\ref{fig:geometry_improvement}.
These artifacts arise because the NeRF model cannot capture the scene's accurate and clean geometry, resulting in floaters, and cannot complete the 4D light field, resulting in holes. 
When physical simulation, such as adding flood and snow into the scene, is incorporated, the artifacts are particularly notable since human perception is more sensitive to these unnatural phenomena.

A good NeRF model for simulation must meet three requirements: 
1) accuracy and completeness of geometry, which allows the scene to interact with physical objects; 
2) semantic awareness, which enables the interaction to reflect the scene's characteristics; and 
3) coherence and completeness of the light field, which enables the use of NeRF for realistic lighting of the inserted physical entities. 
To this end, we systematically study multiple variants of techniques to improve NeRF. We use instant-NGP~\cite{muller2022instant} as our base representation to reconstruct the scene for its efficiency and flexibility. This NeRF alternative explicitly stores multi-resolution features %
for scene representation, making it suitable as we can edit local features easily. 

A physical simulation needs access to the surface normal of any point to compute interactions with
snow and water, and it needs access to the point's semantics (in the sense of semantic segmentation) to
transfer style.  
We expand the NGP and allow it to output both semantic logic $\bs$ and surface normal $\bn$. 
There is no semantic or surface normal ground truth during training. 
We use an off-the-shelf pretrained monocular semantic
segmentation network~\cite{xie2021segformer} to produce semantic maps for each image.
We use density gradients
$\hat{\mathbf{n}}=-\frac{\nabla \sigma}{\|\nabla\sigma\|}$~\cite{boss2021nerd, srinivasan2021nerv}
(cf. Ref-NeRF~\cite{verbin2022refnerf}) to guide the predicted surface normals $\mathbf{n}$ with a weighted MSE loss.

To simulate (say) a blizzard, we must add snow to the scene and turn trees dark, but we should not change the shape of the
house.  
To keep spatial features intact at the stylization stage, we disentangle instant-NGP's latent feature
(as in~\cite{chen2022tensorf, sun2022direct}).   
For each voxel in the NGP model, we split the latent
feature into geometry features $\gamma_\text{geo}$ and appearance features $\gamma_\text{app}$. The geometry features
are trained to render density. The appearance features are used for rendering color, semantics, and normals.
We will freeze the geometry feature vector during the stylization stage and change only the appearance feature vector.  

\paragraph{Improving NeRF geometry to reduce simulation errors:}
The non-ideal geometry in NeRF, as seen in~\cite{mildenhall2020nerf, muller2022instant}, produces severe artifacts in simulation. To address this issue, we incorporate per-frame appearance embedding~\cite{martin2021nerf} and distortion loss~\cite{barron2022mip} to eliminate floaters. 
Additionally, we utilize a surface normal MLP to enhance the smoothness of the scene normals while introducing a novel sky loss function $\mathcal{L}_{\text{sky}}=\sum_{\br \in \text{sky}}e^{-d(\br)}$ to eliminate floaters and fill in holes in the sky. 
Through these improvements in geometry, we have significantly improved the realism of the simulation (more details can be found in the supplementary material).

Our final NeRF renders density $\sigma$, color $\bc$, semantic logit $\bs$ and normal $\bn$, given a query point
and ray direction:
\begin{equation}
\label{eq:ngp}
    (\sigma, \mathbf{c}, \mathbf{s}, \mathbf{n}) = F_{\theta}(\mathbf{x}, \mathbf{d}; \gamma_\text{geo} , \gamma_\text{app}).
\end{equation}
Table.~\ref{tab:geometry_techniques} compares the techniques used in NeRF variants and summarize what we find useful to improve the simulation. 
\input{tables/geometry_techniques}

\input{figures/method/snow_render}

\subsection{Stylization}
\label{sec:stylization}

Deciduous trees drop their leaves in winter, and physical simulation is not efficient for capturing effects like this.
We use FastPhotoStyle~\cite{li2018closed} to transfer style to rendered images from a pre-trained model $F_{\theta}$. 
We only transfer only to regions 'terrain', 'vegetation', or 'sky' regions
to mimic natural weather change phenomena.
The resulting images look realistic but are not necessarily view-consistent. Hence, a student instant-NGP model is fine-tuned to ensure the view consistency of the style-transferred scene.  This is trained to minimize the color difference between our student NeRF-rendered results and style-transferred images. %
We  keep the geometry intact and alter only  appearance to achieve this goal, so only the appearance feature code $\gamma_\text{app}$ is optimized during the style transfer stage: $\min_{\gamma_\text{app}} \sum_{\mathbf{r} \in \mathcal{R}}\left\|C(\mathbf{r})-C_\text{stylized}(\mathbf{r})\right\|_{2}^{2}$, where $C(\br)$ is renderd color and $C_\text{stylized}$ is the style-transferred in the same view. This gives us a new NGP model $(\sigma, \mathbf{c}') = F_{\theta}'(\mathbf{x}, \mathbf{d}, \ell^{(a)}_{i})$ which encodes the style. This optional style transfer step simulates composite effects, such as a flood in Fall,  where the original images were captured in Spring.

\subsection{Representing and Rendering Climate Effects}\label{sec:simulation}

We want to generate scenes with new physical entities -- snow, water, smog -- in place.  We must
determine where they are (the job of physical simulation) and what the resulting image looks like (the job of
rendering).   {\em How} the simulation represents results is important because results must be accessible to the
rendering process.  Rendering will always involve computing responses to ray queries, so computing
radiance at $\mathbf{u}$ in direction $\mathbf{v}$.
We must represent simulation results in terms of densities, and we must be able to compute normals and surface
reflectance properties.  
Generally, we write $O_{\phi}(\bx; F_{\theta}): \bbR^3 \rightarrow \bbR$ for a density
resulting from a physical simulation; $N_\phi(\bx; F_{\theta}): \bbR^3 \rightarrow \bbS^2$ for normals;
and $B_\phi(\bx, \bomega_o, \bomega_i; F_{\theta}): \bbR^9 \rightarrow \bbR$ for BRDF.  Each depends on the existing
scene $F_\theta$. Choice of $B_\phi$ can simulate various effects, including the atmospheric effect of smog, refraction
and reflections on water surfaces, and scattering of accumulated snow.  
$\{O_\phi, N_\phi, B_\phi\}$ differs drastically across different physical simulations (details per effect in Sect.~\ref{sec:implementation}).

Once the physical entities are defined by functions $O_{\phi}, N_\phi, B_\phi$, we can render them into
the image by modeling the light transport between the physical entities and the scene.    Given the query point position
$\bx$, the simulation estimates the density and color of physical entities at position $\bx$ through
physically based rendering: 
$\sigma_{\phi} = O_{\phi}(\mathbf{x}; F_{\theta})$, 
$\mathbf{c}_{\phi} = \int_{\Omega} L(\mathbf{x}, \bomega_i)B_{\phi}(\mathbf{x}, \mathbf{d}, \bomega_i; F_{\theta})(\bomega_i \cdot N_\phi(\bx; F_{\theta})) \text{d}\bomega_i$,
where entity color $\bc_{\phi}$ is approximated with physically-based rendering equation~\cite{kajiya1986rendering} with
normal $N_\phi$ and BRDF $B_\phi$.  Importantly, we approximate the incident illumination $L(\bx, \bomega_i)$ with
radiance by tracing a ray $\br(t) = \bx-t\bomega_i$ opposite to the incident direction in the learned NeRF, \ie $L(\bx,
\bomega_i) = C(\br)$.  %
Depending on the physical entities, we use analytical or sampling-based solutions for the integral. Multiple bounces can be simulated through sampling.   

We follow the volumetric rendering process defined in two passes. For every point along a camera ray, we query the opacity and color of the physical entities as described in Sec.~\ref{sec:simulation}. The system also retrieves the original density $\sigma_{\theta}$ and color $\bc_\theta$ through Eq.~\ref{eq:ngp}. \method~estimates final density and color of the simulated scene by:
$
     {\sigma}_\text{final} = \sigma_{\theta} + \sigma_{\phi}, \quad
     \bc_\text{final}= \frac{\sigma_{\theta}\mathbf{c}_{\theta} + \sigma_{\phi}\mathbf{c}_{\phi}}{\sigma_{\theta} + \sigma_{\phi}},
$
Once ${\sigma}_\text{final}, \bc_\text{final}$ are estimated for each ray points $\{\mathbf{x}_i\}$, the pixel color is
calculated by volume rendering.
Fig.~\ref{fig:3_rendering} depicts the entire physically
inspired simulation and rendering process.

\input{tables/4_quality_smog.tex}
\input{tables/4_quality_flood.tex}

\subsection{Climate Effect Details}
\label{sec:implementation}

\method~is able to simulate smog, flood, and snow through various choices of $\{O_{\phi}, N_\phi, B_\phi \}$.
\paragraph{Smog Simulation} We assume that smog is formed by tiny absorbing particles, uniformly distributed
in empty space.  
In empty space, the NeRF density $\sigma_{\theta}=0$.  The Beer-Lambert law (originally~\cite{beer,
  lambert}; in~\cite{fox}) means we can model smog density in free space by simply adding a non-negative constant
to the density.  
Inside high-density regions of the NeRF, adding the constant does not significantly change the
integral, so we have
$
  O_\phi(\bx; F_\theta)   = \sigma_{\text{smog}} 
$. 
where $\sigma_\phi$ decides the density of the smog. Smog particles have
a constant color $\bc_{\text{smog}}$.  Both $\bc_\text{smog}$ and $\sigma_{\text{smog}}$ are controllable
parameters. Fig.~\ref{tab:4_control} depicts the effects of various smog densities.  

\paragraph{Flood Simulation}

The water surface of the flooded scene is approximately a horizontal plane: $\bn_w(\bx - \bo_w) = 0$, where the gravity
direction normal $\bn_w$ is estimated with camera poses and vanishing points detection~\cite{lu20172}, and plane origin
$\bo_w=(0, 0, h)$ determines the water height. But there are water ripples, which we implement
following~\cite{horvath2015empirical} with Fast Fourier Transform (FFT) based ripples and waves. The
FFT wave takes random spectral coefficients as input and outputs a spatiotemporal surface normal based on wind speed,
direction, and spatial and temporal frequencies. As shown in Fig.~\ref{tab:4_quality_flood}, compared against still water,
FFT-based water surface simulation significantly improves the realism of the water surfaces. We simulate opacity and
micro-facet ripples that make the water look glossy (details in supplementary).
 To approximate the integral of the rendering equation, we adopt the sigma-point method~\cite{menegaz2015systematization, wan2000unscented} and sample 5 rays from $\bx$, including reflection direction
$\bd_r$ and nearby four rays. 
\input{figures/refraction}

\paragraph{Snow Simulation}
Snow is more likely to be accumulated on surfaces facing upward, and the deeper part of the snow is denser due to
gravity. We simulate density over %
object surfaces using metaballs~\cite{nishita1997modeling, keselman2022fuzzy} centered on surfaces and with density
$\sigma_o$ at the center. Realism is ensured by considering both the scene geometry and snowfall direction (\eg, areas such as under the table receiving less accumulated snow). 
The density distribution within a metaball can be formulated by kernel function $\mathbf{K}(r,
R, \sigma_o)$, which leads to a smooth decrease of density as the distance $r$ from the metaball's center grows.   
We follow~\cite{van2007point} and tuned it for better quality: 
$
    \mathbf{K}(r, R, \sigma)=\frac{315}{64\pi1.5^7}(1.5^2-(\frac{r}{R})^2)^3\sigma  
    $.

For any point $\mathbf{x}$ in the space, we calculate the snow's density of $\mathbf{x}$ using a parzen window density
estimator over $N$ local nearest neighbors.  The final density of the snow surface is decided accordingly to local geometry  (details in supplementary). 
We use a spatially-varying diffuse color $\mathbf{c}_\phi(\bx_i^{(p)})$ (which is close to pure white
multiplied by the average illumination of the scene) to approximate BRDF,
and apply a subsurface scattering effect~\cite{nishita1997modeling} to light the snow's shadowed part (further detail in
supplementary). 
{We re-cast shadows on snow by leveraging 2D shadow prediction~\cite{chen2020multi} and bake shadow map using NeRF geometry.}
Fig.~\ref{tab:snow_render} depicts the snow simulation process.

%% file: figures/method/flood_render.tex
\begin{table*}[]
    \centering
    \resizebox{\linewidth}{!}{
\setlength{\tabcolsep}{0.05em} %
    \begin{tabular}{ccccc}
    \footnotesize
        \includegraphics[width=0.22\linewidth, trim={0 7cm 0 0},clip]{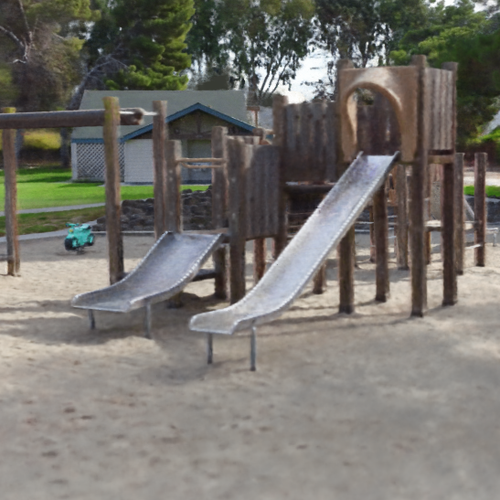} & 
        \includegraphics[width=0.22\linewidth, trim={0 7cm 0 0},clip]{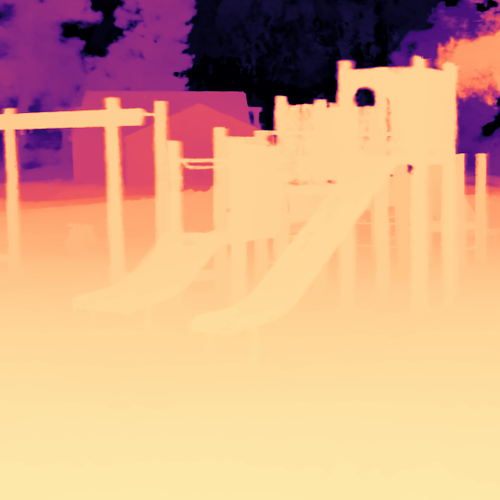}& 
        \includegraphics[width=0.22\linewidth, trim={0 7cm 0 0},clip]{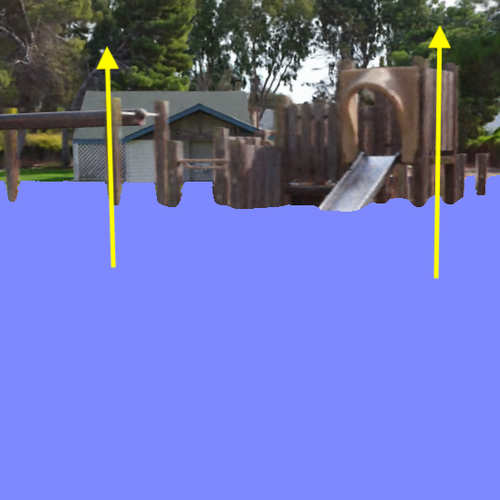}& 
        \includegraphics[width=0.22\linewidth, trim={0 7cm 0 0},clip]{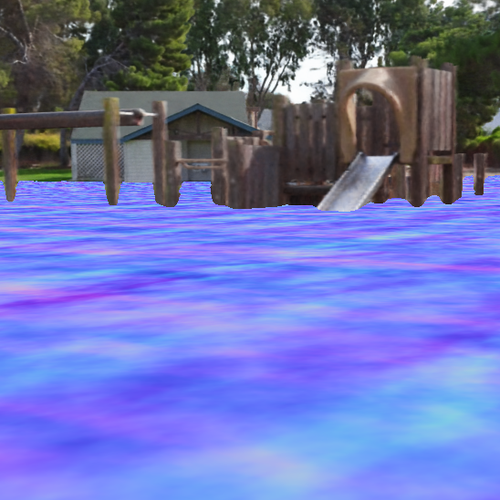} & 
        \includegraphics[width=0.22\linewidth, trim={0 7cm 0 0},clip]{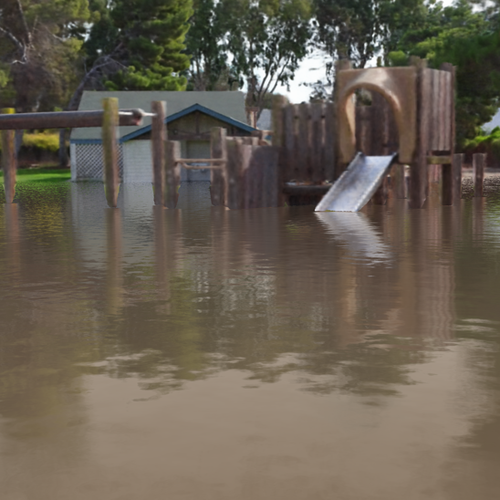}\\
        (a) Original NeRF & 
        (b) Depth map & 
        (c) Water surface & 
        (d) Normal map with wave & 
        (e) Final ClimateNeRF
    \end{tabular}
    }
    \vspace{-3mm}
\captionof{figure}{\textbf{Flooding simulation.} 
We first estimate the vanishing point direction based on the original image (a) and depth (b).
With the vertical vanishing direction (yellow arrows painted (c)), we can insert a planar water surface $\bn_w (x - \bo_w)$. 
We use FFT based water surface simulation to produce a spatiotemporal surface normal map in (d).
Our ClimateNeRF renders the scene with the simulated flood through ray tracing NeRF (e).
}
    \label{tab:flood_render}
    \vspace{-5mm}
\end{table*}

%% file: tables/geometry_techniques.tex
\begin{table}[t]
    \centering
    \resizebox{\linewidth}{!}{
    \begin{tabular}{lccccc}
        \hline 
        Method & \makecell[c]{Appearance \\ Embed.~\cite{martin2021nerf}} & \makecell[c]{Distortion \\ Loss~\cite{barron2022mip}} & \makecell[c]{Predicted \\ Normal} & \makecell[c]{Predicted \\ Semantics~\cite{Zhi:etal:ICCV2021}} & \makecell[c]{Sky \\ Loss} \\
         \hline 
        NGP~\cite{muller2022instant} & & & & & \\
        Mip-NeRF360~\cite{barron2022mip} & \ding{51} & \ding{51} & & & \\
        NeRFacto~\cite{tancik2023nerfstudio} & \ding{51} & \ding{51} & \ding{51} & & \\
        Ours & \ding{51} & \ding{51} & \ding{51} & \ding{51} & \ding{51} \\
         \hline 
    \end{tabular}
    }
    \vspace{-3mm}
    \caption{\textbf{Improving NeRF for simulation.}}
    \label{tab:geometry_techniques}
    \vspace{-6mm}
\end{table}

%% file: figures/method/snow_render.tex
\begin{table*}[]
    \centering
    \resizebox{\linewidth}{!}{
\setlength{\tabcolsep}{0.05em} %
    \begin{tabular}{ccccc}
    \footnotesize
        \includegraphics[width=0.22\linewidth]{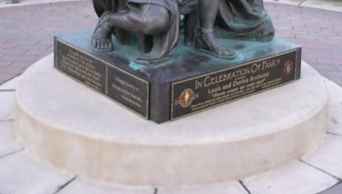} & 
        \includegraphics[width=0.22\linewidth]{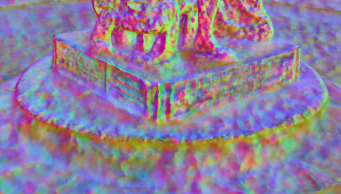}& 
        \includegraphics[width=0.22\linewidth]{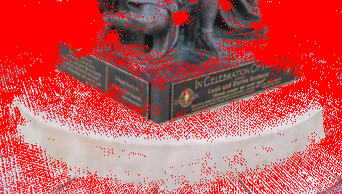}& 
        \includegraphics[width=0.22\linewidth]{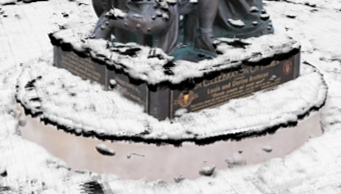} & 
        \includegraphics[width=0.22\linewidth]{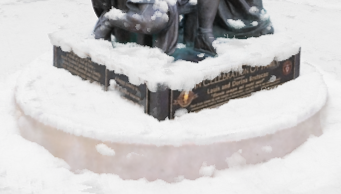}\\
        (a) Original NeRF & 
        (b) Surface normal & 
        (c) Metaball centers (red) & 
        (d) Snow with diffuse model & 
        (e) Snow with scattering
    \end{tabular}
    }
    \vspace{-3mm}
\captionof{figure}{\textbf{Snow simulation.}  
We first locate metaballs on object surfaces facing upward based on surface normal values (b). 
With metaballs, we can estimate the density $\sigma_{snow}(\mathbf{x})$ and color $\mathbf{c}(\mathbf{x})$ with a parzen window density estimator. 
(d) and (e) show the differences between fully diffuse model and scattering approximations, shadowed parts in (d) are lit in (e).}
    \label{tab:snow_render}
    \vspace{-5mm}
\end{table*}

%% file: tables/4_quality_smog.tex
\begin{table}[t]
    \centering
    \small
    \setlength\tabcolsep{0.05em} %
    \resizebox{\linewidth}{!}{
    \begin{tabular}{ccc}

    \tikz{
        \node[draw=black, line width=.3mm, inner sep=0pt] (family) at (0,0) {
        \includegraphics[width=0.33\linewidth]{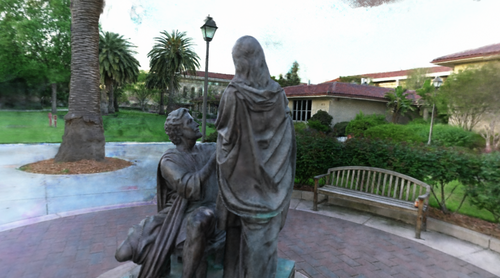}}}
    & 
    \tikz{
        \node[draw=black, line width=.3mm, inner sep=0pt] (family_smog_gan) at (0,0) {
        \includegraphics[width=0.33\linewidth]{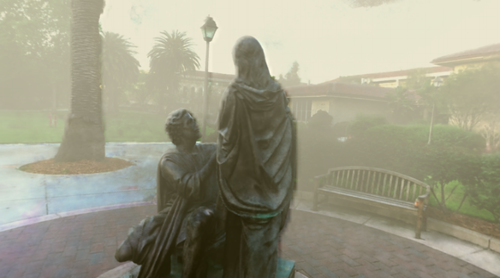}}}
    & 
    \tikz{
        \node[draw=black, line width=.3mm, inner sep=0pt] (family_smog_ours) at (0,0) {
        \includegraphics[width=0.33\linewidth]{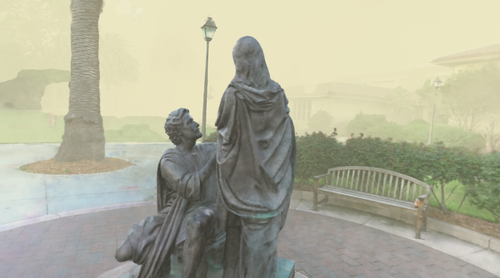}}} \\

    \tikz{
        \node[draw=black, line width=.3mm, inner sep=0pt] (horse) at (0,0) {
        \includegraphics[width=0.33\linewidth]{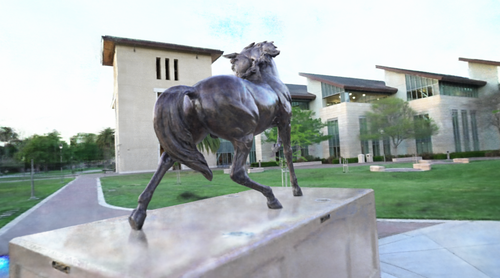}}}
    & 
    \tikz{
        \node[draw=black, line width=.3mm, inner sep=0pt] (horse_smog_gan) at (0,0) {
        \includegraphics[width=0.33\linewidth]{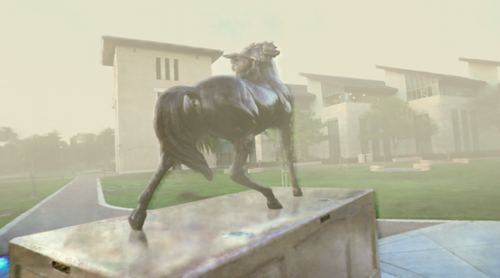}}}
    & 
    \tikz{
        \node[draw=black, line width=.3mm, inner sep=0pt] (horse_smog_ours) at (0,0) {
        \includegraphics[width=0.33\linewidth]{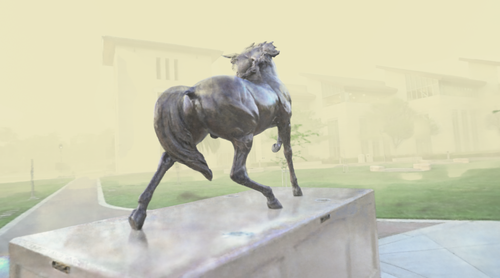}}} \\
    
     Original & ClimateGAN~\cite{schmidt2021climategan} & Ours \\

    \end{tabular}
    }
    \vspace{-3mm}
    \captionof{figure}{\textbf{Smog simulation comparison.} \method~simulate smog in view-consistent manner, and separate foreground objects from background better.}
    \label{tab:4_quality_smog}
    \vspace{-5mm}
\end{table}

%% file: tables/4_quality_flood.tex
\begin{table*}[t]
    \centering
    \footnotesize
    \setlength\tabcolsep{0.05em} %
    \resizebox{\textwidth}{!}{
    \begin{tabular}{cccc}

    \tikz{
        \node[draw=black, line width=.3mm, inner sep=0pt] (garden) at (0,0) {
    \includegraphics[width=0.23\textwidth]{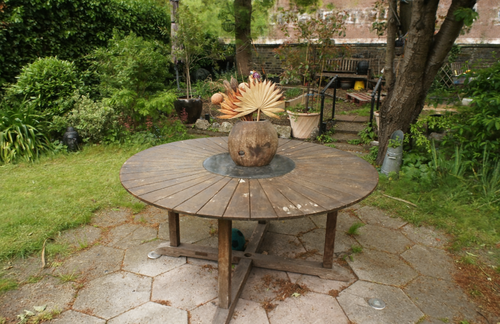} }}
    & 
    \tikz{
        \node[draw=black, line width=.3mm, inner sep=0pt] (garden_gan) at (0,0) {
    \includegraphics[width=0.23\textwidth]{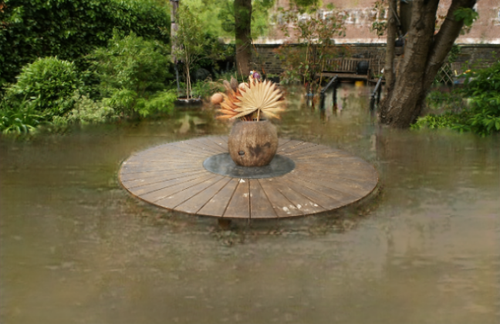} }}
    & 
    \tikz{
        \node[draw=black, line width=.3mm, inner sep=0pt] (garden_diffusion) at (0,0) {
    \includegraphics[width=0.23\textwidth]{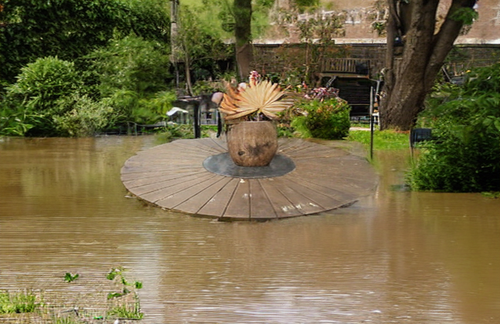} }}
    & 
    \tikz{
        \node[draw=black, line width=.3mm, inner sep=0pt] (garden_ours) at (0,0) {
    \includegraphics[width=0.23\textwidth]{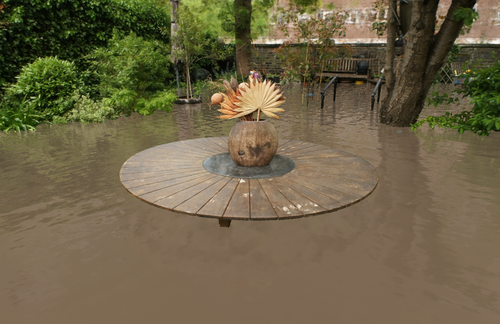} }}
    \\
    
     Original & ClimateGAN++~\cite{schmidt2021climategan} & Stable Diffusion~\cite{rombach2021highresolution} & Ours \\

    \end{tabular}
    }
    \vspace{-3mm}
    \captionof{figure}{\textbf{Flood simulation comparison.} 
     {Both ClimateGAN++~\cite{schmidt2021climategan} and Stable Diffusion~\cite{rombach2021highresolution} inpaint flood using water masks from our reconstructions, but suffer from inaccurate reflections or randomly generated contents. \method~simulates photorealistic reflections and ripples.}
     }
    \label{tab:4_quality_flood}
    \vspace{-5mm}
\end{table*}

%% file: figures/refraction.tex
\begin{figure}[t]
  \captionsetup[subfigure]{labelformat=empty}
  \centering

  \begin{subfigure}{0.4\linewidth}
    \centering
    
    \tikz{
        \node[draw=black, line width=.3mm, inner sep=0pt] (clear) at (0,0) {
    \includegraphics[width=\textwidth]{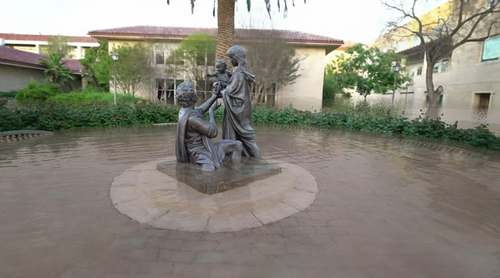} }}
    \vspace{-5mm}
    \caption{(a) Clear water}
    \vspace{-5mm}
    \label{fig:clear}
  \end{subfigure}
  \begin{subfigure}{0.4\linewidth}
    \centering
    \tikz{
        \node[draw=black, line width=.3mm, inner sep=0pt] (muddy) at (0,0) {
    \includegraphics[width=\textwidth]{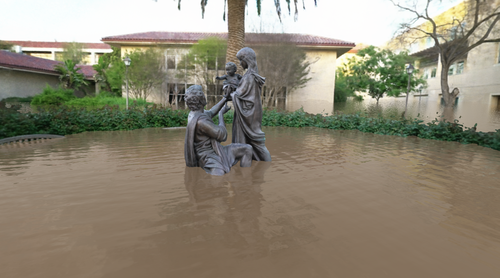} }}
    \vspace{-5mm}
    \caption{(b) Muddy water}
    \vspace{-5mm}
    \label{fig:muddy}
  \end{subfigure}
 \caption{\textbf{Refraction in flood simulation} 
 \vspace{-6mm}
}
  \label{fig:refraction}
\end{figure}

%% file: 4_experiment.tex
\input{tables/4_quality_snow.tex}

\section{Experiments}

\input{tables/4_user_study.tex}

We evaluate ClimateNeRF and show simulated results over various scenes across different climate effects. We compare our results with state-of-the-art 2D synthesis and 3D stylization to show the quality and consistency of rendered frames. Experimental results demonstrate that our method is more realistic and faithful than existing 2D synthesis and NeRF model finetuned on stylized images while we also maintain temporal consistency and physical plausibility. We encourage readers to watch supplementary videos for a better demonstration of our method's quality. 

\subsection{Experimental Details}

\paragraph{Datasets.}
We conduct experiments on various outdoor scenes: \textit{Playground}, {\it Family}, {\it Horse}, {\it Truck}, and {\it Train} from \textit{Tanks and Temples} dataset~\cite{knapitsch2017tanks}, {\it Garden} from {\it Mip-NeRF360}~\cite{barron2022mip} and {\it Seq 00} from {\it KITTI-360}~\cite{liao2022kitti}. The scenes vary significantly in contents, layouts, and viewpoints. 

\paragraph{Baselines.} 
We compare \method~ with the state-of-the-art 2D image editing methods, such as stable diffusion inpainting~\cite{rombach2021highresolution}, ClimateGAN~\cite{schmidt2021climategan}, as well as state-of-the-art 3D NeRF stylization~\cite{zhang2022arf}. For all 2D synthesis approaches, we first build a NeRF using NGP, render at the target view, and conduct synthesis. For all the 3D methods, we re-use the improved version of NGP-based NeRF. \textbf{ClimateGAN}~\cite{schmidt2021climategan} uses monocular depth to predict masks and uses GAN to inpaint the climate-related effects, including smog and flood; \textbf{ClimateGAN++} is an improved version for flood simulation using our method's water mask, yielding better geometry consistency; \textbf{Swapping Autoencoder}~\cite{park2020swapping} is a photorealistic 2D style transfer method. We use the model pre-trained on Flickr Mountains dataset and Flickr Waterfall dataset~\cite{park2020swapping} for snow. 
\textbf{Stable Diffusion}~\cite{rombach2021highresolution} is the state-of-the-art guided image inpainting method based on latent diffusion model. We feed accurate water masks produced by \method~ and use text prompts of "flooding" for inpainting.  \textbf{3D Stylization} leverages FastPhotoStyle~\cite{li2018closed}. To simulate white snow coverings, we stylize regions labeled {\it road}, {\it terrain}, {\it vegetation}, and {\it sky} while keeping the geometry. Please refer to supplementary materials for additional implementation details for all competing methods including ours. 

\subsection{Experimental Results}

\paragraph{Qualitative Results}

Fig.~\ref{tab:4_quality_smog} depicts qualitative results from smog simulation. Our method delivers better realism and physical plausibility (see the different transmission levels across foreground and background). ClimateGAN~\cite{schmidt2021climategan} generates visually reasonable results but fails to provide sharp boundaries. 
Additionally, video results further show our method is better at view consistency. 

We also report flood simulation results in Fig.~\ref{tab:4_quality_flood}. ClimateGAN++~\cite{schmidt2021climategan} produces waters with wrong reflection and blurry artifacts. Stable Diffusion~\cite{rombach2021highresolution} provides realistic and diverse colors and reflectance but hallucinates additional contents (e.g., cars, trees) and lacks consistency.  \method~renders realistic reflections with Fresnel effects and water ripples. Videos show that \method~ is view-consistent and provides fluid dynamics. {Fig.~\ref{fig:refraction} shows that our method simulates both clear and muddy water.}

We report snow simulation results in Fig.~\ref{tab:4_quality_snow}. As the figure shows, 3D Stylization changes the floor texture but cannot add physical entities to the scene, limiting its realism. Swapping Autoencoder~\cite{park2020swapping} changes the overall appearance but hallucinates unrealistic textures (e.g., car texture). On the other hand, \method~simulates photorealistic winter effects, including accumulated snow and change of sky and tree colors, etc. ClimateNeRF even piles snow on tiny structures like pedals, as shown in the figure. 
\input{tables/4_control.tex}

{
Using our proposed NeRF significantly enhances the quality of climate simulation. As demonstrated in Fig.~\ref{fig:geometry_improvement}, comparing to Instant-NGP, our improved geometry results in a more uniform smog density that is aligned with depth, and a more natural intersection between the border of the water surface and the scene. In addition, the appearance of white floaters is reduced and the ground is more evenly covered by snow in the snow scene.}

\paragraph{User Study}
We perform a user study to validate our approach quantitatively. Users are asked to watch pairs of synthesized images or videos of the same scene and pick the one with higher realism. 37 users participated in the study, and in total, we collected 2664 pairs of comparisons. Results are reported in Fig.~\ref{tab:user}. ClimateNeRF has consistently been favored among all video simulation comparisons thanks to its high realism and view consistency.  Single image comparison does not consider view consistency. In this case, ClimateNeRF still outperforms most baselines except diffusion models on flooding, which also produces realistic water reflectances. Users find diffusion models tend to produce more reflective water surfaces and diverse ripples. 

\paragraph{Controllability.}
One unique advantage of ClimateNeRF is its controllability. 
Fig.~\ref{tab:4_control} depicts the changes 
in smog density, water height, and snow thickness. 
We show additional controllability results, such as ripple size, flood color, water reflectance, and smog color in supp materials.

\input{tables/4_kitti360.tex}

\paragraph{Adverse weather simulation for self-driving.}
\method~ can be applied to any NeRF scene. In Fig.~\ref{tab:4_kitti360}, \method~ simulates climate effects in driving scenes of KITTI-360~\cite{liao2022kitti}, demonstrating the potential of ClimateNeRF to train and test self-driving under adverse weather. 

\paragraph{Limitations}
The simulation performance is closely tied to the quality of NeRF. Despite our proposed improvement, there might exist challenging cases leading to simulation artifacts. Additionally \method~is not real-time currently, and thus acceleration is an area for future work.

%% file: tables/4_quality_snow.tex
\begin{table*}[t]
    \centering
    \footnotesize
    \setlength\tabcolsep{0.05em} %
    \resizebox{\textwidth}{!}{
    \begin{tabular}{cccc}

    \tikz{
        \node[draw=black, line width=.3mm, inner sep=0pt] (truck) at (0,0) {
    \includegraphics[width=0.24\textwidth]{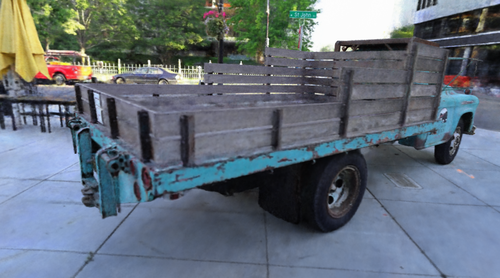} }}
    & 
    \tikz{
        \node[draw=black, line width=.3mm, inner sep=0pt] (truck_swap) at (0,0) {
    \includegraphics[width=0.24\textwidth]{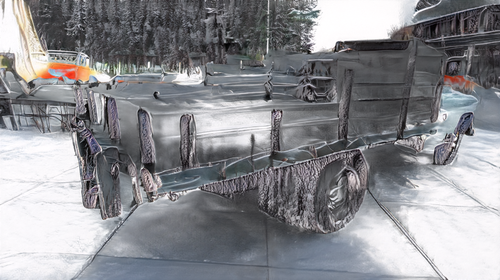} }}
    & 
    \tikz{
        \node[draw=black, line width=.3mm, inner sep=0pt] (truck_sty) at (0,0) {
    \includegraphics[width=0.24\textwidth]{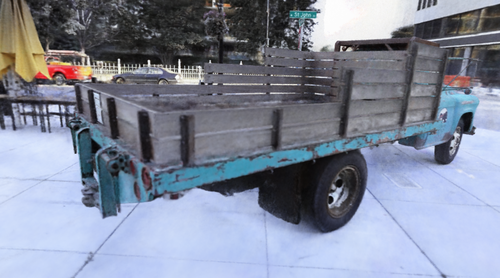} }}
    & 
    \tikz{
        \node[draw=black, line width=.3mm, inner sep=0pt] (truck_ours) at (0,0) {
    \includegraphics[width=0.24\textwidth]{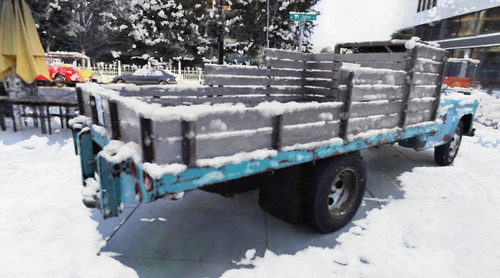} }}
    \\

     Original & Swapping Autoencoder~\cite{park2020swapping} & 3D Stylization & Ours \\

    \end{tabular}
    }
    \vspace{-3mm}
    \captionof{figure}{\textbf{Snow simulation comparison.}
    Swapping-Autoencoder~\cite{park2020swapping} captures appearance changes but ignores the geometry of both truck and train. 
    3D Stylization preserves the geometry of original scene well but doesn't accumulate snow on horizontal surfaces. In contrast, \method~has convincing snow accumulation both on ground and on objects.  Note small snow accumulations on the bogies and running board on the train and the boards and bonnet of the truck.}
    \label{tab:4_quality_snow}
    \vspace{-5mm}
\end{table*}

%% file: tables/4_user_study.tex
\begin{table}[t]
    \centering
    \setlength\tabcolsep{1.0pt} %
        \footnotesize
    \resizebox{\linewidth}{!}{
    \begin{tabular}{ccl}
      & Images & \shiftright{30pt}{Videos}  \\
      \raisebox{1.5\normalbaselineskip}[0pt][0pt]{\rotatebox[origin=c]{0}{Smog}} &\includegraphics[width=0.38\linewidth]{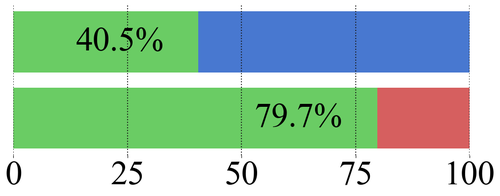} & \includegraphics[width=0.51\linewidth]{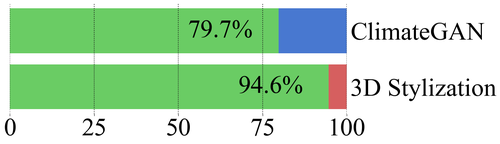} \\
      \hline 

      \raisebox{2.25\normalbaselineskip}[0pt][0pt]{\rotatebox[origin=c]{0}{Flood}} 
      & \includegraphics[width=0.38\linewidth]{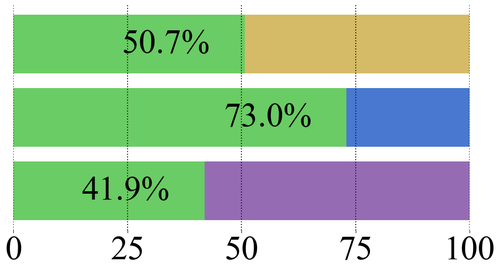} & \includegraphics[width=0.53\linewidth]{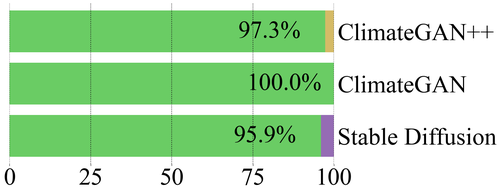}  \\
      \hline 
    
      \raisebox{1.5\normalbaselineskip}[0pt][0pt]{\rotatebox[origin=c]{0}{Snow}} &\includegraphics[width=0.38\linewidth]{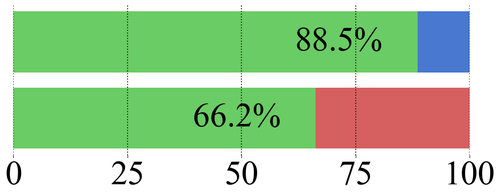} & \includegraphics[width=0.51\linewidth]{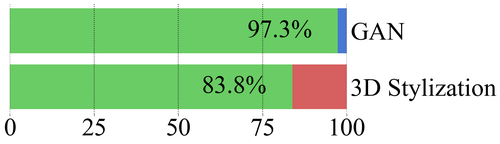} 
      \\
    \end{tabular}
    }
    \vspace{-2mm}
    \captionof{figure}{\textbf{User Study.} The bar length indicates the \% of users voting for higher realism. The video quality of ClimateNeRF significantly outperforms all baselines. 
    }
    \label{tab:user}
    \vspace{-3mm}
\end{table}

%% file: tables/4_control.tex
\begin{table}[t]
    \centering
    \setlength\tabcolsep{0.5pt} %
    \footnotesize
    \begin{tabular}{ccc}

    \tikz{
        \node[draw=black, line width=.3mm, inner sep=0pt] (smog_0) at (0,0) {
     \includegraphics[width=0.16\textwidth]{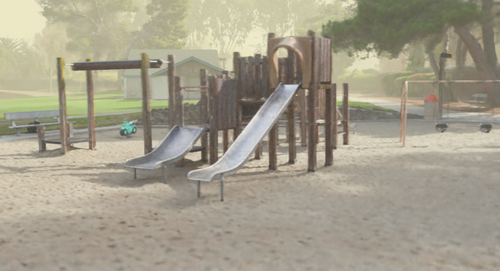} }}
     & 
     \tikz{
        \node[draw=black, line width=.3mm, inner sep=0pt] (smog_1) at (0,0) {
     \includegraphics[width=0.16\textwidth]{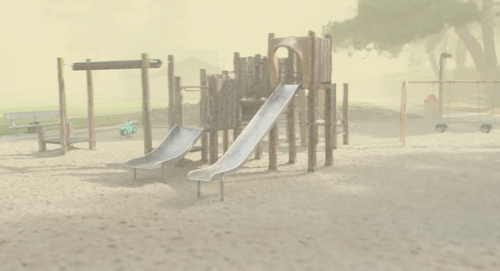} }}
     & 
     \tikz{
        \node[draw=black, line width=.3mm, inner sep=0pt] (smog_2) at (0,0) {
     \includegraphics[width=0.16\textwidth]{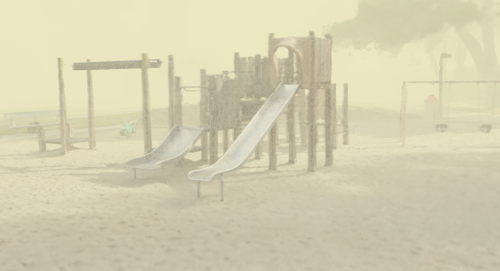} }} \\

     \tikz{
        \node[draw=black, line width=.3mm, inner sep=0pt] (flood_0) at (0,0) {
     \includegraphics[width=0.16\textwidth]{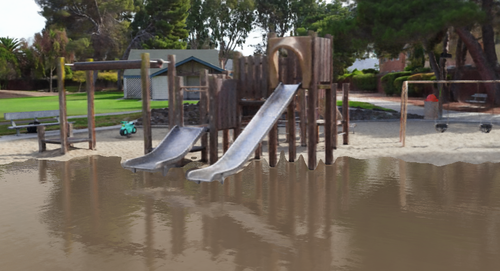} }}
     & 
     \tikz{
        \node[draw=black, line width=.3mm, inner sep=0pt] (flood_1) at (0,0) {
     \includegraphics[width=0.16\textwidth]{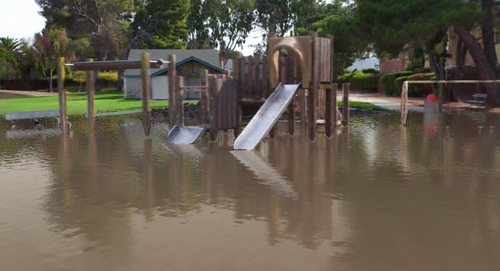} }}
     & 
     \tikz{
        \node[draw=black, line width=.3mm, inner sep=0pt] (flood_2) at (0,0) {
     \includegraphics[width=0.16\textwidth]{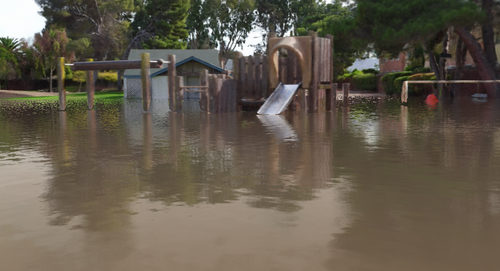} }} \\

     \tikz{
        \node[draw=black, line width=.3mm, inner sep=0pt] (snow_0) at (0,0) {
     \includegraphics[width=0.16\textwidth]{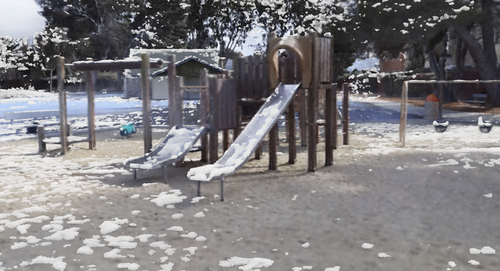} }}
     & 
     \tikz{
        \node[draw=black, line width=.3mm, inner sep=0pt] (snow_1) at (0,0) {
     \includegraphics[width=0.16\textwidth]{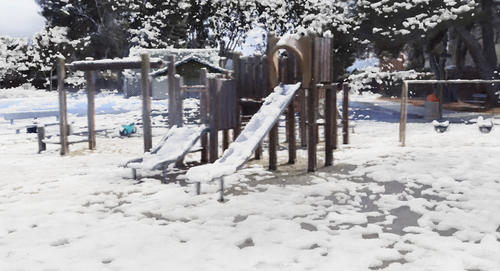} }}
     & 
     \tikz{
        \node[draw=black, line width=.3mm, inner sep=0pt] (snow_2) at (0,0) {
     \includegraphics[width=0.16\textwidth]{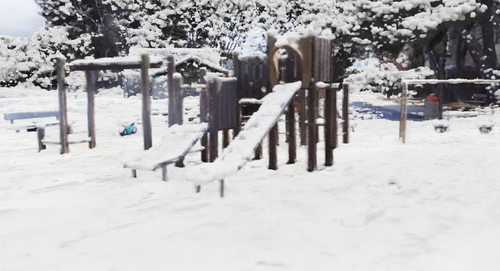} }} \\
     
    \end{tabular}
    \vspace{-3mm}
    \captionof{figure}{\textbf{Controllable Simulation}. 
    ClimateNeRF is highly controllable by users. \textbf{smog density} (1st row); \textbf{flood levels} (2nd row); \textbf{snow accumulations} (3rd row). %
    }
    \label{tab:4_control}
    \vspace{-6mm}
\end{table}

%% file: tables/4_kitti360.tex
\begin{table}[t]
    \centering
    \setlength\tabcolsep{0.05em} %
    
    \begin{tabular}{cc}
    
    \tikz{
        \node[draw=black, line width=.3mm, inner sep=0pt] (kitti_0) at (0,0) {
    \includegraphics[width=0.49\linewidth]{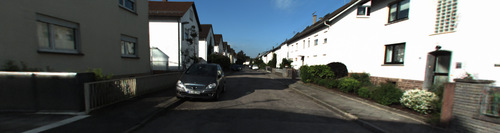} }} & 
    \tikz{
        \node[draw=black, line width=.3mm, inner sep=0pt] (kitti_1) at (0,0) {
    \includegraphics[width=0.49\linewidth]{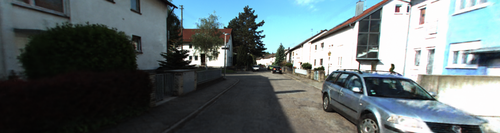} }} \\
    \tikz{
        \node[draw=black, line width=.3mm, inner sep=0pt] (kitti_0_smog) at (0,0) {
    \includegraphics[width=0.49\linewidth]{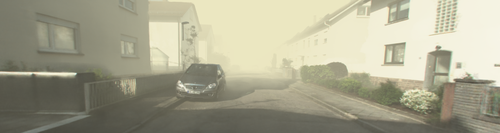} }} & 
    \tikz{
        \node[draw=black, line width=.3mm, inner sep=0pt] (kitti_1_smog) at (0,0) {
    \includegraphics[width=0.49\linewidth]{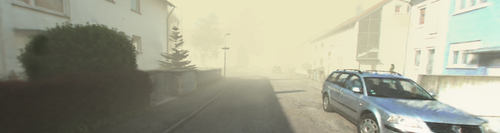} }} \\
    \tikz{
        \node[draw=black, line width=.3mm, inner sep=0pt] (kitti_0_flood) at (0,0) {
    \includegraphics[width=0.49\linewidth]{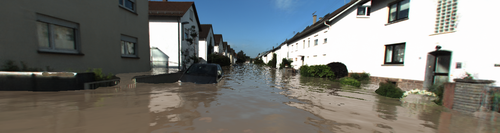} }} & 
    \tikz{
        \node[draw=black, line width=.3mm, inner sep=0pt] (kitti_1_flood) at (0,0) {
    \includegraphics[width=0.49\linewidth]{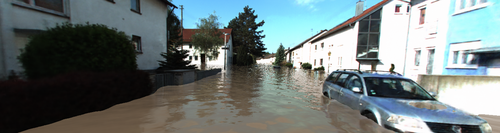} }} \\
    \tikz{
        \node[draw=black, line width=.3mm, inner sep=0pt] (kitti_0_snow) at (0,0) {
    \includegraphics[width=0.49\linewidth]{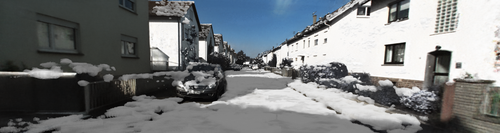} }} & 
    \tikz{
        \node[draw=black, line width=.3mm, inner sep=0pt] (kitti_1_snow) at (0,0) {
    \includegraphics[width=0.49\linewidth]{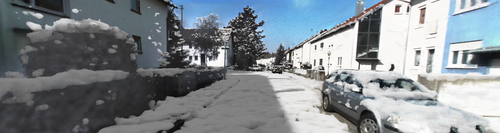} }} \\
    
    \end{tabular}
    \vspace{-3mm}
    \captionof{figure}{\textbf{Simulation on Driving Scenes~\cite{liao2022kitti}.} 
    }
    \vspace{-6mm}
    \label{tab:4_kitti360}
\end{table}

%% file: 5_conclusion.tex
\section{Conclusion}

We propose a novel NeRF editing framework that applies physical simulation to NeRF models of scenes. Leveraging this framework, we build \method, allowing us to render realistic climate change effects, including smog, flood, and snow. Our synthesized videos are realistic, view-consistent, physically plausible, and highly controllable. 
We demonstrate the potential of \method~to help raise climate change awareness in the community and enhance self-driving robustness to adverse weather conditions.

%% file: supp/abstract.tex
In this supplementary material, we describe implementation details of the simulation framework in Section~\ref{sect:implementation} and provide an ablation study to validate our design choices in Section~\ref{sect:ablation}. 
\method~is highly controllable and is demonstrated in Sect~\ref{sect:control}.
We present extensive qualitative and quantitative results in Section~\ref{sect:qualitative} and Section~\ref{sect:quantitative}, respectively.

\input{figures/sydney}
\input{figures/rising}

%% file: figures/sydney.tex
\begin{table*}[h]
    \centering
    \setlength\tabcolsep{0.05em}
    \resizebox{\textwidth}{!}{
    \begin{tabular}{cccc}
        \includegraphics[width=0.24\textwidth]{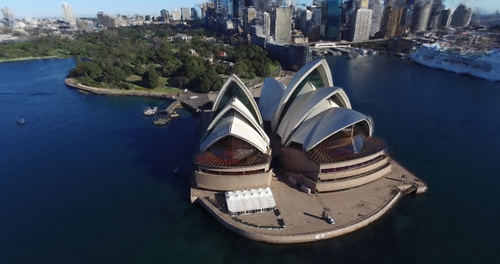} & \includegraphics[width=0.24\textwidth]{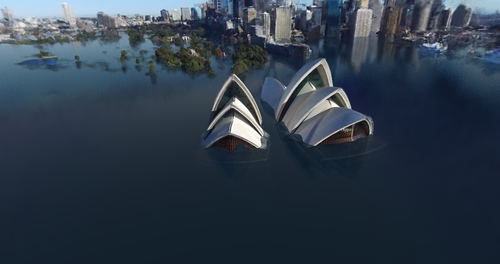} & \includegraphics[width=0.24\textwidth]{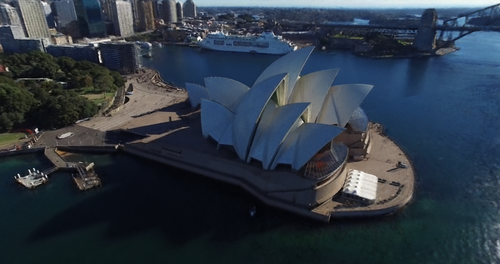} & \includegraphics[width=0.24\textwidth]{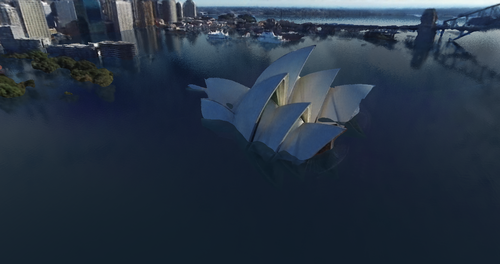} \\
        Reconstruction 1 & Simulation 1 & Reconstruction 2 & Simulation 2
    \end{tabular}
    }
    
    \captionof{figure}{\textbf{Rising Sea-level Simulation.} We simulate rising sea level at the Sydney Opera House from various view points.}
    \vspace{-3mm}
    \label{fig:sydney}
\end{table*}

%% file: figures/rising.tex
\begin{table*}[h]
    \centering
    \setlength\tabcolsep{0.05em}
    \resizebox{\textwidth}{!}{
    \begin{tabular}{lcccc}
        \raisebox{12mm}[0pt][0pt]{\rotatebox[origin=c]{90}{{\footnotesize{Reconstruction}}}} & \includegraphics[width=0.24\textwidth]{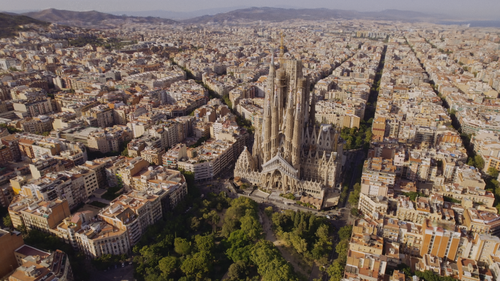} & \includegraphics[width=0.24\textwidth]{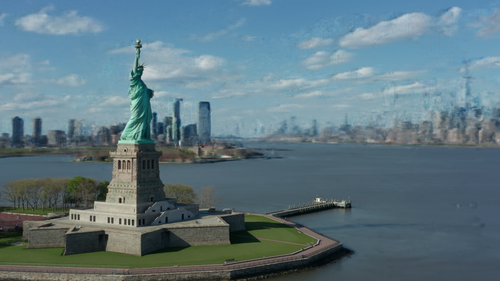} & \includegraphics[width=0.24\textwidth]{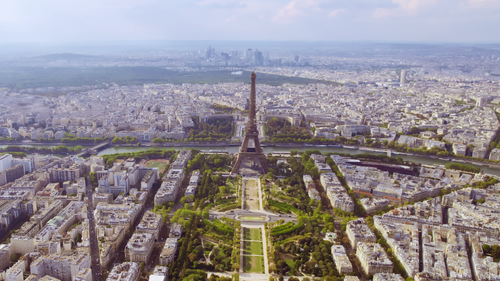} & \includegraphics[width=0.24\textwidth]{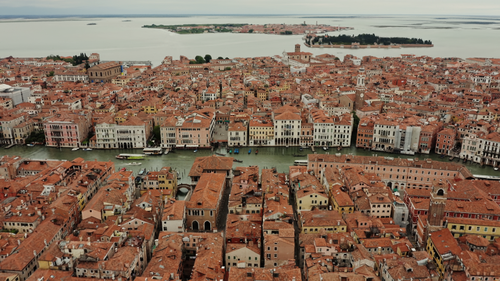} \\

        \raisebox{12mm}[0pt][0pt]{\rotatebox[origin=c]{90}{{\footnotesize{Simulation}}}} & \includegraphics[width=0.24\textwidth]{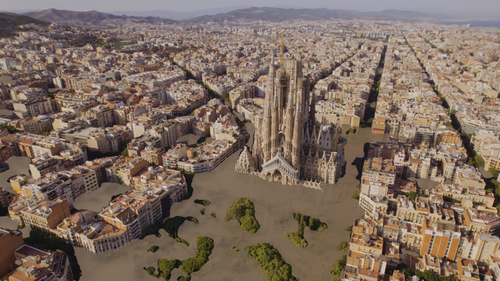} & \includegraphics[width=0.24\textwidth]{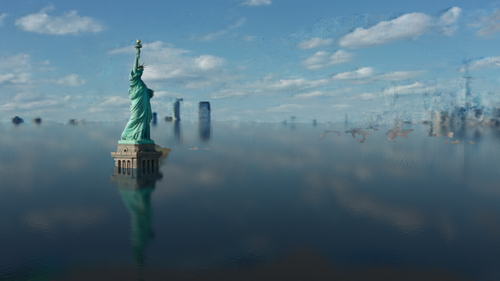} & \includegraphics[width=0.24\textwidth]{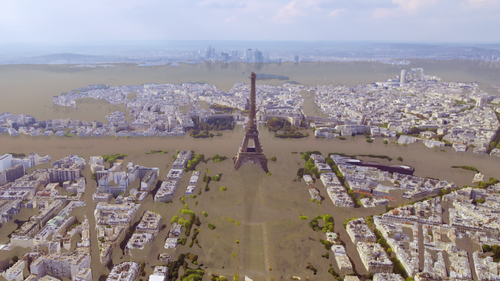} & \includegraphics[width=0.24\textwidth]{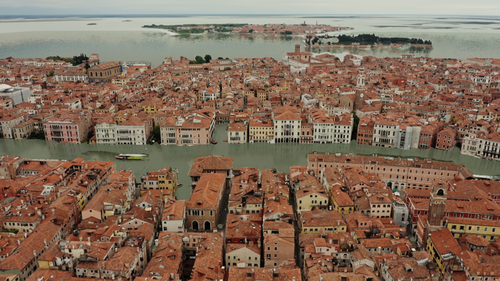} \\
        & La Sagrada Familia & Statue of Liberty & Eiffer Tower & Venice
    \end{tabular}
    }
    \captionof{figure}{\textbf{Rising Sea-level Simulation.} Visualizing the consequences of rising sea levels in world-famous scenes~\cite{lu2023large} can raise general public awareness about climate change.}
    \label{fig:rising}
\end{table*}

%% file: supp/implementation.tex
\section{Implementation Details}\label{sect:implementation}

\paragraph{3D Scene Reconstruction}

Our implementation is based on ~\cite{queianchen_ngp}. 
We use spatial hashing grid~\cite{muller2022instant} to represent the 3D scene. 
The entire space contains a multi-resolution feature grid $\{\mathrm{enc}(\mathbf{x};\theta^l)\}_{l=1}^{L}$, where $L$ is the total number of resolution levels; $\theta^l$ are learnable parameters at each level.
For a 3D point $\mathbf{x}\in R^3$, we index grids by a spatial hash function~\cite{teschner2003optimized} and fetch feature by interpolation and concatenation: $\gamma=\mathrm{cat}\{\mathrm{interp}(\mathbf{x},\mathrm{enc}(\mathbf{x};\theta^l))\}_{l=1}^{L}$.
To keep spatial features intact at the stylization stage and separate gradient flows between geometry and color, we develop a disentangled version of instant-NGP inspired by ~\cite{chen2022tensorf, sun2022direct}. 
Each voxel maintains a \emph{geometry code} $\gamma_{\sigma}$ and \emph{appearance code} $\gamma_{\text{app}}$. The geometry code $\gamma_{\sigma}$ encodes opacity and the appearance code $\gamma_{\text{app}}$ contains color, semantic and normal information:
 \begin{align}\label{eq:hash_enc}
    \gamma_{\sigma}=\mathrm{cat}\{\mathrm{interp}(\mathbf{x},\mathrm{enc}(\mathbf{x};\theta^l_{\sigma}))\}_{l=1}^{L_{\sigma}}, \quad
    \gamma_{\text{app}}=\mathrm{cat}\{\mathrm{interp}(\mathbf{x},\mathrm{enc}(\mathbf{x};\theta^l_{app}))\}_{l=1}^{L_{app}} \ ,
\end{align}
where $\mathrm{interp}$ stands for linear interpolation, $\mathrm{cat}$ denotes concatenation.

Inspired by ~\cite{muller2022instant}, we use shallow MLPs to predict densities $\sigma$, colors $\mathbf{c}$, semantic logits $\mathbf{s}$ and surface normal values $\mathbf{n}$ respectively from geometry features $\gamma_{\sigma}$ and appearance features $\gamma_{\text{app}}$. 
We also incorporate low-dimensional per-frame appearance embeddings $\{\ell^{(a)}_{i}\}^N_{i=1}$~\cite{martin2021nerf} to balance different lighting conditions across images. 
With hash grids and MLPs, we have the following function to reconstruct the original scene:
\begin{equation}\label{eq:full_model}
    (\sigma, \mathbf{c}, \mathbf{s}, \mathbf{n}) = F_{\theta}(\mathbf{x}, \mathbf{d}, \ell^{(a)}_{i}, \gamma_{\sigma}, \gamma_{\text{app}})
\end{equation}

Following volume rendering and alpha blending~\cite{mildenhall2020nerf}, we render the color $C(\mathbf{r})$ for ray $\mathbf{r}$ and its semantic logit $S(\mathbf{r})$~\cite{Zhi:etal:ICCV2021}.
\begin{align}\label{eq:nerf_volume}
    C(\mathbf{r}) = \sum_{i=1}^{N}T_i(1-\text{exp}(-\sigma_i\delta_i))\mathbf{c}_i; \quad
    S(\mathbf{r}) = \sum_{i=1}^{N}T_i(1-\text{exp}(-\sigma_i\delta_i))\mathbf{s}_i, \text{where } T_i = \text{exp} \left( -\sum_{j=1}^{i-1}\sigma_j\delta_j \right)
\end{align}
We apply softmax to semantic logits $S(\mathbf{r})$ to obtain semantic probabilities $\{p(\mathbf{r})^l\}_{l=1}^L$ for all labels.
During training, we perform MSE loss $L_C$ and cross-entropy loss $L_S$ for rendered colors and semantic logits using ground truth color $C_{gt}(\mathbf{r})$ and 2D semantic logits $\{\hat{p}(\mathbf{r})^l\}_{l=1}^L$ predicted by segformer~\cite{xie2021segformer} pretrained on cityscape dataset~\cite{Cordts2016Cityscapes}. 
Moreover, since we leverage pseudo-semantic labels, we detach densities $\sigma$ when rendering $S(\mathbf{r})$. 
Similar to Ref-NeRF~\cite{verbin2022refnerf}, we use the density gradient normals $\hat{\mathbf{n}}=-\frac{\nabla \sigma}{\|\nabla\sigma\|}$~\cite{boss2021nerd, srinivasan2021nerv} to guide the predicted surface normals $\mathbf{n}$ using a weighted MSE loss: 
\begin{align}\label{eq:losses}
    L_{C} &= \sum_{\mathbf{r} \in \mathcal{R}}\left\|C(\mathbf{r})-C_{gt}(\mathbf{r})\right\|_{2}^{2} \ , \\
    L_{S} &= -\sum_{\mathbf{r} \in \mathcal{R}} [\sum_{l \in L}\hat{p}(\mathbf{r})^l\log p(\mathbf{r})^l] \ ,\\
    L_{\mathbf{n}} &= \sum_{\mathbf{r} \in \mathcal{R}}\sum_{i=N}^{N}w_i\|\mathbf{n_i}-\hat{\mathbf{n}}_i\|_{2}^{2} \ ,
\end{align}
where $w_i=T_i(1-\text{exp}(-\sigma_i\delta_i))$ denotes the detached weight in Eq.\ref{eq:nerf_volume}. 
We also leverage distortion loss $L_{dist}$ ~\cite{barron2022mip, sun2022improved} to mitigate floaters in the reconstruction results:
\begin{align}\label{eq:dist_loss}
    L_{\text{dist}}=&\sum_{i=0}^{N-1} \sum_{j=0}^{N-1} w_{i} w_{j}\left|\frac{t_{i}+t_{i+1}}{2}-\frac{t_{j}+t_{j+1}}{2}\right| %
    +\frac{1}{3} \sum_{i=0}^{N-1} w_{i}^{2}\left(t_{i+1}-t_{i}\right)
\end{align}
However, ngp model tends to create big 'blobs' in the sky with distortion loss. 
We alleviate this by applying a simple penalty $L_{sky}=\sum_{\mathbf{r} \in \mathcal{R}} \mathrm{e}^{-D(\mathbf{r})}\cdot \mathbbm{1}\{\hat{p}(\mathbf{r})=\hat{p}_{sky}\}$ where $D(\mathbf{r})$ denotes the depth following Eq.~\ref{eq:nerf_volume}~\cite{yu2022monosdf} and $\mathbbm{1}\{\cdot\}$ is an indicator function using 2D predicted semantic logits $\hat{p}$. 
Moreover, we incorporate opacity loss $L_O= -\sum_{\mathbf{r} \in \mathcal{R}} O(\mathbf{r})\log O(\mathbf{r})$ ~\cite{queianchen_ngp} to encourage ray opacity being either 0 or 1 to avoid semi-transparent regions in reconstruction results.
During our training time, our model's total loss is a weighted sum of the aforementioned losses:
\begin{align}\label{eq:total_loss}
    L=L_C+\lambda_{S}L_S+&\lambda_{\mathbf{n}}L_{\mathbf{n}}+\lambda_{dist}L_{dist}
             +\lambda_{sky}L_{sky}+\lambda_{O}L_O
\end{align}

\paragraph{Transient Object Occlusion}
To occlude transient objects like pedestrians or vehicles across views in tanks and temples dataset ~\cite{knapitsch2017tanks}, we follow ~\cite{chen2022hallucinated} and create per-image learnable masks:
\begin{equation}\label{eq:mask}
    M_{i}(u, v) = F_{\psi}(u, v, i, \gamma_{M}) \ ,
\end{equation}
where $(u,v)\in\mathbb{R}^2$, $i$ denotes image coordinates and image index in all training images. $F_{\psi}$ denotes a shallow MLP and $\gamma_{M}$ is the output of hash grids.

In such case, we change color reconstruction loss~\ref{eq:losses} following ~\cite{chen2022hallucinated}:
\begin{equation}\label{eq:mask_rgb_loss}
    L_{C} = \sum_{\mathbf{r} \in \mathcal{R}}M(\mathbf{r})\|C(\mathbf{r})-C_{gt}(\mathbf{r})\|^{2}_{2} + \lambda_{M}(1-M({\mathbf{r}})) \ ,
\end{equation}
where the second term is used to prevent the mask from predicting everything transient.

\paragraph{Geometry improvements}

As mentioned in the limitation subsection in the main paper and can be seen in Fig.~\ref{fig:6_limitations}, ClimateNeRF strongly relies upon high-quality geometry. 
To improve geometry estimations for KITTI-360 dataset~\cite{liao2022kitti}, we further use the monocular dense depth $D_{\text{mono}}$ and impose a depth loss:
\begin{equation}\label{eq:loss_depth}
    L_{\text{depth}} = \sum_{\mathbf{r} \in \mathcal{R}} \|((wD(\mathbf{r})+d)-D_{\text{mono}}(\mathbf{r}))\|^2_2 \ ,
\end{equation}
where $w$ and $d$ are used to align the predicted depth from NeRF $D(\mathbf{r})$ and monocular depth cue $D_{\text{mono}}(\mathbf{r})$ with a least-square criterion~\cite{yu2022monosdf, eigen2014depth, ranftl2020towards}. 
As shown in Fig.\ref{fig:geometry_improve}, holes on the ground are ``filled" thanks to the supervisions from monocular dense depth.
\input{tables/geometry.tex}
\paragraph{Training Details}

Our implementations of the hash grids and distortion loss follow~\cite{tiny-cuda-nn, queianchen_ngp}.
For the scale and resolution of the hash grid in our model, we get the inspiration from ~\cite{sun2022direct}, where appearance features allocate more grids with finer resolutions. 
Geometry hash grids have 16 levels and $2^{19}$ entries at most per level, while appearance hash grids have 32 levels and $2^{21}$ entries at most per level. For Tanks and Temples dataset~\cite{knapitsch2017tanks} and MipNeRF-360 dataset~\cite{barron2022mip}, side length of hash grids is $16$. 
Geometry MLP and appearance MLP have 128 neurons per layer and one hidden layer, while semantic MLP and normal predicting MLP have 32 neurons per layer and one hidden layer. 
For our default loss weights in Eq.~\ref{eq:total_loss}, we set $\lambda_S=4^{-2}$, $\lambda_{\mathbf{n}}=7^{-4}$, $\lambda_{dist}=3^{-4}$, $\lambda_{sky}=1^{-1}$ and $\lambda_{O}=2^{-4}$.

\subsection{Flood Simulation}\label{sect:flood_sim}

Given that water is mostly muddy and non-transparent, we approximate the opacity by checking a point \emph{above} or \emph{under} the water's surface:
\begin{equation}
    O_\phi(\bx; F'_\theta) = 
        \begin{cases}
            \infty   & \quad \text{if } \bn_w(\bx - \bo_w) = 0 \\
            0        & \quad \text{otherwise}
        \end{cases}
\end{equation}

Water is highly reflective, yet the microfacet ripples sometimes make the water look glossy. 
To simulate these effects, we leverage a Spherical Gaussian (SG) to approximate the BRDF on the reflective water surface:
\begin{equation}
B_{\phi}(\mathbf{x}, -\mathbf{d}, \bomega_i, N_\phi(\bx)) = \exp^{\lambda(-\bomega_i \cdot \bd_r - 1)}
\end{equation}
where SG lobe axis is $\bd_r = \bd - 2(\bd \cdot N_\phi(\bx))N_\phi(\bx)$,, and $\lambda \in \bbR_+$ is the lobe sharpness, controlling the glossy effects. 

We use sigma point-based sampling for rendering water. 
Specifically, the camera ray $\br(t)=\bo + t\bd$ is cast from the camera and hits the water surface at position $\bx$. 
We compute the observed color by a sampling-based approximation of the rendering equation:
\begin{equation}\label{eq:flood_color}
    \bc_{\phi} = (1 - R)\bc_{w} + R\displaystyle\sum_{i=1}^5 L(\mathbf{x}, \bomega_i)e^{\lambda(-\bomega_i \cdot \bd_r - 1)},
\end{equation} 
where $L(\bx, \bomega_i) = C(\br)$ is the NeRF ray color (Eq.~\ref{eq:nerf_volume}), representing the incident light color hitting $\bx$ along direction $\bomega_i$.  $R \in (0, 1)$ is the reflectance index determined by viewing direction $d$ and normal $N_\phi(\bx)$, which enables the system to simulate the Fresnel effect on the water.
To approximate the integral in rendering equation~\cite{kajiya1986rendering}, we adopt the sigma-point method~\cite{menegaz2015systematization, wan2000unscented} and sample 5 rays from the position $\bx$, including reflection direction $\bd_r$ and nearby four rays. 
In short, \method~simulates Fresnel effect, glossy reflection, and wave dynamics.]

\paragraph{Fresnel Effect}
When the light hits the water's surface, the amount of reflection and transmission is determined by the incident and normal directions and is described by the Fresnel effect. 
The angle between normal and incident rays is denoted by $\theta_i$, and the angle between normal and refracted ray in water is $\theta_t$. 
According to Snell's Law: $r_i\theta_i = r_t\theta_t$, where $r_i = 1$ is the refraction index of air and $r_t=1.33$ is the refraction index of water in our experiments, which is also consistent with real-world water properties. 
Next, the reflectance $R$ in Eq.~\ref{eq:flood_color} is computed by:

\begin{equation}
    R = \frac{R_s+R_p}{2}, \quad R_s=\left[ \frac{\text{sin}(\theta_t-\theta_i)}{\text{sin}(\theta_t+\theta_i)} \right]^2, \quad R_p = \left[ \frac{\text{tan}(\theta_t-\theta_i)}{\text{tan}(\theta_t+\theta_i)} \right]^2
\end{equation}
where $R_s$ and $R_p$ are the reflectances for s-polarized and p-polarized light, respectively. 
Modeling the Fresnel effect in our flood simulation pipeline makes the water far from the camera (larger $\theta_i$) have higher $R$ and looks more like a mirror. 
The water nearby (smaller $\theta_i$) has lower $R$ and shows watercolor, enhancing the simulation's realism.

\paragraph{Refraction Effect} 
Refraction occurs when light transports from one medium to another.
Modeling such an effect helps simulate clear water.
With a ray direction, surface normal, and refraction index $r_t$ of the water, \method~calculates the refraction direction $\bomega_t$ according to Snell's Law, and retrieves color with volume rendering along $\bomega_t$.
Next, we can simulate underwater scenes following ~\cite{sethuraman2022waternerf}.
\begin{equation}\label{eq:flood_refraction}
    \bc_{\phi} = (1 - R)(t
_c(\mathbf{x})L(\bx, \bomega_t)+(1-t_c(\mathbf{x})\bc_{w}) + R\displaystyle\sum_{i=1}^5 L(\mathbf{x}, \bomega_i)e^{\lambda(-\bomega_i \cdot \bd_r - 1)},
\end{equation} 
where $t_c(\mathbf{x})=e^{-\beta_cD(x,\bomega_t)}$ and $D(x,\bomega_t)$ denotes the geometry depth alone refraction direction.
Consequently, \method~can simulate reflection and refraction, simulating clear and muddy water with controllable parameter $\beta_c$.

\subsection{Snow Simulation}
For any point $\mathbf{x}$ in the space, we calculate the snow's density of $\mathbf{x}$ in a particle-based manner. 
We first figure out a set of $N$ particles as metaballs' centers $\{\mathbf{x}^{(p)}_i\}_{i=1}^{N}$ with densities $\{\sigma^{(p)}_i\}_{i=1}^N$ and metaball radius $\{R^{(p)}_i\}_{i=1}^N$ around $\mathbf{x}$. Then we sum up the densities calculated by kernel function $\mathbf{K}(r,
R, \sigma_o)$. 
\begin{align}\label{eq:snow_density}
\begin{split}
    \sigma_{\text{snow}}(\mathbf{x})&=\sum_{i=1}^{N} \sigma_{\mathbf{K}}(\mathbf{x}, \mathbf{x}^{(p)}_i), \\
    \text{where } \sigma_{\mathbf{K}}(\mathbf{x}, \mathbf{x}^{(p)}_i) &= \mathbf{K}(\|\mathbf{x}-\mathbf{x}^{(p)}_i\|_2, R^{(p)}_i, \sigma^{(p)}_i)V_g(\mathbf{x}^{(p)}_i, \bomega_s),
\end{split}
\end{align}
where $\sigma^{(p)}_i$ is defined by weights during volume rendering~\ref{eq:nerf_volume} for $\sigma_{\theta}$ of $F'_{\theta}$. 
More details are shown in Section~\ref{extensions}. 
We denote $V_g(\mathbf{x}^{(p)}_i, \bomega_s)$ as the transparency along snow falling direction $\bomega_s$ retrieved from a pre-trained visibility network based on pre-trained NeRF's geometry to simulate snow occlusions. 
When training the visibility network, we perturb snow-falling directions within a small angle to soften snow boundaries. 
During rendering, we identify snow surface by a threshold $\tau_\text{snow}$ and a hyperparameter $a$:
\begin{equation}\label{eq:snow_density_final}
    O_\phi(\bx; F'_\theta) = 
    \frac{1}{1+e^{-a(x\sigma_{\text{snow}}-\tau_\text{snow})}} \sigma_{\text{snow}},
\end{equation}

The BRDF of snow particles is set as spatially-varying diffuse color $\mathbf{c}_\phi(\bx_i^{(p)})$ close to pure white multiplied by the average illumination of the scene. 
Furthermore, since the snow is semi-transmissive, the subsurface scattering effect~\cite{nishita1997modeling} will light the snow's shadowed part. 
To simulate such effect, we leverage warp lighting function~\cite{green2004real} $\Phi(\mathbf{n}_\mathbf{K}, \mathbf{n}_l, \gamma_{\Phi})$ based on normalized surface normal $\mathbf{n}_\mathbf{K}$, light vector $\mathbf{n}_l$ and hyperparameter $\gamma_{\Phi}$. 
We use 2D shadow predictions~\cite{chen2020multi} and bake shadow confidence $V_s(\mathbf{x})\in[0, 1]$ on pre-trained NeRF's geometry using $L_2$ loss. 
For an arbitrary point $\bx$ in space, the color of point $\mathbf{x}$ is a weighted sum of $\{\mathbf{c}^{(p)}_i\in\bbR^1\}_{i=1}^N$ based on kernel function:
\begin{align}\label{eq:snow_color}
\begin{split}
    \mathbf{c}_{\phi}(\mathbf{x}) &= \frac{\sum_{i=1}^{N}\sigma_{\mathbf{K}}(\mathbf{x}, \mathbf{x}^{(p)}_i)\frac{\mathbf{c}^{(p)}_i+\mathbf{c}_0}{1+\mathbf{c}_0}(\beta_sV_s(\mathbf{x}^{(p)}_i)+V_{s0})}
    {\sum_{i=1}^{N}\sigma_{\mathbf{K}}(\mathbf{x}, \mathbf{x}^{(p)}_i)} \Phi(\mathbf{n}_\mathbf{K}(\mathbf{x}), \mathbf{n}_l, \gamma_{\Phi}) \ ,
\end{split}
\end{align}
where $\Phi(\mathbf{n}_\mathbf{K}(\mathbf{x}), \mathbf{n}_l, \gamma_{\Phi})=\frac{\mathbf{n}(\mathbf{x})\cdot\mathbf{n}_l+\gamma_{\Phi}}{1+\gamma_{\Phi}}$ and 
$\frac{\mathbf{c}'^{(p)}_i+\mathbf{c}_0}{1+\mathbf{c}_0}$ is used to approximate a high albedo for snow and $\mathbf{c}_0$ is a hyperparameter. To recast shadow on snow, we let $\beta_s$ and $V_{s0}$ in Eq.~\ref{eq:snow_color} be hyperparameters. See Figure~\ref{fig:kitti} for example. Surface normal values of metaballs are still calculated in a gradient-based manner. 
Moreover, we stylize the scene to match the color tones of different weather conditions, as shown in Fig.~\ref{fig:4_stylization}.

\subsection{Extensions}\label{extensions}

\paragraph{Bake Editing}
When simulating physical entities, especially snow, rendering will be time-consuming if we straightly let snow fall from the sky and do collision detection. 
Moreover, a lack of supervision in a bird's eye view makes depth estimation for rays cast from the sky inaccurate. 
To mitigate the aforementioned issues, we fit the distribution of metaballs' densities and colors in a new model and fetch them in a particle-based manner. 
The new model outputs high densities where metaballs locate. 
We use $w_i=T_i(1-\text{exp}(-\sigma_i\delta_i))$ in Eq.~\ref{eq:nerf_volume} for surface detection since $w(\mathbf{x})\in[0,1]$ is close to 1 when $\mathbf{x}$ is close to surfaces. 
To identify surfaces where snow accumulates, we incorporate surface normals $\mathbf{n}$ and vertical axis $\mathbf{n}_\perp$ to figure out metaballs' density weights: $w^{(p)}_i=\frac{1}{1+e^{-a'(\mathbf{n}_i\cdot\mathbf{n}_\perp-\cos(\theta_0))}}w_i$ where $\theta_0$ is a hyperparameter. 
We then bake $w^{(p)}_i$ into a new model $G_{\phi_w}$:
\begin{equation}\label{eq:bake_geometry}
    w^{(p)}_{\phi_w}(\mathbf{x})=G_{\phi_w}(\mathbf{x})
\end{equation}
 We also bake the gray scale~\cite{stokes1996standard} of $\mathbf{c}_{\theta}$ from $F_{\theta}$ into a new model $G_{\phi_\mathbf{c}}$ to capture an approximation for light intensities and shadows:
\begin{equation}\label{eq:bake_color}
    \mathbf{c}^{(p)}_{\phi_\mathbf{c}}(\mathbf{x})=G_{\phi_\mathbf{c}}(\mathbf{x})
\end{equation}

\input{figures/extensions/voxel_nest}

Then, we leverage the pre-trained $G_{\phi_w}$, $G_{\phi_\mathbf{c}}$ and do voxel sampling to fetch $\{\sigma^{(p)}_i\}_{i=1}^N$ and $\{\mathbf{c}^{(p)}_i\}_{i=1}^N$ from 8 vertices. 
Also, to automatically alter metaballs' radiuses according to the size of the surface, we sample nested grids with different side lengths defined in a geometric progression. 
Moreover, we define metaballs' radiuses by girds' side lengths. 
Hence, Eq.~\ref{eq:snow_density} and Eq.~\ref{eq:snow_color} can be rewritten as:
\begin{align}\label{eq:voxel_cascade}
    \sigma_{\text{snow}}(\mathbf{x}) = \sum_{i=1}^{8N_{n}} \sigma_{\mathbf{K}}(\mathbf{x}, \mathbf{x}^{(p)}_{i}); \quad
    \mathbf{c}_{\phi}(\mathbf{x}) = \frac{\sum_{i=1}^{8N_{n}}\sigma_{\mathbf{K}}(\mathbf{x}, \mathbf{x}^{(p)}_i)\frac{\mathbf{c}^{(p)}_i+\mathbf{c}_0}{1+\mathbf{c}_0}(\beta_sV_s(\mathbf{x}^{(p)}_i)+V_{s0})}
    {\sum_{i=1}^{8N_{n}}\sigma_{\mathbf{K}}(\mathbf{x}, \mathbf{x}^{(p)}_i)} \Phi(\mathbf{n}_\mathbf{K}(\mathbf{x}), \mathbf{n}_l, \gamma_{\Phi}),
\end{align}
where $N_n$ is the number of nests. We calculate density $\sigma^{(p)}$ and albedo color $\mathbf{c}^{(p)}$ for metaball centered at $\mathbf{x}^{(p)}$ by $\sigma^{(p)}=w^{(p)}_{\phi_w}(\mathbf{x}^{(p)}) \sigma_0$ and $\mathbf{c}^{(p)}=\mathbf{c}^{(p)}_{\phi_\mathbf{c}}(\mathbf{x})$ where $\sigma_0$ is a hyperparameter. 
See Fig.~\ref{fig:voxel_cascade} for a visualization of this sampling strategy. 
If stylization is done on the scene, we leverage the stylized model $F'_{\theta}$ and finetune the $G_{\phi_\mathbf{c}}$ to match new illumination conditions while remaining $G_{\phi_w}$ intact since $F'_{\theta}$ shares the same spatial information with $F_{\theta}$.

\paragraph{Anti-Aliasing}
When rendering with simulation, the high-frequency normal map changes on the physical entity surface would lead to an aliasing effect. 
To alleviate such artifacts, we can render four times larger images with higher resolution and perform anti-aliasing downsampling to the original resolution.

%% file: tables/geometry.tex
\begin{table*}[t]
  \centering
  \setlength\tabcolsep{0.05em}
  \resizebox{\linewidth}{!}{
    \begin{tabular}{cc}
        \includegraphics[width=0.45\linewidth]{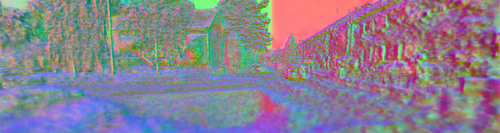} &
        \includegraphics[width=0.45\linewidth]{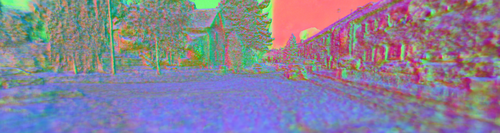} \\
        (a) w/o monocular dense depth $D_{\text{mono}}$ &
        (b) w/ monocular dense depth $D_{\text{mono}}$
    \end{tabular}
  }
   \captionof{figure}{\textbf{Ablation on monocular dense depth supervision} Normal estimations in (b) shows that leveraging monocular dense depth removes artifacts on the road.}
   \label{fig:geometry_improve}
\end{table*}

%% file: figures/extensions/voxel_nest.tex
\begin{table*}[t]
  \centering
  \resizebox{0.5\textwidth}{!}{
    \begin{tabular}{cc}
        \includegraphics[width=0.26\textwidth]{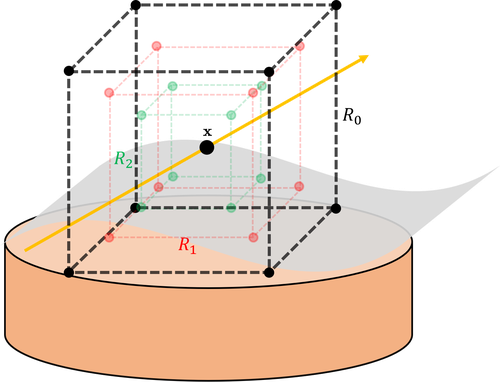} &
        \includegraphics[width=0.2\textwidth]{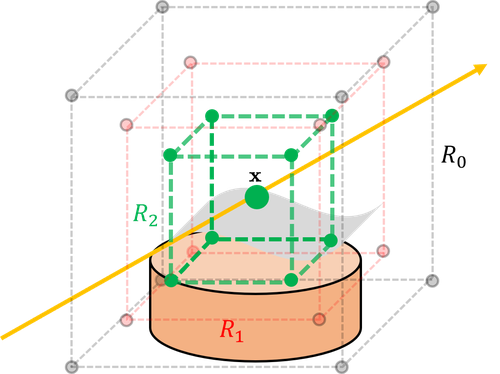} \\
        
        (a) Sample near a large surface &
        (b) Sample near a tiny surface
    \end{tabular}
  }
   \captionof{figure}{\textbf{Visualization of voxel+nest sampling for $N_n=3$.} Instead of storing metaball information in point cloud, we approximate metaball's center density distributions with $G_{\phi_w}$. Our voxel+nest sampling helps capture different geometric level of details~\cite{takikawa2021neural}, which makes large metaballs dominate at flat surfaces, and put smaller snow balls on thin structures. }
   \label{fig:voxel_cascade}
\end{table*}

%% file: supp/ablation.tex
\section{Ablation Study}\label{sect:ablation}
To justify our design choices, we perform an ablation study of flood simulation, and the results are shown in Fig~\ref{fig:flood_ablation}. 
Specifically, we report the simulation results without certain technical components depicted in Sec.~\ref{sect:flood_sim} of the main paper. 

Fig~\ref{fig:flood_ablation} shows that all components are essential for realism. For example, vanishing point detection~\cite{lu20172} makes the water plane follow gravity direction; wave simulation adds ripples to the water surface; the Fresnel effect makes the water reflectance view-dependent and physically plausible; the Glossy effect mimics realistic microfacet water surfaces with ripples; anti-aliasing removes far-away high-frequency noises. In short, all components contribute to the realism of the simulation.

We also perform an ablation study on snow simulation to validate our approximate scattering rendering in Fig.~\ref{fig:snow_ablation}. We compare 1) pure white metaballs with spatial variant colors, 2) metaballs in a fully diffuse model, and 3) our full simulation. Results demonstrate that our choice provides a more realistic rendering of accumulated snow. %

To identify how scene geometry improvements benefit simulations, the ablation study of various technical components is illustrated in Fig.~\ref{fig:geometry_improvement}. For example, appearance embeddings mitigate the floaters caused by varying exposures in training views, and sky loss reduces blobs in the sky. Results show that our improvements in geometry reconstructions help simulate dense snow covering in both foreground and background, thus offering more visually pleasing results.

\input{tables/flood_ablation.tex}

\input{tables/snow_ablation.tex}

\input{tables/recon_ablation.tex}

%% file: tables/flood_ablation.tex
\begin{table*}[t]
  \centering
  \footnotesize
  \setlength\tabcolsep{0.05em}
  \resizebox{\textwidth}{!}{
    \begin{tabular}{ccc}

        \includegraphics[width=0.3\textwidth]{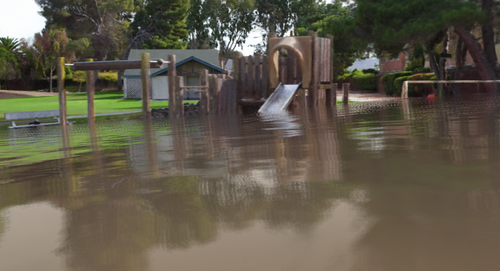} & 
        \includegraphics[width=0.3\textwidth]{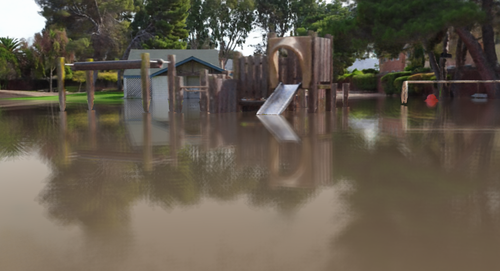} &
        \includegraphics[width=0.3\textwidth]{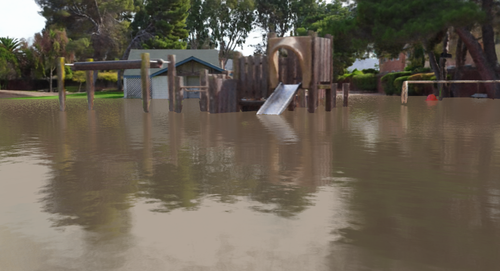} \\
        (a) Inaccurate plane geometry & (b) No wave simulation & (c) No Fresnel effect \\
        \includegraphics[width=0.3\textwidth]{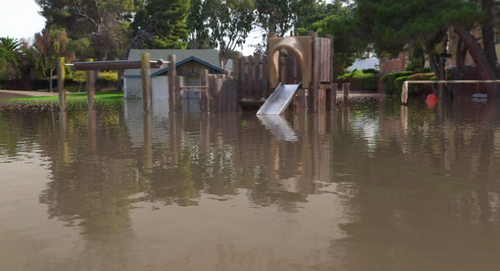} & 
        \includegraphics[width=0.3\textwidth]{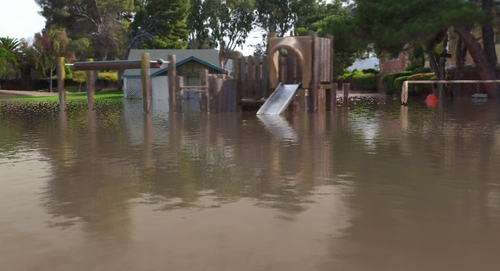} &
        \includegraphics[width=0.3\textwidth]{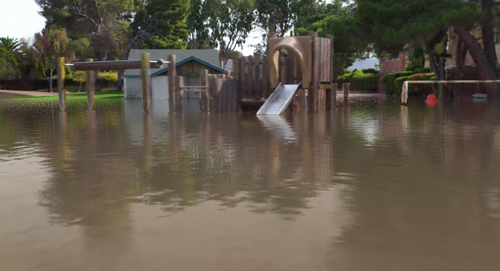} \\
        (d) No glossy effect & (e) No anti-aliasing & (f) Full simulation

    \end{tabular}
  }
   \captionof{figure}{\textbf{Ablation Study of Flood Simulation.} (a) Without accurate plane geometry estimation with vanishing point detection~\cite{lu20172}, the water surface deviates from gravity direction. (b) The surface is perfect planar without wave simulation, which is not natural. (c) Fresnel effect makes the water far from the camera (with a larger incident angle) have higher reflectance, and is consistent with physical rules. (d) The glossy effect makes the reflection more blurry and realistic (e) There is much high-frequency noise around the water border without an anti-aliasing trick. (f) Our full flood simulation results.}
   \label{fig:flood_ablation}
\end{table*}

%% file: tables/snow_ablation.tex
\begin{table*}[t]
  \centering
  \footnotesize
  \setlength\tabcolsep{0.05em}
  \resizebox{\textwidth}{!}{
    \begin{tabular}{cccc}
        \includegraphics[width=0.2\textwidth]{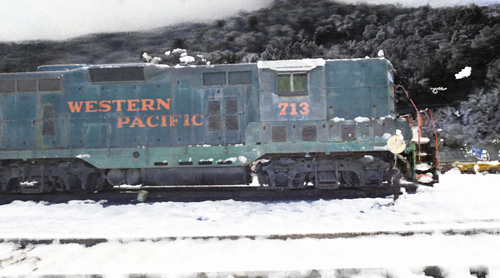} &
        \includegraphics[width=0.2\textwidth]{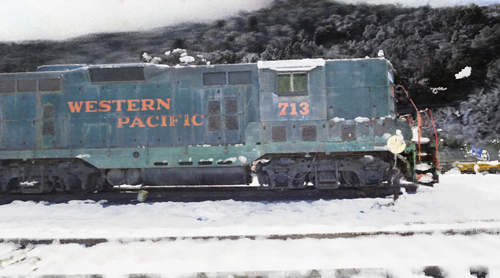} &
        \includegraphics[width=0.2\textwidth]{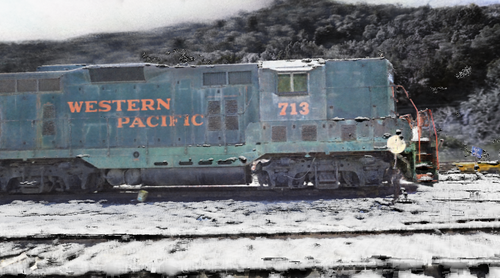} &
        \includegraphics[width=0.2\textwidth]{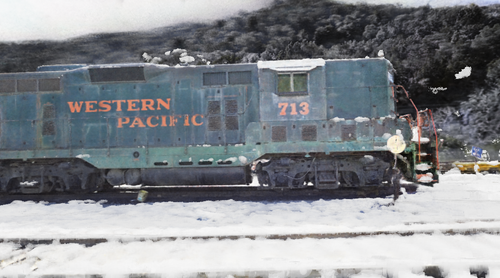} \\
        
        \includegraphics[width=0.2\textwidth]{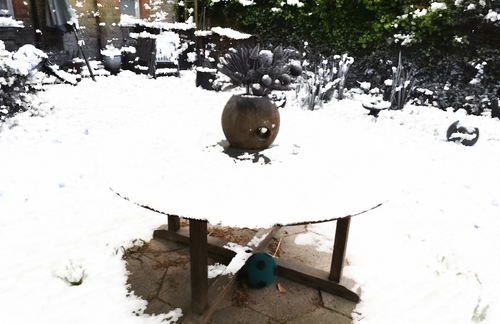} &
        \includegraphics[width=0.2\textwidth]{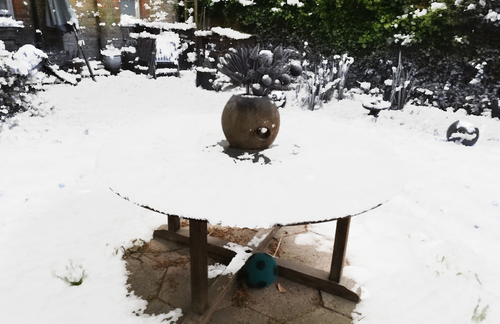} &
        \includegraphics[width=0.2\textwidth]{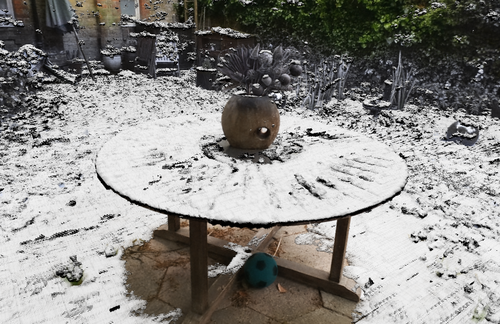} &
        \includegraphics[width=0.2\textwidth]{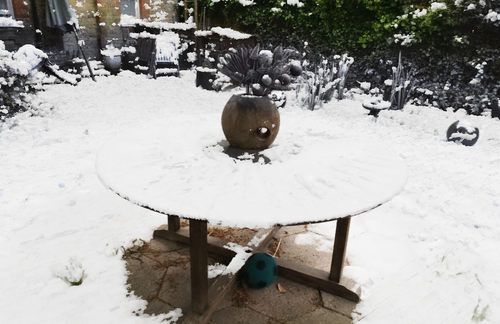} \\
        
        Pure White Snow &
        Only spatial variant colors &
        Only diffuse model &
        Full model
    \end{tabular}
  }
   \captionof{figure}{\textbf{Ablation Study for Snow Simulation} Though spatial variant colors capture local illumination conditions, they fail to offer a sense of depth for snow. Moreover, the diffuse model cannot simulate snow's scattering effects. }
   \label{fig:snow_ablation}
\end{table*}

%% file: tables/recon_ablation.tex
\begin{table*}[t]
  \centering
  \footnotesize
  \setlength\tabcolsep{0.05em}
  \resizebox{0.7\textwidth}{!}{
    \begin{tabular}{ccccc}

        \raisebox{5mm}[0pt][0pt]{\rotatebox[origin=c]{90}{\makecell[c]{\tiny{NGP~\cite{muller2022instant}}}}} & \includegraphics[width=0.15\textwidth]{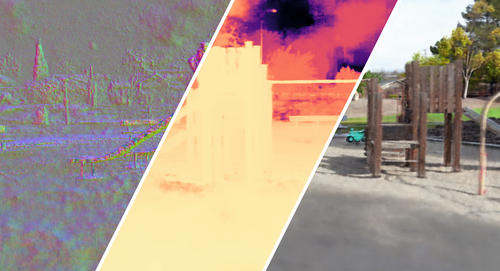} &  
        \includegraphics[width=0.15\textwidth]{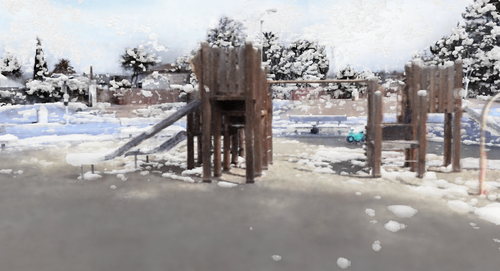} \\
        
        \raisebox{6mm}[0pt][0pt]{\rotatebox[origin=c]{90}{\makecell[c]{\tiny{w/o appearance} \\ \tiny{embedding}}}} & \includegraphics[width=0.15\textwidth]{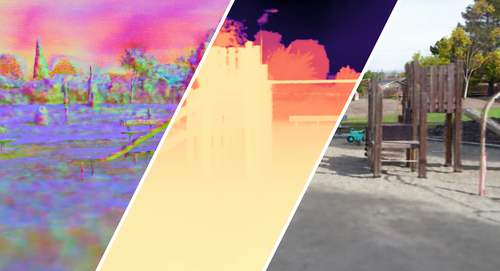} &  
        \includegraphics[width=0.15\textwidth]{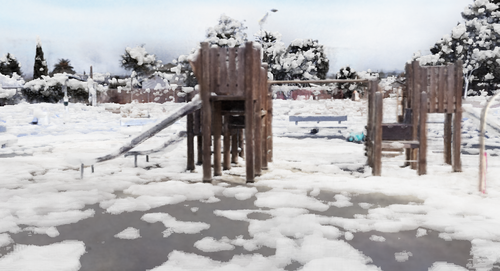} \\
        
        \raisebox{6mm}[0pt][0pt]{\rotatebox[origin=c]{90}{\makecell[c]{\tiny{w/o distortion loss}}}} & \includegraphics[width=0.15\textwidth]{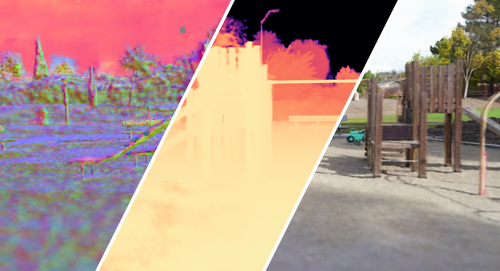} &  
        \includegraphics[width=0.15\textwidth]{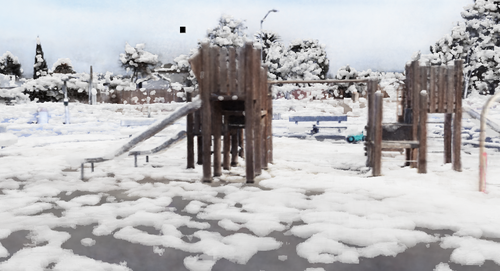} \\
        
        \raisebox{6mm}[0pt][0pt]{\rotatebox[origin=c]{90}{\makecell[c]{\tiny{w/o normal} \\ \tiny{predictions}}}} & \includegraphics[width=0.15\textwidth]{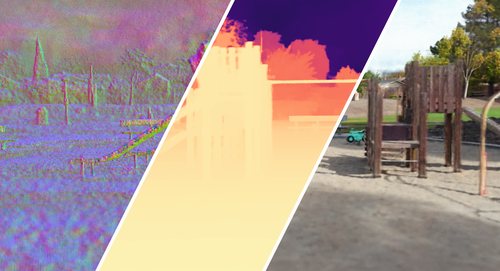} &  
        \includegraphics[width=0.15\textwidth]{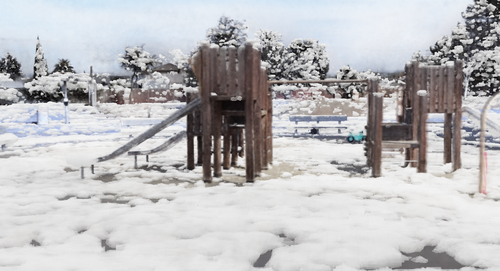} \\
        
        \raisebox{6mm}[0pt][0pt]{\rotatebox[origin=c]{90}{\makecell[c]{\tiny{w/o sky loss}}}} & \includegraphics[width=0.15\textwidth]{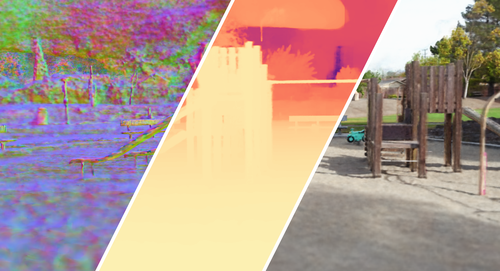} &  
        \includegraphics[width=0.15\textwidth]{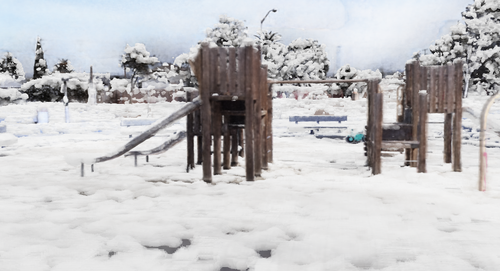} \\
        
        \raisebox{6mm}[0pt][0pt]{\rotatebox[origin=c]{90}{\makecell[c]{\tiny{Ours}}}} & 
        \includegraphics[width=0.15\textwidth]{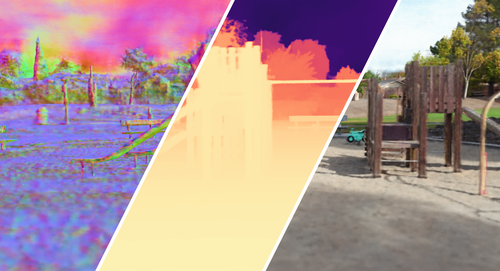} & 
        \includegraphics[width=0.15\textwidth]{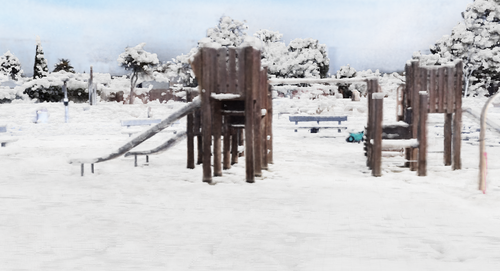} \\
       & \tiny{(a) Normal/Depth/RGB} & \tiny{(d) Snow} \\

    \end{tabular}
  }
  \vspace{-3mm}
   \captionof{figure}{\textbf{Geometry Improvement for Simulation.} Appearance embeddings~\cite{martin2021nerf} and distortion loss~\cite{barron2022mip} mitigate floaters or incomplete geometries while normal prediction loss~\cite{verbin2022refnerf} smooths predicted normals and leads a well interpolation in background(see the snow coverage in the background). Sky loss offers primitives for sky depth and alleviates 'blobs' in the sky.} 
   \label{fig:geometry_improvement}
   \vspace{-3mm}
\end{table*}

%% file: supp/control.tex
\section{Controllability}\label{sect:control}

We further demonstrate that \method~is highly controllable during the simulation process. In Fig.~\ref{fig:control}, our method simulate different colors of smog and flood, water clearness, varying spatial frequency of water ripples, and distinct heights of accumulated snow. The results show that our simulation framework is highly controllable by the users. Consequently, scientists can use this framework to simulate accurate climate conditions depending on the projected climate in the future and visualize the consequences corresponding to different actions taken by policymakers and the general public.

\input{tables/controllability.tex}

%% file: tables/controllability.tex
\begin{table*}[t]
  \centering
  \footnotesize
  \setlength\tabcolsep{0.05em}
  \resizebox{\textwidth}{!}{
    \begin{tabular}{ccccc}

        \includegraphics[width=0.2\textwidth]{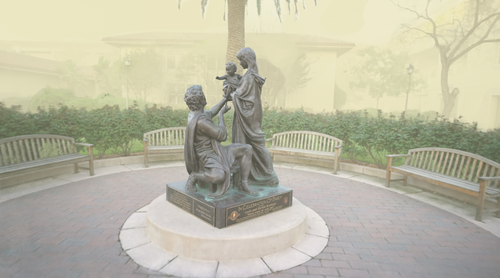} &
        \includegraphics[width=0.2\textwidth]{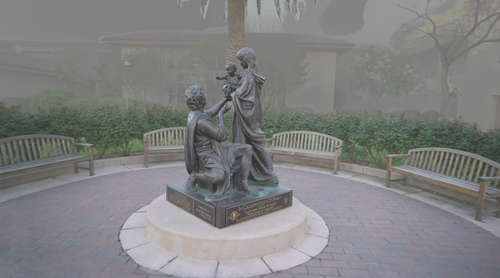} &
        \includegraphics[width=0.2\textwidth]{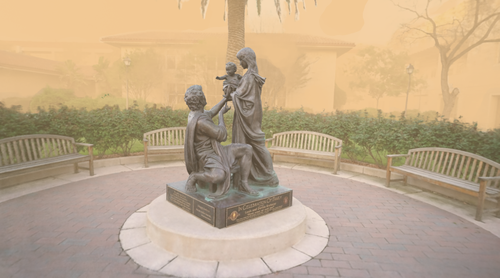} &
        \includegraphics[width=0.2\textwidth]{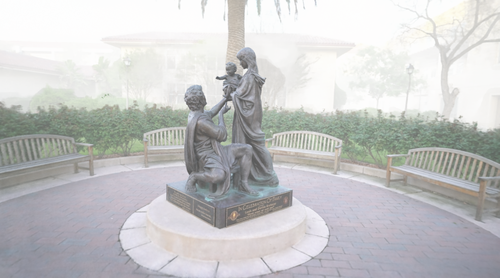} &
        \includegraphics[width=0.2\textwidth]{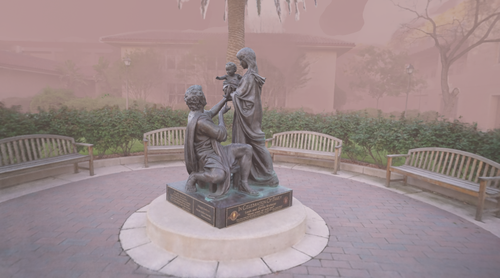} \\

        \includegraphics[width=0.2\textwidth]{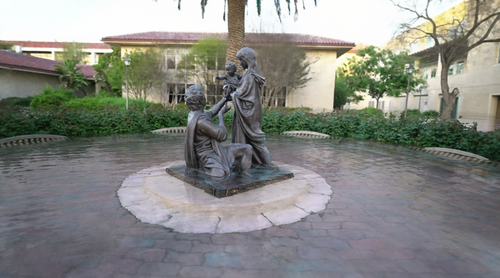} &
        \includegraphics[width=0.2\textwidth]{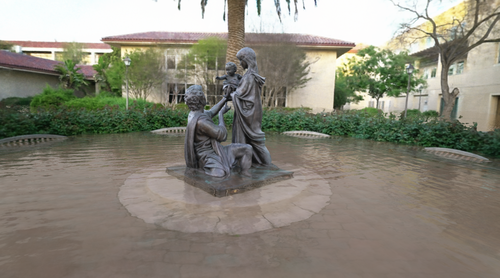} &
        \includegraphics[width=0.2\textwidth]{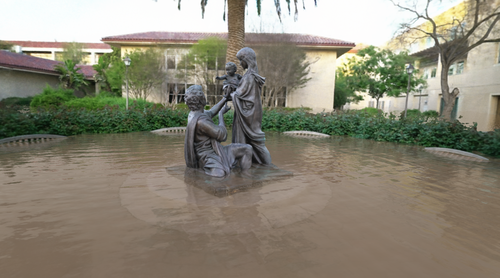} &
        \includegraphics[width=0.2\textwidth]{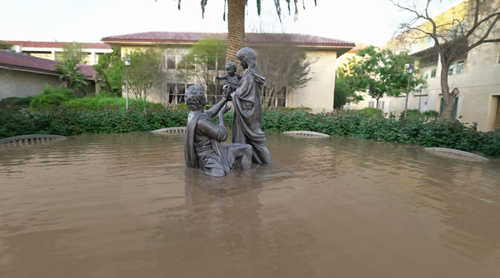} &
        \includegraphics[width=0.2\textwidth]{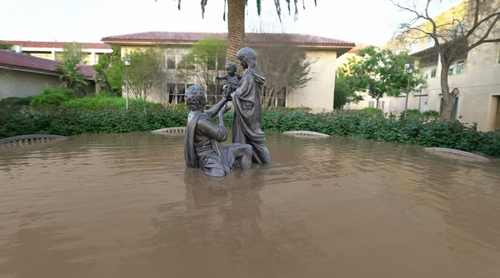} \\

        \includegraphics[width=0.2\textwidth]{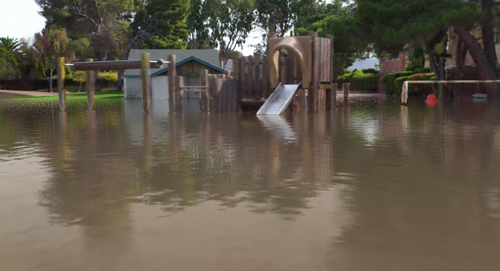} &
        \includegraphics[width=0.2\textwidth]{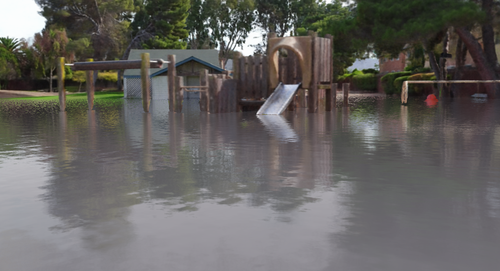} &
        \includegraphics[width=0.2\textwidth]{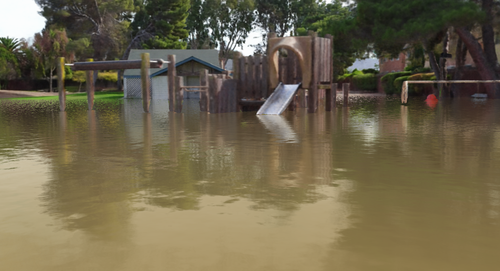} &
        \includegraphics[width=0.2\textwidth]{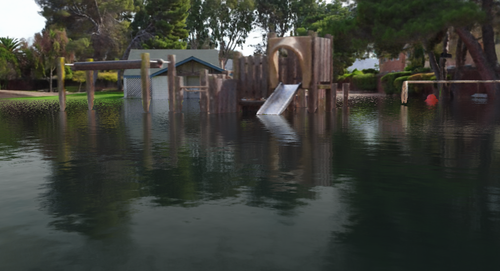} &
        \includegraphics[width=0.2\textwidth]{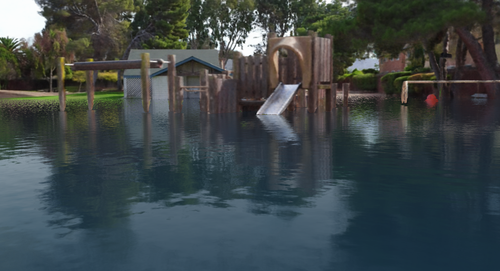} \\

        \includegraphics[width=0.2\textwidth]{figures/images/controllability/flood/flood-000-default.png} &
        \includegraphics[width=0.2\textwidth]{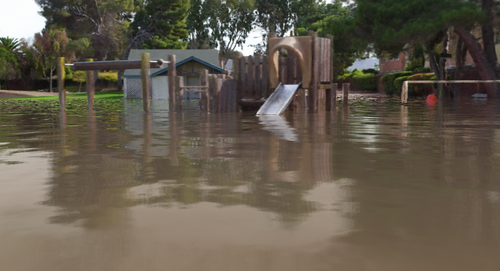} &
        \includegraphics[width=0.2\textwidth]{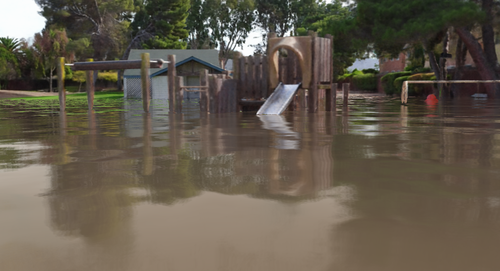} &
        \includegraphics[width=0.2\textwidth]{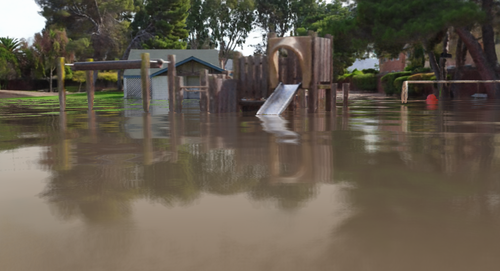} &
        \includegraphics[width=0.2\textwidth]{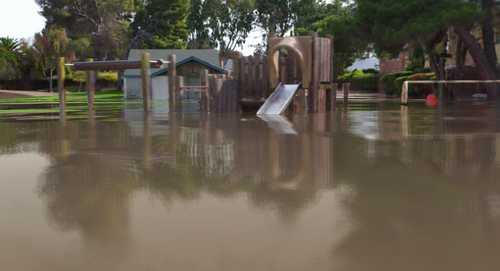} \\

        \includegraphics[width=0.2\textwidth]{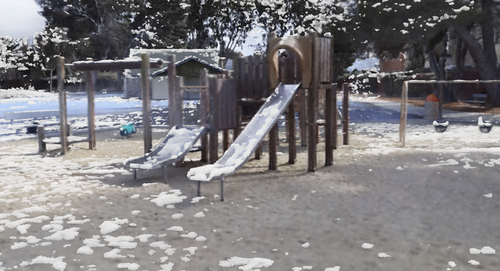} &
        \includegraphics[width=0.2\textwidth]{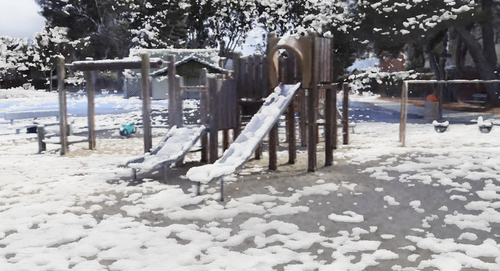} &
        \includegraphics[width=0.2\textwidth]{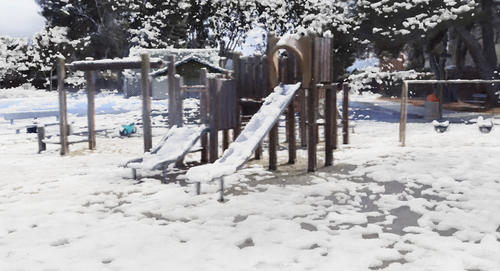} &
        \includegraphics[width=0.2\textwidth]{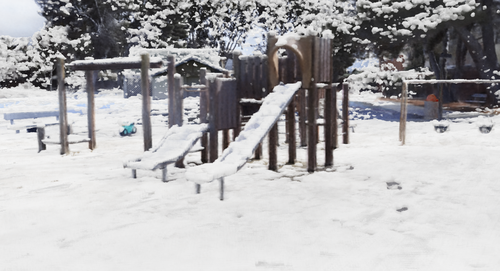} &
        \includegraphics[width=0.2\textwidth]{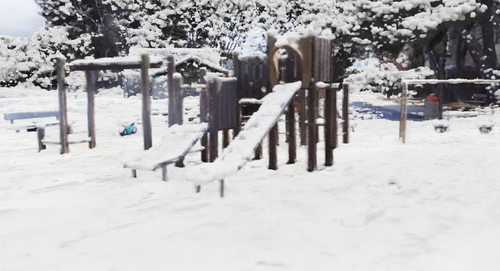} \\
        
        \includegraphics[width=0.2\textwidth]{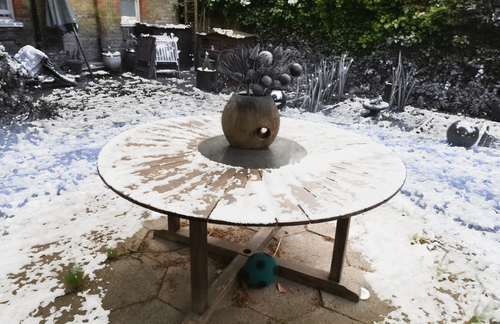} &
        \includegraphics[width=0.2\textwidth]{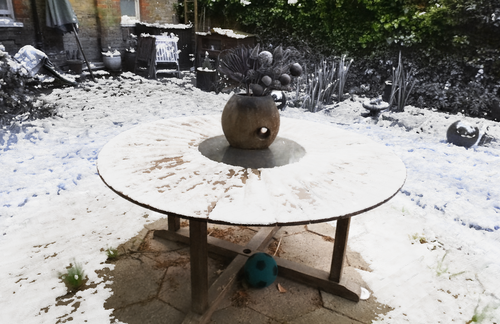} &
        \includegraphics[width=0.2\textwidth]{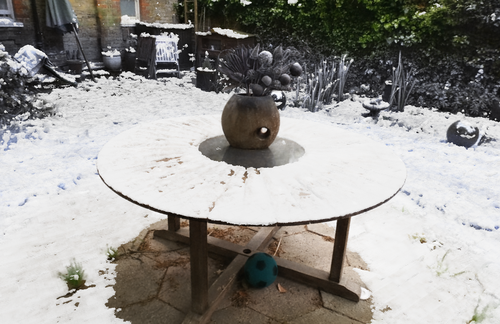} &
        \includegraphics[width=0.2\textwidth]{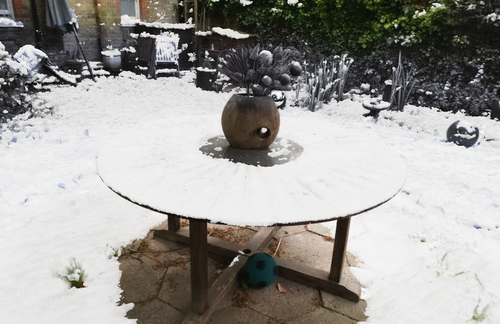} &
        \includegraphics[width=0.2\textwidth]{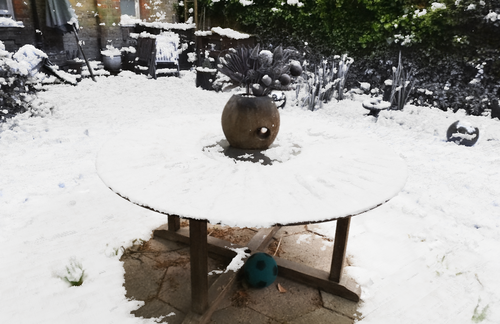} \\
    \end{tabular}
  }
   \captionof{figure}{\textbf{Controllable Climate Simulation.} From the top row to the bottom row, we control (1) smog color (2) water clearness (3) watercolor (4) spatial frequency of the wave. The frequency decreases from left to right. (5) snow height. We control snow height by adjusting the threshold density $\tau_{\text{snow}}$. }
   \label{fig:control}
\end{table*}

%% file: supp/qualitative.tex
\section{Qualitative Results}\label{sect:qualitative}
We demonstrate more qualitative results is Fig.~\ref{fig:quality_smog}, Fig.~\ref{fig:quality_flood}, Fig.~\ref{fig:quality_flood_2}, Fig.~\ref{fig:quality_snow}, and Fig.~\ref{fig:kitti}. For smog scene images in Fig.~\ref{fig:quality_smog}, ClimateGAN~\cite{schmidt2021climategan} generates visually plausible results but fails to provide sharp boundaries, and 3D stylization attempts to change the surface texture but makes the images overall darker. Our method simulates realistic visibility reduction effects caused by smog, thanks to the geometry reconstruction.  

The flood images are shown in Fig.~\ref{fig:quality_flood}. ClimateGAN++~\cite{schmidt2021climategan} cannot reconstruct realistic reflection on the water surface, Stable Diffusion~\cite{rombach2021highresolution} synthesize realistic water appearance but also produce random objects (e.g., cars, signs) in the scene, which is not consistent across views. \method~simulates realistic reflection and water ripples while being view-consistent. This is better demonstrated in the supp video and website. In Fig.~\ref{fig:quality_flood_2}, we compare with other NeRF-editing baselines. We first edit training images with ClimateGAN~\cite{schmidt2021climategan} or 2D stylization~\cite{li2018closed} and optimize the neural radiance field. While these baseline methods produce more view-consistent rendering, they either generate severe floating artifacts to compensate for the image inconsistency or cannot perform geometric editing, which sabotages realism.

We also compare our FastPhotoStyle~\cite{li2018closed} based stylization method with Artistic Radiance fields~\cite{zhang2022arf}. As shown in Fig.~\ref{fig:arf_compare}, we sustain more appearance details from the original scene. \method~can apply flood simulation pipeline to synthesize rising sea levels in large-scale scenes, as shown in Fig.~\ref{fig:sydney},~\ref{fig:rising}. Visualizing the consequences of such effects can raise general public awareness and call for government actions against climate change

\input{tables/quality_smog.tex}

\input{tables/quality_flood.tex}

\input{tables/quality_flood_2}

\input{tables/quality_snow.tex}

\input{tables/kitti360.tex}
\input{figures/method/stylization}
\input{tables/arf_compare.tex}
\input{tables/4_view_consistency.tex}
\input{figures/limitations/limitations.tex}

%% file: tables/quality_smog.tex
\begin{table*}[t]
    \centering
    \footnotesize
    \setlength\tabcolsep{0.05em} %
    \resizebox{\textwidth}{!}{
    \begin{tabular}{cccc}

    \includegraphics[width=0.24\textwidth]{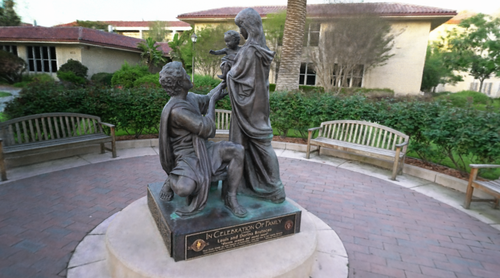} 
    & \includegraphics[width=0.24\textwidth]{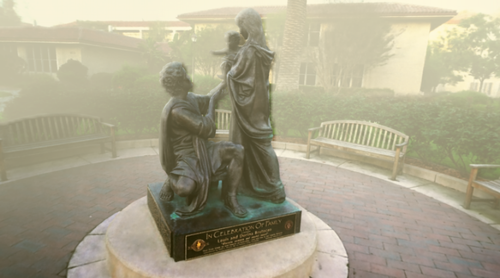} 
    & \includegraphics[width=0.24\textwidth]{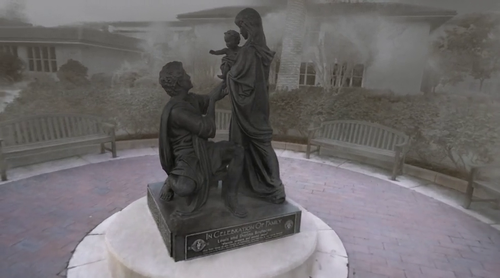}
    & \includegraphics[width=0.24\textwidth]{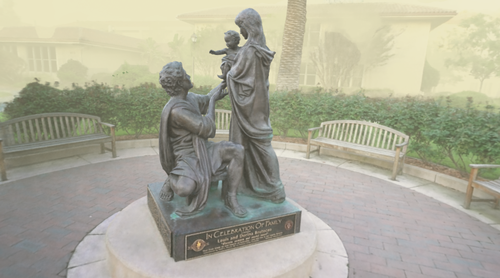}
    \\

    \includegraphics[width=0.24\textwidth]{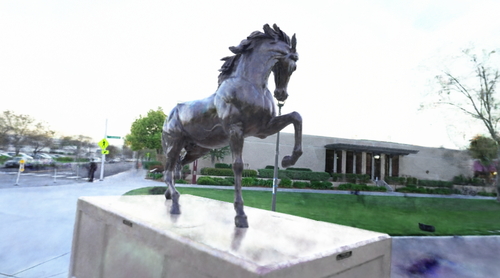} 
    & \includegraphics[width=0.24\textwidth]{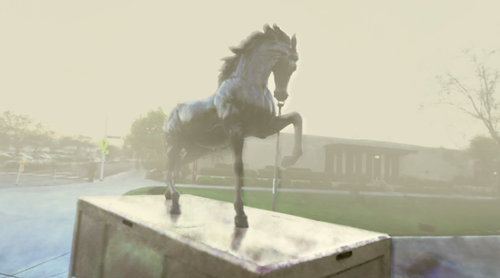} 
    & \includegraphics[width=0.24\textwidth]{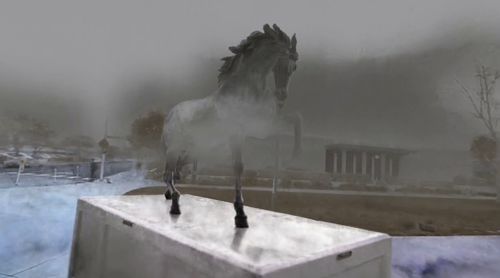}
    & \includegraphics[width=0.24\textwidth]{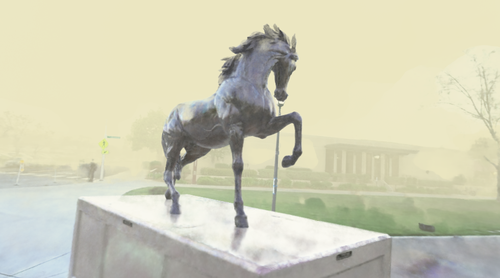}
    \\

    \includegraphics[width=0.24\textwidth]{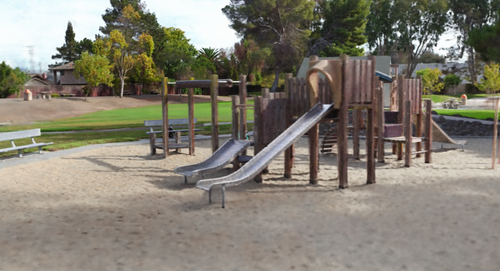} 
    & \includegraphics[width=0.24\textwidth]{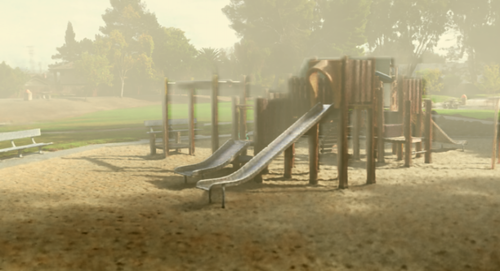} 
    & \includegraphics[width=0.24\textwidth]{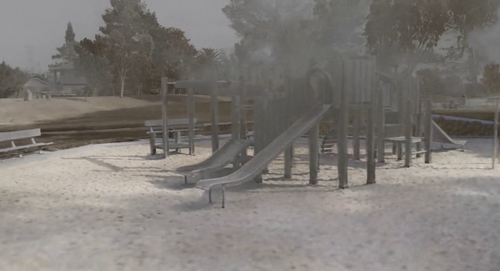}
    & \includegraphics[width=0.24\textwidth]{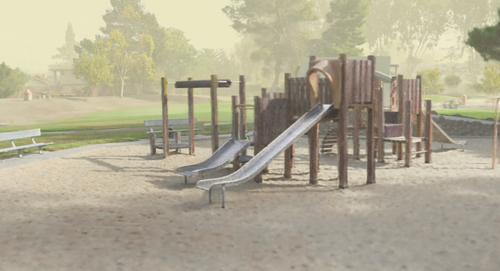}
    \\

    \includegraphics[width=0.24\textwidth]{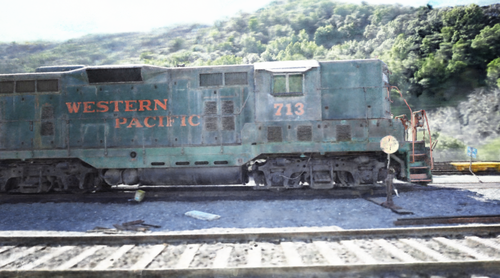} 
    & \includegraphics[width=0.24\textwidth]{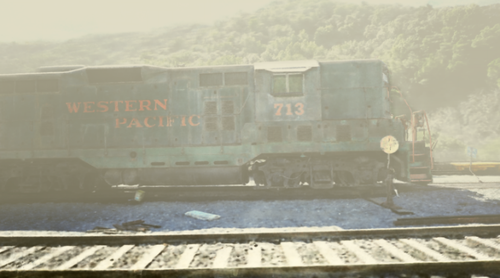} 
    & \includegraphics[width=0.24\textwidth]{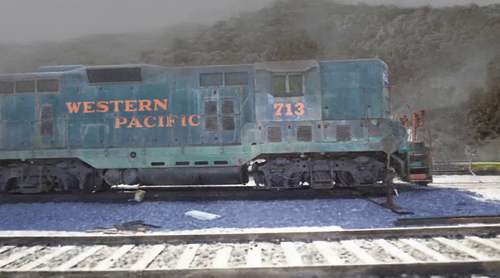}
    & \includegraphics[width=0.24\textwidth]{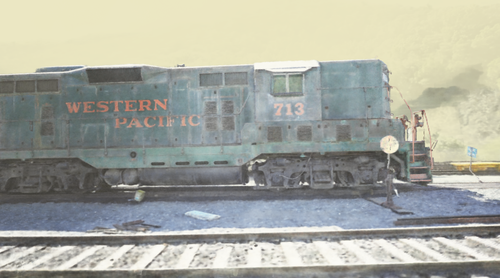}
    \\

    \includegraphics[width=0.24\textwidth]{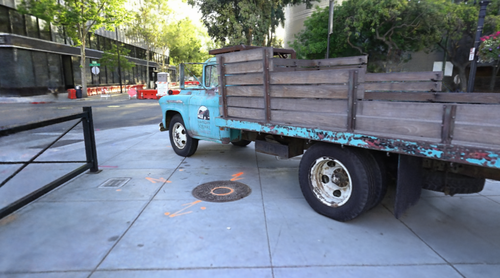} 
    & \includegraphics[width=0.24\textwidth]{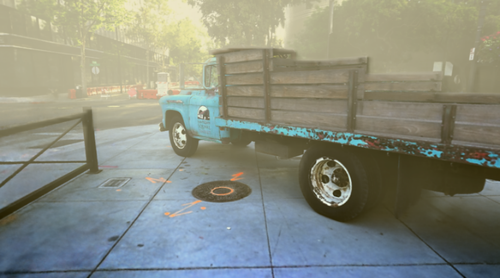} 
    & \includegraphics[width=0.24\textwidth]{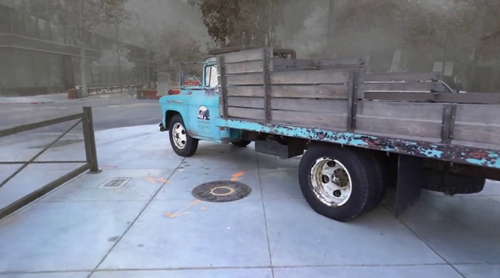}
    & \includegraphics[width=0.24\textwidth]{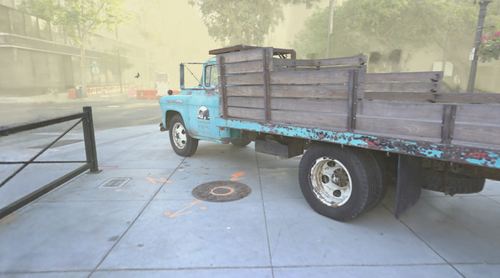}
    \\
    
     Original & ClimateGAN~\cite{schmidt2021climategan} & 3D Stylization & Ours \\

    \end{tabular}
    }
    \captionof{figure}{\textbf{Smog simulation comparison.} }
    \label{fig:quality_smog}
\end{table*}

%% file: tables/quality_flood.tex
\begin{table*}[t]
    \centering
    \footnotesize
    \setlength\tabcolsep{0.05em} %
    \resizebox{\textwidth}{!}{
    \begin{tabular}{cccc}

    \includegraphics[width=0.24\textwidth]{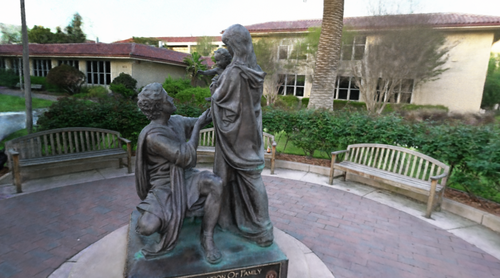} 
    & \includegraphics[width=0.24\textwidth]{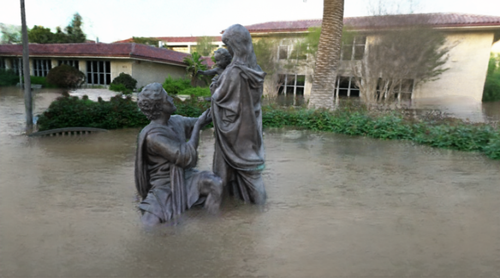} 
    & \includegraphics[width=0.24\textwidth]{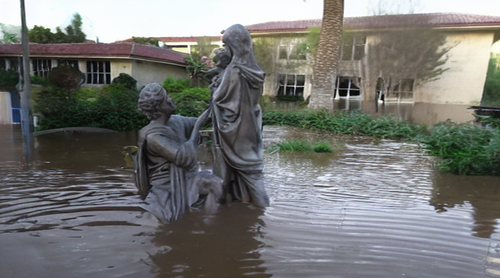}
    & \includegraphics[width=0.24\textwidth]{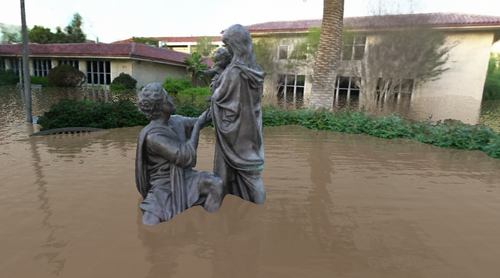}
    \\

    \includegraphics[width=0.24\textwidth]{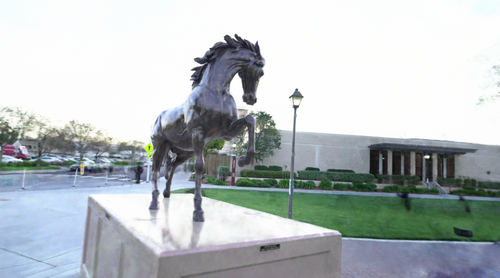} 
    & \includegraphics[width=0.24\textwidth]{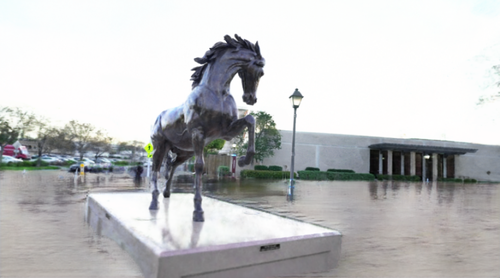} 
    & \includegraphics[width=0.24\textwidth]{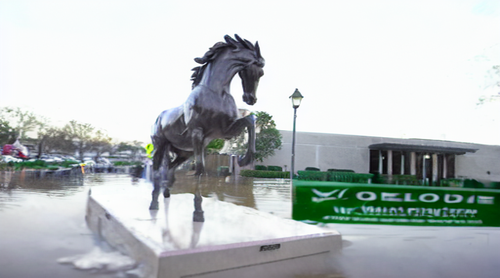}
    & \includegraphics[width=0.24\textwidth]{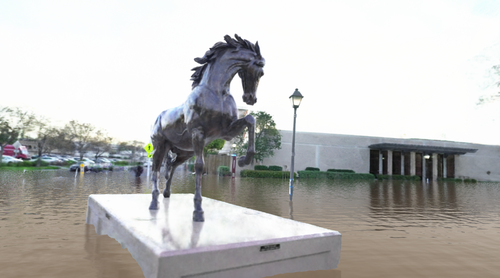}
    \\

    \includegraphics[width=0.24\textwidth]{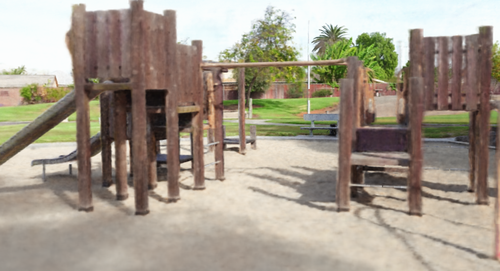} 
    & \includegraphics[width=0.24\textwidth]{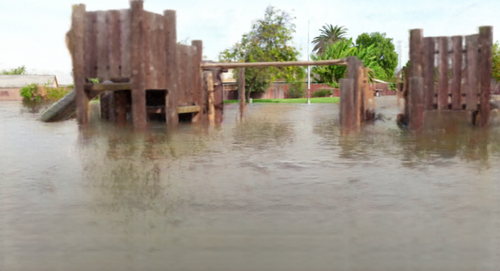} 
    & \includegraphics[width=0.24\textwidth]{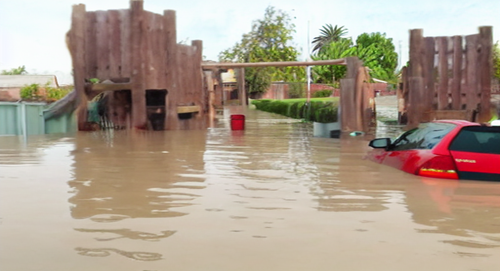}
    & \includegraphics[width=0.24\textwidth]{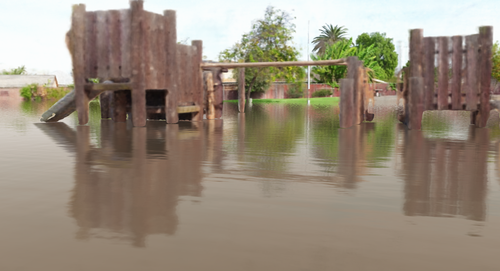}
    \\

    \includegraphics[width=0.24\textwidth]{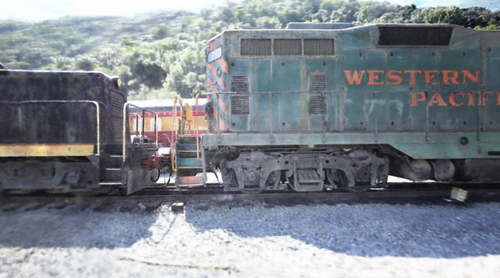} 
    & \includegraphics[width=0.24\textwidth]{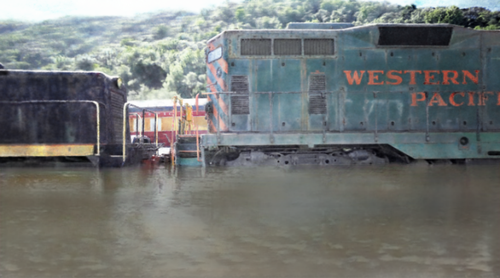} 
    & \includegraphics[width=0.24\textwidth]{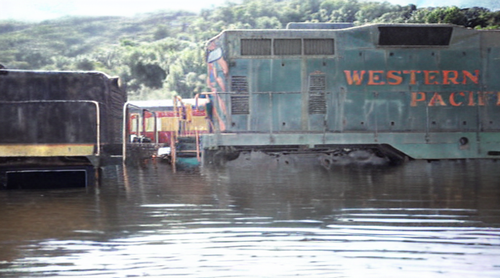}
    & \includegraphics[width=0.24\textwidth]{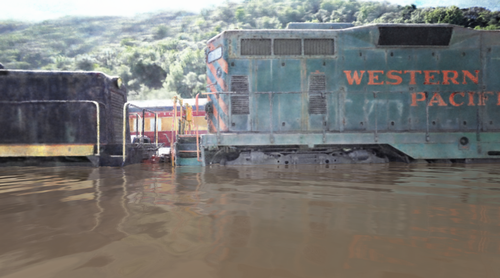}
    \\
    
    \includegraphics[width=0.24\textwidth]{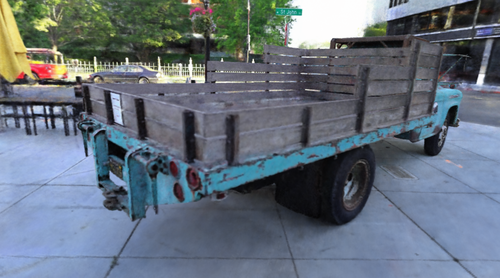} 
    & \includegraphics[width=0.24\textwidth]{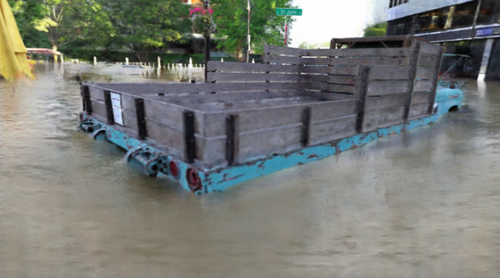} 
    & \includegraphics[width=0.24\textwidth]{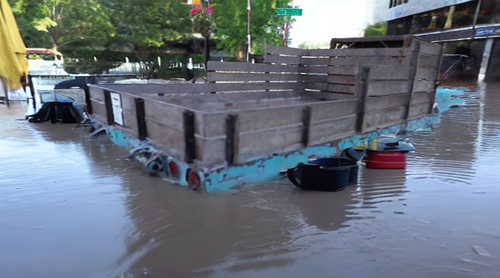}
    & \includegraphics[width=0.24\textwidth]{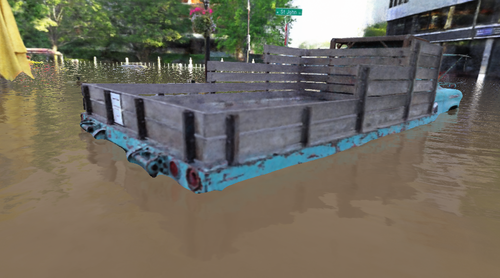}
    \\

     Original & ClimateGAN++~\cite{schmidt2021climategan} & Stable Diffusion~\cite{rombach2021highresolution} & Ours \\

    \end{tabular}
    }
    \captionof{figure}{\textbf{Flood simulation comparison.} }
    \label{fig:quality_flood}
\end{table*}

%% file: tables/quality_flood_2.tex
\begin{table*}[t]
    \centering
    \footnotesize
    \setlength\tabcolsep{0.05em} %
    \resizebox{\textwidth}{!}{
    \begin{tabular}{cccc}

    \includegraphics[width=0.24\textwidth]{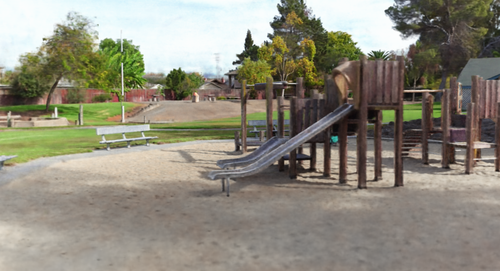} 
    & \includegraphics[width=0.24\textwidth]{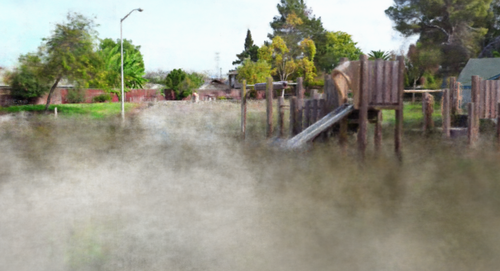} 
    & \includegraphics[width=0.24\textwidth]{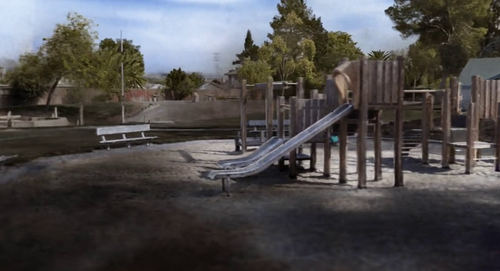}
    & \includegraphics[width=0.24\textwidth]{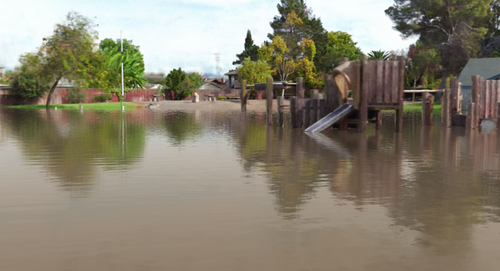}
    \\

    \includegraphics[width=0.24\textwidth]{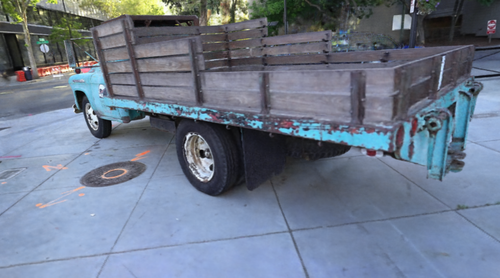} 
    & \includegraphics[width=0.24\textwidth]{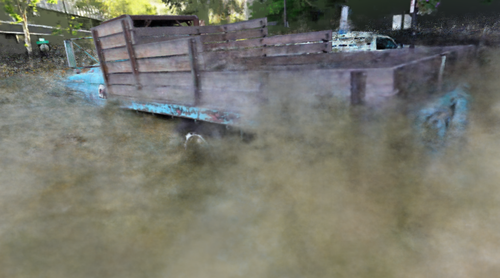} 
    & \includegraphics[width=0.24\textwidth]{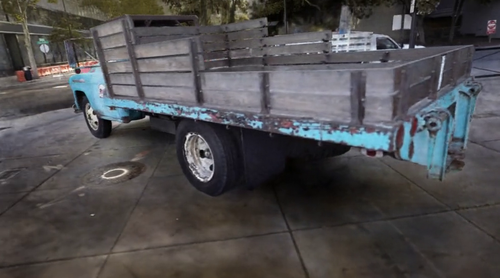}
    & \includegraphics[width=0.24\textwidth]{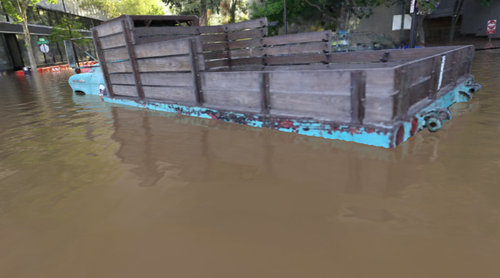}
    \\
     Original & ClimateGAN~\cite{schmidt2021climategan}+NeRF & 3D Stylization & Ours \\

    \end{tabular}
    }
    \captionof{figure}{\textbf{NeRF editing for flood simulation.} 
We tested the {\it ClimateGAN image + NeRF training} baseline, but the results are blurry due to inconsistent inputs. {\it 3D stylization} baseline is a NeRF trained on stylized images.}
    \label{fig:quality_flood_2}
\end{table*}

%% file: tables/quality_snow.tex
\begin{table*}[t]
    \centering
    \footnotesize
    \setlength\tabcolsep{0.05em} %
    \resizebox{\textwidth}{!}{
    \begin{tabular}{cccc}

    \includegraphics[width=0.24\textwidth]{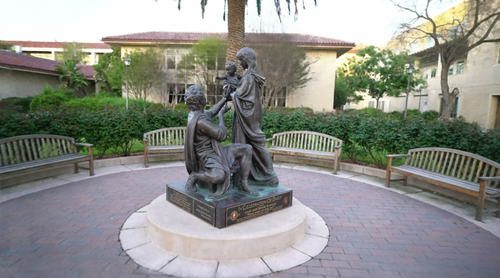} 
    & \includegraphics[width=0.24\textwidth]{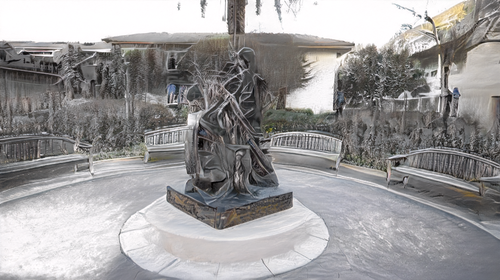} 
    & \includegraphics[width=0.24\textwidth]{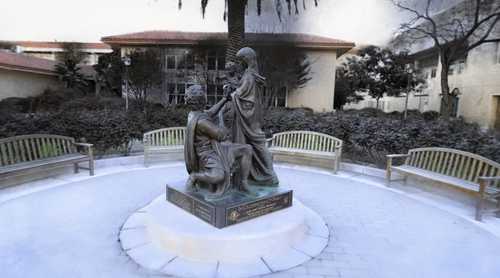} 
    & \includegraphics[width=0.24\textwidth]{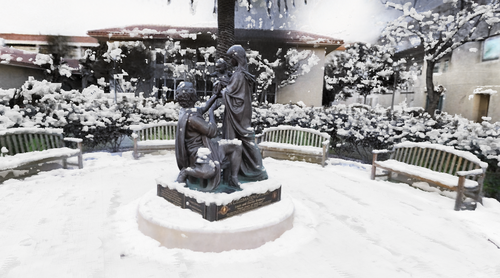} 
    \\

    \includegraphics[width=0.24\textwidth]{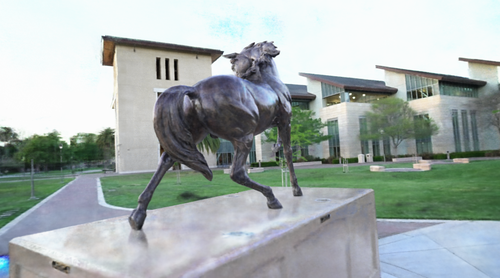} 
    & \includegraphics[width=0.24\textwidth]{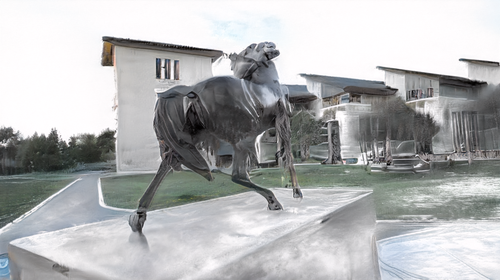} 
    & \includegraphics[width=0.24\textwidth]{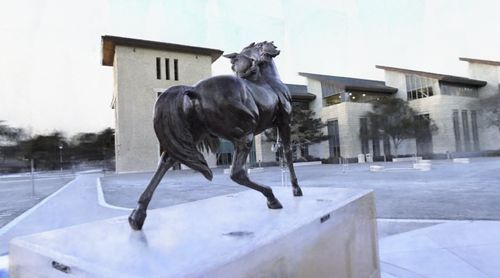}
    & \includegraphics[width=0.24\textwidth]{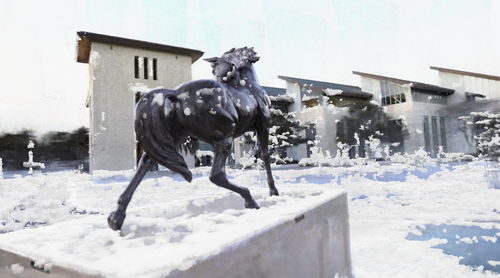}
    \\

    \includegraphics[width=0.24\textwidth]{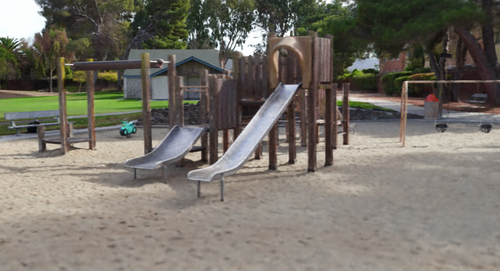} 
    & \includegraphics[width=0.24\textwidth]{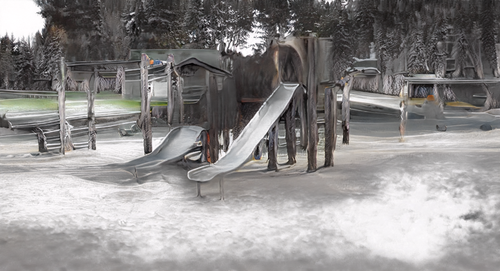} 
    & \includegraphics[width=0.24\textwidth]{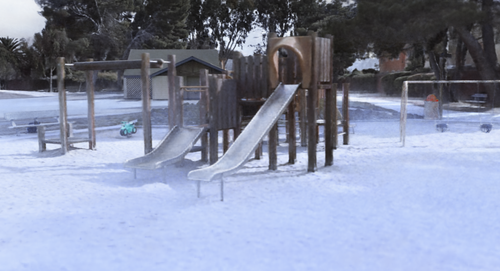}
    & \includegraphics[width=0.24\textwidth]{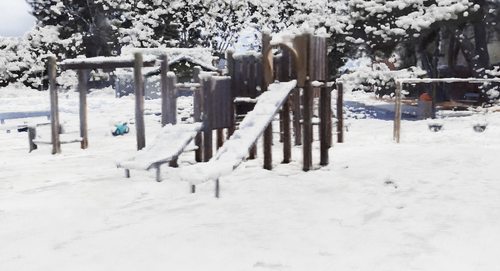}
    \\

    \includegraphics[width=0.24\textwidth]{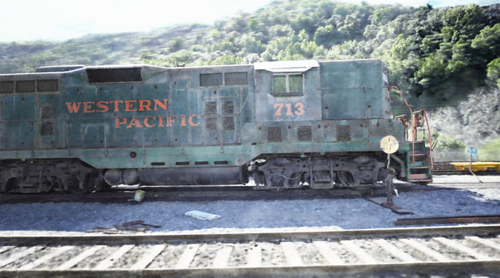} 
    & \includegraphics[width=0.24\textwidth]{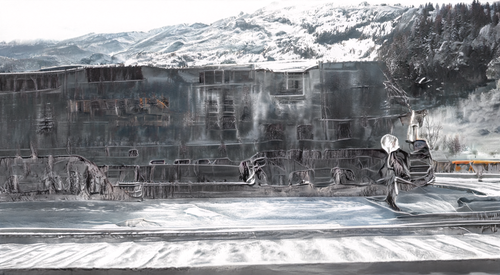} 
    & \includegraphics[width=0.24\textwidth]{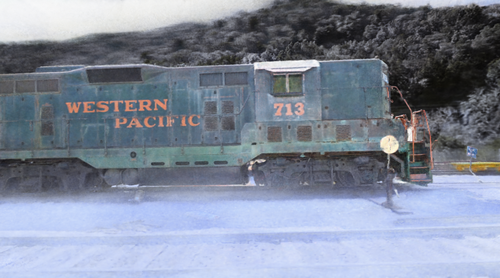}
    & \includegraphics[width=0.24\textwidth]{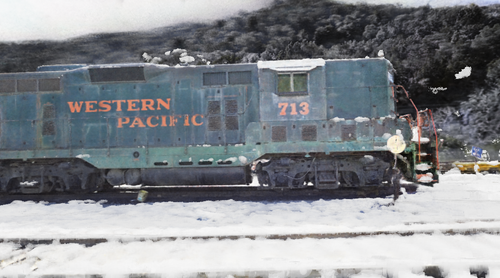}
    \\
    
    \includegraphics[width=0.24\textwidth]{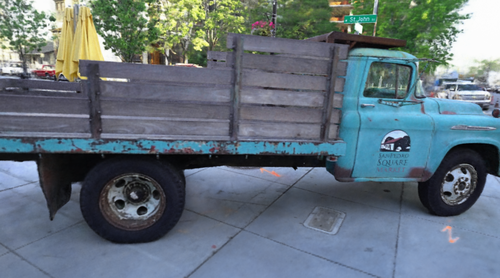} 
    & \includegraphics[width=0.24\textwidth]{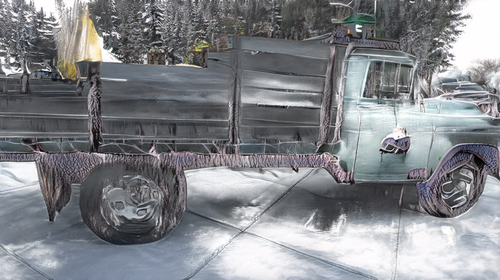} 
    & \includegraphics[width=0.24\textwidth]{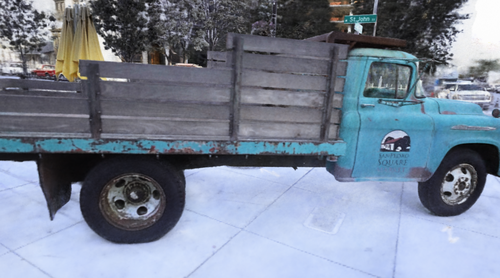}
    & \includegraphics[width=0.24\textwidth]{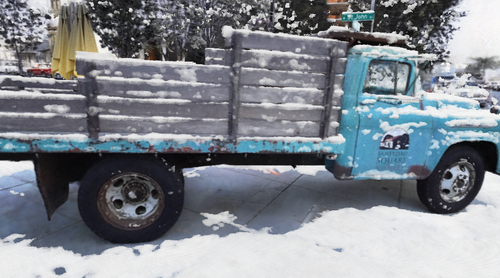}
    \\

    \includegraphics[width=0.24\textwidth]{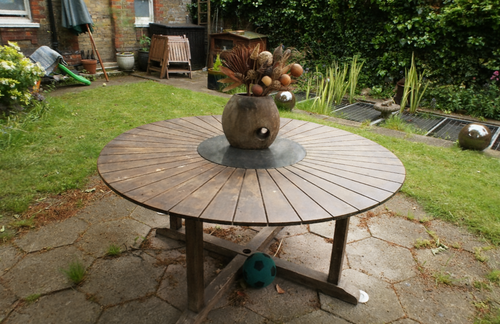} 
    & \includegraphics[width=0.24\textwidth]{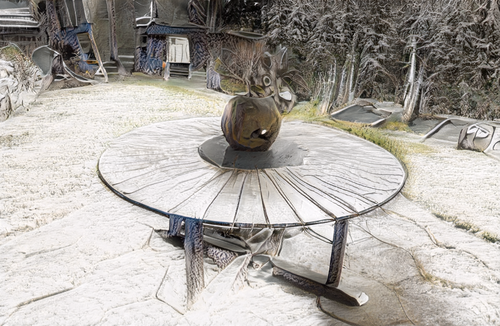} 
    & \includegraphics[width=0.24\textwidth]{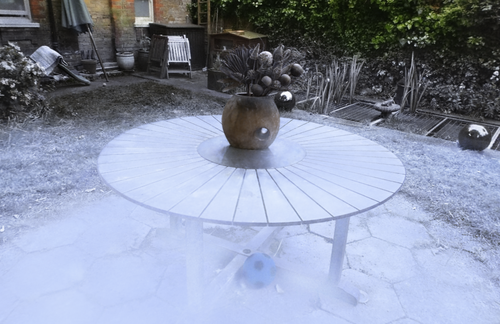}
    & \includegraphics[width=0.24\textwidth]{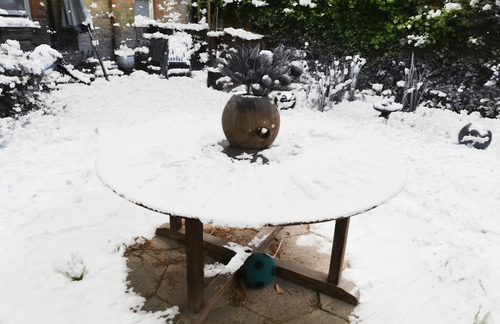}
    \\
    
     Original & Swapping Autoencoder~\cite{park2020swapping} & 3D stylization & Ours \\

    \end{tabular}
    }
    \captionof{figure}{\textbf{Snow simulation comparison.} }
    \label{fig:quality_snow}
\end{table*}

%% file: tables/kitti360.tex
\begin{table*}[t]
    \centering
    \footnotesize
    \setlength\tabcolsep{0.05em} %
    \resizebox{\textwidth}{!}{
    \begin{tabular}{ccc}

    \includegraphics[width=0.3\textwidth]{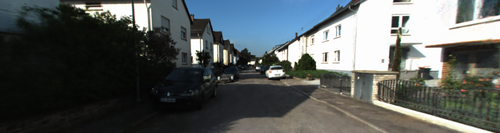} 
    & \includegraphics[width=0.3\textwidth]{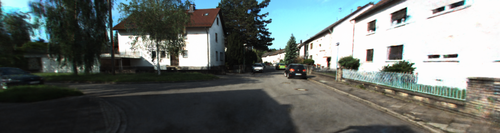} 
    & \includegraphics[width=0.3\textwidth]{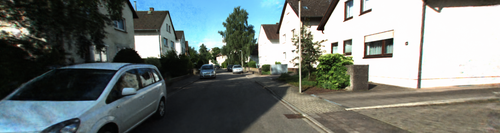}
    \\

    \includegraphics[width=0.3\textwidth]{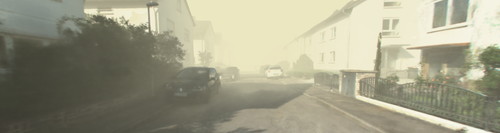} 
    & \includegraphics[width=0.3\textwidth]{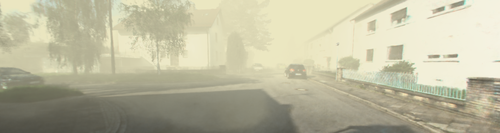} 
    & \includegraphics[width=0.3\textwidth]{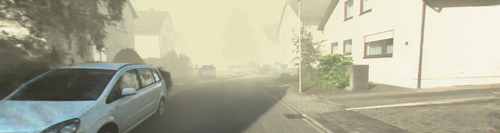}
    \\

    \includegraphics[width=0.3\textwidth]{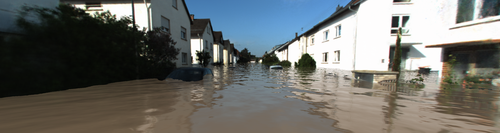} 
    & \includegraphics[width=0.3\textwidth]{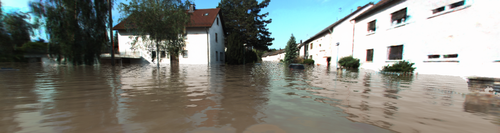} 
    & \includegraphics[width=0.3\textwidth]{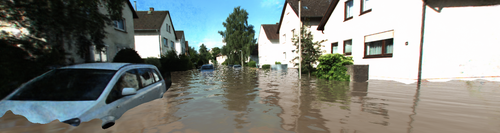}
    \\

    \includegraphics[width=0.3\textwidth]{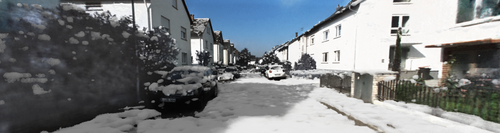} 
    & \includegraphics[width=0.3\textwidth]{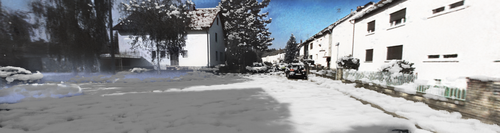} 
    & \includegraphics[width=0.3\textwidth]{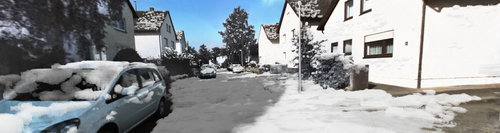}

    \end{tabular}
    }
    \captionof{figure}{\textbf{Simulation on Urban Driving Scenes.} }
    \label{fig:kitti}
\end{table*}

%% file: figures/method/stylization.tex
\begin{table*}[t]
  \centering
  \footnotesize
  \setlength\tabcolsep{0.05em}
  \resizebox{\textwidth}{!}{
    \begin{tabular}{cccc}

        \includegraphics[width=0.24\textwidth]{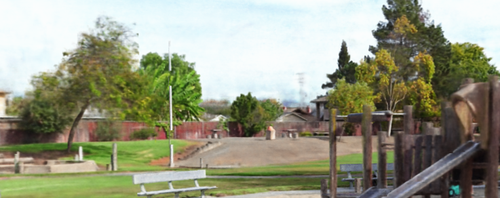} &
        \includegraphics[width=0.24\textwidth]{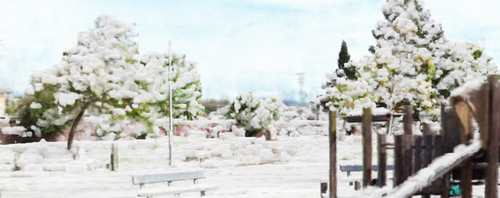} &
        \includegraphics[width=0.24\textwidth]{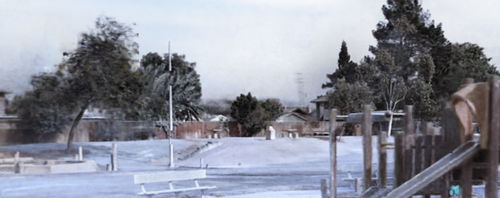} &
        \includegraphics[width=0.24\textwidth]{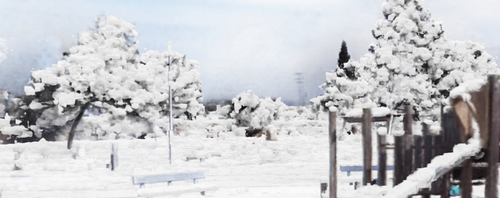} \\
        
        (a) Original NeRF rendering & (b) Original NeRF + snow & (c) NeRF with stylization & (d) NeRF with stylization + snow

    \end{tabular}
  }

   \captionof{figure}{\textbf{Stylization.} 
Performing snow simulation with the original NeRF model leads to incompatible scene appearance (e.g., snow on green vegetations).
We address this issue by first finetuning the appearance of the NeRF model to match the style of the provided style image. 
Our snow simulation on the stylized NeRF model shows visually more appealing results. }
   \label{fig:4_stylization}
\end{table*}

%% file: tables/arf_compare.tex
\begin{table*}[t]
  \centering
  \footnotesize
  \setlength\tabcolsep{0.05em}
  \resizebox{\textwidth}{!}{
    \begin{tabular}{cccc}
        \includegraphics[trim=0cm 3cm 0cm 0.5cm, clip, width=0.2\textwidth]{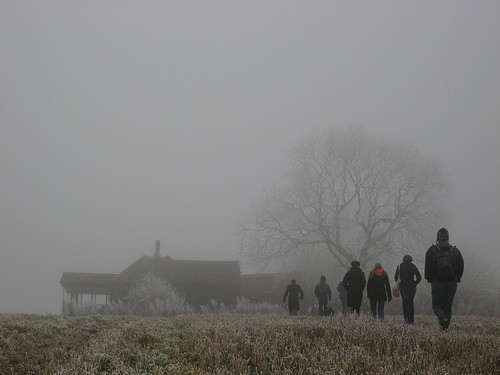} &
        \includegraphics[width=0.2\textwidth]{figures/quality_snow/Family.png} &
        \includegraphics[width=0.2\textwidth]{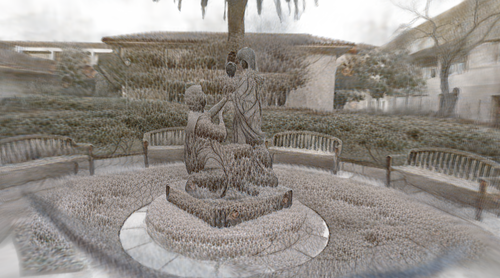} &
        \includegraphics[width=0.2\textwidth]{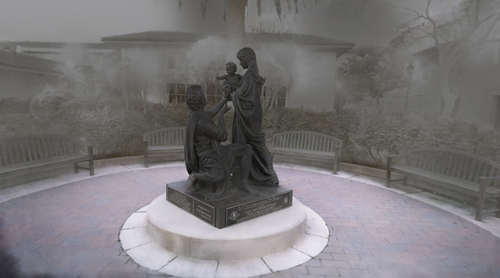} \\
        
        \includegraphics[trim=0cm 1.8cm 0cm 0.5cm, clip, width=0.2\textwidth]{figures/arf/flood_style_image.jpg} &
        \includegraphics[width=0.2\textwidth]{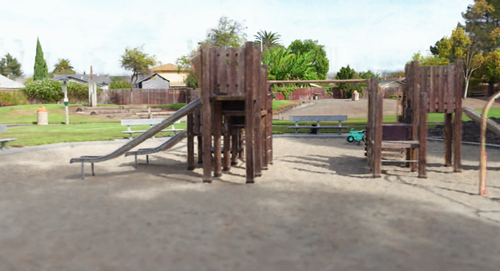} &
        \includegraphics[width=0.2\textwidth]{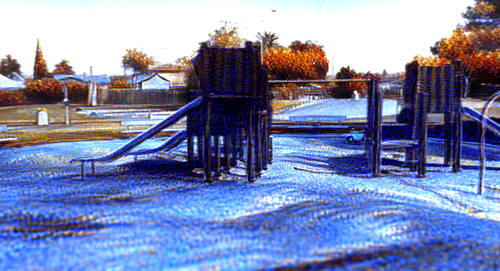} &
        \includegraphics[width=0.2\textwidth]{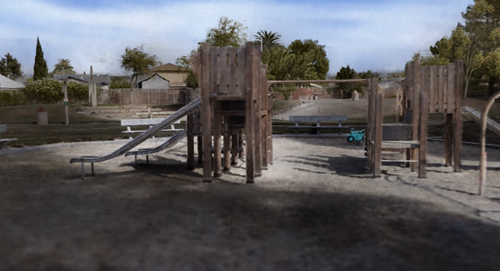} \\
        
        \includegraphics[trim=0cm 3.6cm 0cm 0.5cm, clip, width=0.2\textwidth]{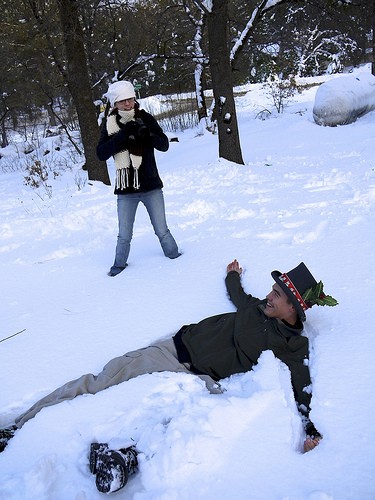} &
        \includegraphics[width=0.2\textwidth]{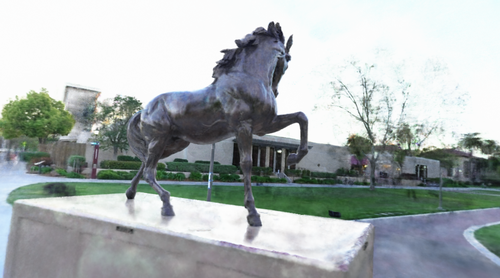} &
        \includegraphics[width=0.2\textwidth]{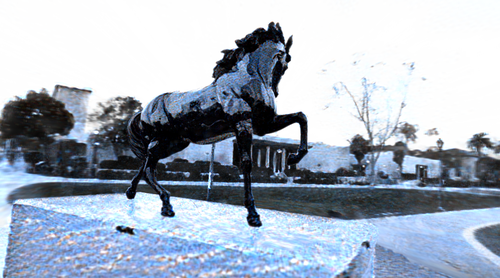} &
        \includegraphics[width=0.2\textwidth]{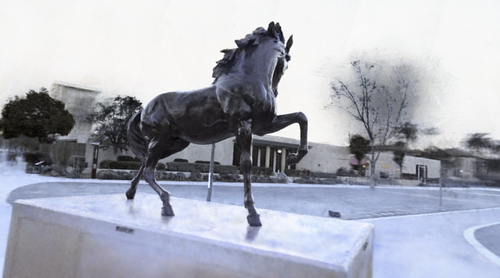} \\
        
        Style Image from ~\cite{young2014image} &
        Original &
        ARF ~\cite{zhang2022arf}&
        3D stylization
    \end{tabular}
  }
   \captionof{figure}{\textbf{Comparison with Artistic Radiance Fields~\cite{zhang2022arf}}}
   \label{fig:arf_compare}
\end{table*}

%% file: tables/4_view_consistency.tex
\begin{table*}[t]
    \centering
    \setlength\tabcolsep{0.5pt} %
    \footnotesize
    \begin{tabular}{lccccc}

     \raisebox{2.2\normalbaselineskip}[0pt][0pt]{\rotatebox[origin=c]{90}{\tiny{ClimateGAN++~\cite{schmidt2021climategan}}}} & 
     \includegraphics[width=0.19\textwidth]{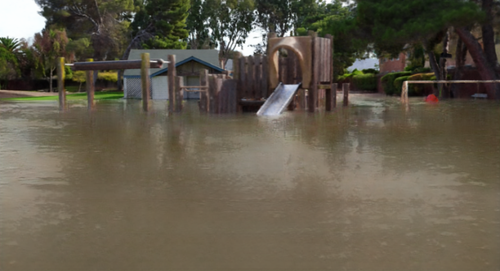} &
     \includegraphics[width=0.19\textwidth]{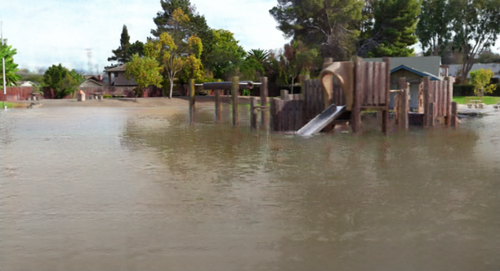}
     & \includegraphics[width=0.19\textwidth]{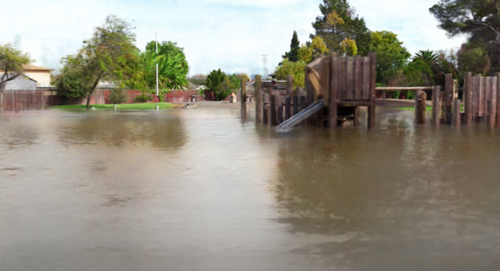} &
     \includegraphics[width=0.19\textwidth]{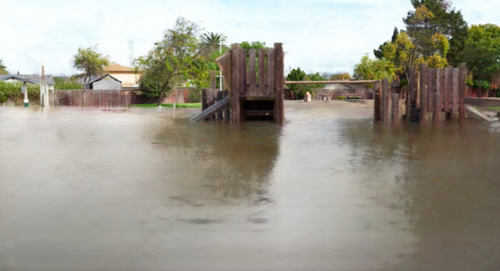}
     & \includegraphics[width=0.19\textwidth]{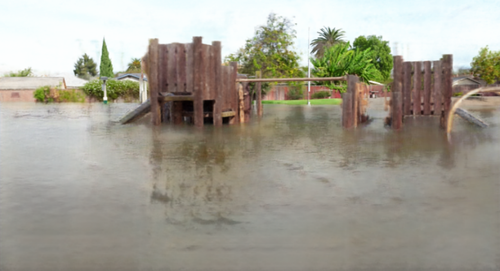} \\

     \raisebox{2.2\normalbaselineskip}[0pt][0pt]{\rotatebox[origin=c]{90}{\tiny{Stable Diffusion~\cite{rombach2021highresolution}}}} & 
     \includegraphics[width=0.19\textwidth]{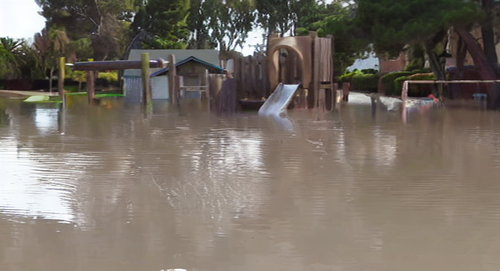}
     & \includegraphics[width=0.19\textwidth]{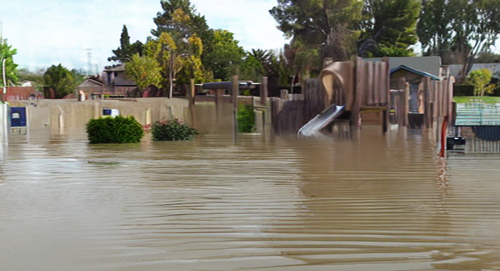}
     & \includegraphics[width=0.19\textwidth]{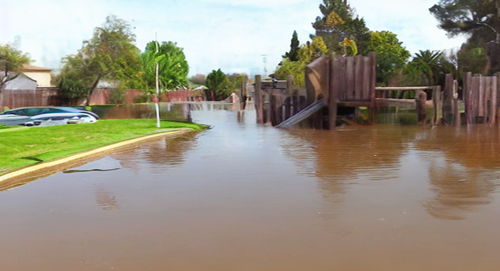}
     & \includegraphics[width=0.19\textwidth]{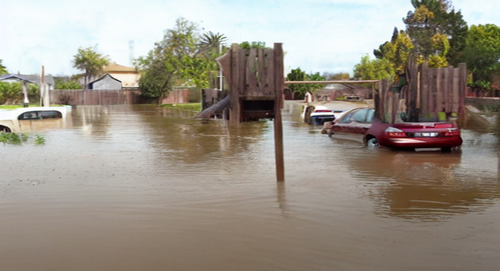}
     & \includegraphics[width=0.19\textwidth]{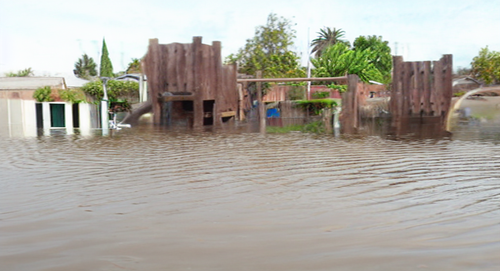} \\

     \raisebox{2\normalbaselineskip}[0pt][0pt]{\rotatebox[origin=c]{90}{\scriptsize{Ours}}} 
     & 
     \includegraphics[width=0.19\textwidth]{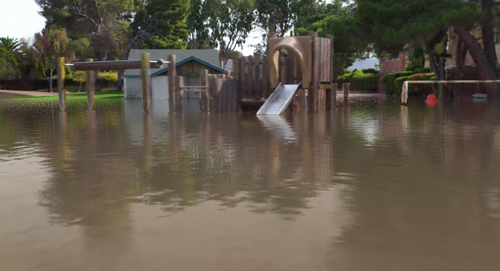}
     & \includegraphics[width=0.19\textwidth]{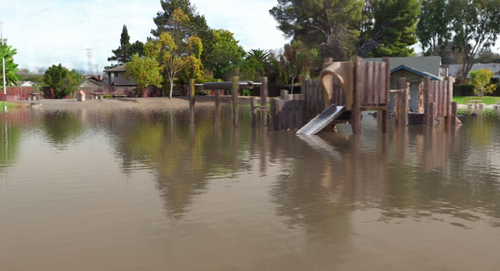}
     & \includegraphics[width=0.19\textwidth]{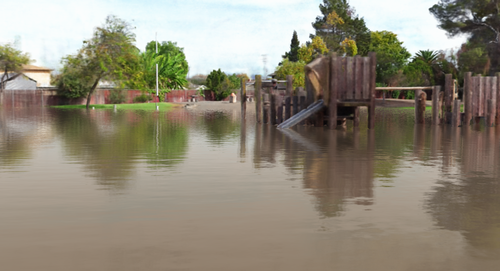}
     & \includegraphics[width=0.19\textwidth]{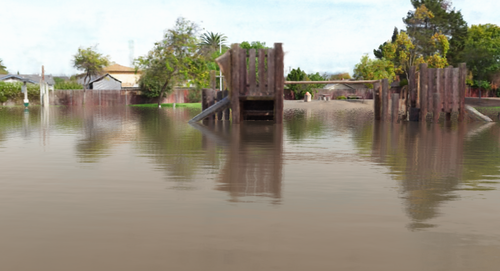}
     & \includegraphics[width=0.19\textwidth]{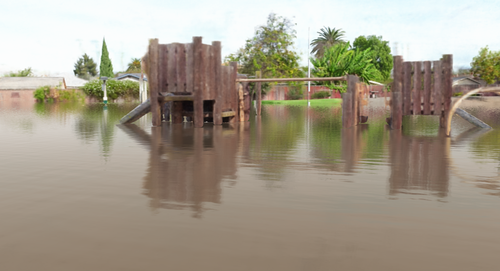} \\
     
     & $t=0$ & $t=40$ & 
     $t=80$ & $t=120$ & 
     $t=160$
    
    \end{tabular}
    \vspace{-2mm}
    \captionof{figure}{\textbf{View Consistency Comparison.}  We show flood simulation in Playground scene and time step $t$ increases from left to right. ClimateGAN++~\cite{schmidt2021climategan} (Top) cannot generate realistic reflection, Stable Diffusion~\cite{rombach2021highresolution} (Middle) synthesizes different objects in each views. Ours (Bottom) simulate photorealistic and consistent reflection and ripples.}
    \label{tab:4_view_consistency}
    \vspace{-3mm}
\end{table*}

%% file: figures/limitations/limitations.tex
\begin{table*}[t]
    \centering
    \tiny
    \setlength\tabcolsep{0.05em} %
    \resizebox{\linewidth}{!}{
    \begin{tabular}{cccc}

    \includegraphics[width=0.24\textwidth]{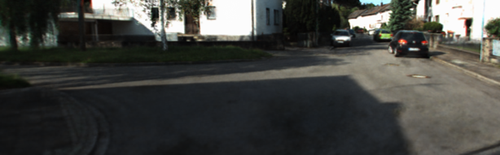}
    & \includegraphics[width=0.24\textwidth]{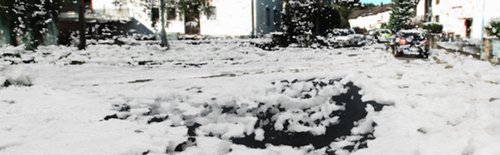} &
    \includegraphics[width=0.24\textwidth]{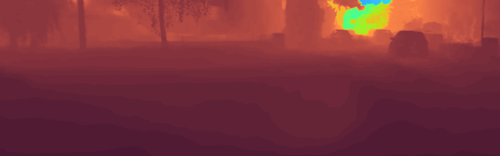}
    & \includegraphics[width=0.24\textwidth]{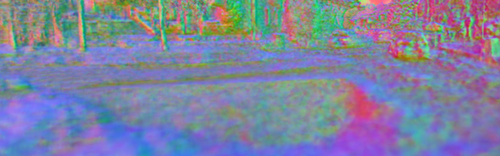}\\
    
    (a) Original NeRF rendering & 
      (b) Climate NeRF rendering  &    
      (c) Depth rendering & 
      (d) Surface normal rendering  

    \end{tabular}
    }
    \vspace{-3mm}
    \captionof{figure}{\textbf{Limitations.} 
    Snow simulation on KITT-360~\cite{liao2022kitti} dataset fails to cover a shadowed road due to wrong geometry. }
    \label{fig:6_limitations}
\end{table*}

%% file: supp/quantitative.tex
\section{Quantitative Results}\label{sect:quantitative}

No automatic quantitative score can holistically evaluate the quality of our weather-simulated movies. In this project, we evaluate the synthesized videos with the state-of-the-art video quality assessment model UVQ~\cite{wang2021rich} and report the results in Table~\ref{tab:vqa}. 
The score ranges between interval $(1, 5)$, where 1 indicates the lowest quality and 5 indicates the highest quality. As the table shows, our smog simulation outperforms all other baselines, while it does not win Stable Diffusion~\cite{rombach2021highresolution} in flood simulation and ClimateGAN~\cite{schmidt2021climategan} in snow simulation. That being said, UVQ prefers sharp videos instead of measuring holistic realism. As shown in Fig.~\ref{fig:quality_snow}, baselines get a better quality score despite providing low-quality snow simulation results, suggesting UVQ might not be a good metric for our task. Hence, despite demonstrating UVQ~\cite{wang2021rich} results, we want to emphasize that such metrics mainly focus on measuring the amount of low-level degradation (such as blurriness and noise), which cannot faithfully reproduce human evaluation on realism. Having a good video quality score on simulation remains an open topic. 

\input{tables/vqa.tex}

%% file: tables/vqa.tex
\begin{table*}[t]
  \centering
  \footnotesize
  \setlength\tabcolsep{0.05em}
  {
    \begin{tabular}{|l|c||c|c|c||c|c|c|c||c|c|c|}
    \hline
    & & \multicolumn{3}{c||}{\textbf{Smog}} & \multicolumn{4}{c||}{\textbf{Flood}} & \multicolumn{3}{c|}{\textbf{Snow}} \\
    & Original & \makecell{GAN} & \makecell{3D Style} & Ours & \makecell{GAN} & \makecell{GAN++} & \makecell{Diffusion} & Ours & \makecell{GAN} & \makecell{3D Style} & Ours \\
 \hline 
 Family     & 3.434 & 3.426 & 3.425 & \textbf{3.437} & 3.408 & 3.413 & 3.416 & \textbf{3.424} & \textbf{3.434} & \textbf{3.434} & 3.431 \\
 \hline 
 Horse      & 3.426 & 3.422 & 3.422 & \textbf{3.432} & 3.409 & 3.412 & 3.416 & \textbf{3.422} & \textbf{3.435} & 3.424 & 3.421 \\
 \hline 
 Playground & 3.409 & 3.402 & 3.403 & \textbf{3.412} & 3.400 & 3.402 & \textbf{3.409} & 3.404 & \textbf{3.427} & 3.412 & 3.415 \\
 \hline 
 Train      & 3.406 & 3.396 & \textbf{3.407} & \textbf{3.407} & 3.401 & 3.402 & \textbf{3.410} & 3.407 & \textbf{3.417} & 3.411 & 3.408 \\
 \hline 
 Truck      & 3.425 & 3.424 & 3.424 & \textbf{3.431} & 3.404 & 3.403 & \textbf{3.424} & 3.413 & \textbf{3.431} & 3.424 & 3.416 \\
 \hline 
    \end{tabular}
  }
   \caption{\textbf{Video Quality Assessment.} We evaluate the video quality with Google's Universal Video Quality (UVQ) model~\cite{wang2021rich}. }
   \label{tab:vqa}
\end{table*}